\documentclass[11pt,oneside,a4paper]{article}
\usepackage[top=1 in, bottom=1 in, left=0.85 in, right=0.85 in]{geometry}
\usepackage{hyperref}
\usepackage{float}
\usepackage{amsmath,amssymb,graphicx,url}
\usepackage{amsthm}
\usepackage{thmtools,thm-restate,wrapfig,enumitem,mathabx}
\usepackage{xcolor}
\usepackage{algorithm}
\usepackage{algorithmic}
\usepackage{bbold}
\usepackage[utf8]{inputenc}
\usepackage[T1]{fontenc}
\usepackage{booktabs}
\usepackage{amsfonts}
\usepackage{nicefrac}
\usepackage{microtype}
\usepackage{xcolor}
\usepackage{subcaption}
\usepackage{multirow}
\usepackage{bbm}
\usepackage{mathtools}
\usepackage[compress]{cite}
\usepackage[thinc]{esdiff}

\newtheorem{assumption}{Assumption}
\newtheorem{theorem}{Theorem}

\newtheorem{lemma}{Lemma}
\newtheorem{corollary}{Corollary}

\definecolor{darkgreen}{RGB}{25,107,36}
\definecolor{darkteal}{RGB}{21,96,130}

\newcommand{\blue}[1]{\textcolor{blue}{#1}}

\newcommand{\expt}{\mathbb{E}}
\newcommand{\prob}{\mathbb{P}}
\let\oldnl\nl
\newcommand{\nonl}{\renewcommand{\nl}{\let\nl\oldnl}}

\newcommand{\lIF}[2]{\STATE \textbf{if} #1 \textbf{then} #2}
\newcommand{\lELSIF}[2]{\STATE \textbf{else if} #1 \textbf{then} #2}
\newcommand{\lELSE}[1]{\STATE \textbf{else} #1}

\allowdisplaybreaks

\title{Near-Optimal Regret-Queue Length Tradeoff\\in Online Learning for Two-Sided Markets}

\author{%
  Zixian Yang\\
  Department of Electrical Engineering and Computer Science\\
  University of Michigan, Ann Arbor\\
  \texttt{zixian@umich.edu}\\
  \and
  Sushil Mahavir Varma \\
  Department of Industrial and Operations Engineering\\
  University of Michigan, Ann Arbor\\
  \texttt{sushilv@umich.edu}
  \and
  Lei Ying \\
  Department of Electrical Engineering and Computer Science\\
  University of Michigan, Ann Arbor\\
  \texttt{leiying@umich.edu}
}
\date{}

\begin{document}

\maketitle

\begin{abstract}
We study a two-sided market, wherein, price-sensitive heterogeneous customers and servers arrive and join their respective queues.
A compatible customer-server pair can then be matched by the platform, at which point, they leave the system.
Our objective is to design pricing and matching algorithms that maximize the platform's profit, while maintaining reasonable queue lengths.
As the demand and supply curves governing the price-dependent arrival rates may not be known in practice, we design a novel online-learning-based pricing policy and establish its near-optimality. In particular, we prove a tradeoff among three performance metrics: $\tilde{O}(T^{1-\gamma})$ regret, $\tilde{O}(T^{\gamma/2})$ average queue length, and $\tilde{O}(T^{\gamma})$ maximum queue length for $\gamma \in (0, 1/6]$, significantly improving over existing results \cite{yang2024learning}. Moreover, barring the permissible range of $\gamma$, we show that this trade-off between regret and average queue length is optimal up to logarithmic factors under a class of policies, matching the optimal one as in \cite{varma2023dynamic} which assumes the demand and supply curves to be known.
Our proposed policy has two noteworthy features: a dynamic component that optimizes the tradeoff between low regret and small queue lengths; and a probabilistic component that resolves the tension between obtaining useful samples for fast learning and maintaining small queue lengths.
\end{abstract}

\section{Introduction}
\label{sec:intro}

We study a two-sided market, wherein, heterogeneous customers and servers arrive into the system and join their respective queues. Any compatible customer-server pair can then be matched, at which point they leave the system instantaneously. More formally, the system is described by a bipartite graph, where the vertices are customer and server queues while the edges represent compatibility between them.

Such a canonical two-sided/matching queueing model is useful in studying matching in emerging applications like online marketplaces \cite{varma2023dynamic}. For example, matching customers and drivers in ride-hailing systems, customers and couriers in meal-delivery platforms, tasks and workers on crowdsourcing platforms, etc. Furthermore, matching queues are a versatile tool for modeling diverse applications beyond online marketplaces. In particular, different variants of matching queues are prevalent in modeling applications like assemble-to-order systems \cite{gurvich2015dynamic}, payment channel networks \cite{varma2021throughput}, and quantum switches \cite{zubeldia_quantum}.  

Motivated by these applications, we study matching in a two-sided queueing model with an additional lever of modulating the arrival rates via pricing. For example, a higher price offered to the customer or a lower price offered to the server reduces their arrival rate. Our objective is to devise a joint dynamic pricing and matching policy to maximize the profit for the system operator and minimize the delay for customers and servers. Specifically, we focus on characterizing the Pareto frontier of regret and average queue length, where regret is defined as the loss in profit compared to the so-called fluid benchmark (see \eqref{equ:regret-def} for a precise definition).

This model was first introduced and studied in \cite{varma2023dynamic}. They proposed a dynamic pricing and matching policy and established a $\Theta(\eta^{1-\gamma})$ regret and $\Theta(\eta^{\gamma/2})$ queue length for $\gamma \in [0, 1]$ in the steady state, where $\eta$ is the system size. They also showed that such a trade-off between regret and queue length is the best possible for a large class of policies. However, the proposed pricing policy of \cite{varma2023dynamic} is based on the optimal solution of the so-called fluid problem which relies on knowing the demand and supply curves. These curves are typically estimated using historical data. Thus, any estimation error leads to an error in the optimal fluid solution, resulting in a $\Theta(\eta)$ regret in the steady state. So, it is essential to simultaneously estimate the demand and supply curves and optimize the pricing policy.

Building on \cite{varma2023dynamic}, the paper \cite{yang2024learning} proposed a learning-based pricing policy that iteratively updates the price based on a zeroth-order stochastic projected gradient ascent on the fluid optimization problem. They establish a $\tilde{\Theta}(T^{1-\gamma})$ regret and $\Theta(T^{\gamma})$ average queue length bounds for $\gamma \in [0, 1/6]$, where $T$ is the time horizon (plays a similar role as $\eta$ in \cite{varma2023dynamic}). So, while this pricing scheme is parameter-agnostic, it is not Pareto optimal as it does not achieve $1-\gamma$ versus $\gamma/2$ tradeoff. One can understand this sub-optimality by observing that their pricing scheme mimics a static policy as $T \rightarrow \infty$ as opposed to a two-price policy as in \cite{varma2023dynamic}.

To this end, we develop a pricing policy that achieves the best of both worlds: a parameter-agnostic pricing scheme that mimics a two-price policy. We establish an improved $\tilde{\Theta}(T^{1-\gamma})$ regret and  $\tilde{\Theta}(T^{\gamma/2})$ average queue length for $\gamma \in [0, 1/6]$. We show that this trade-off,  that is, $1-\gamma$ versus $\gamma/2$ as in \cite{varma2023dynamic}, is optimal up to logarithmic factors under a class of policies.
The only discrepancy is in the permissible value of $\gamma$. In particular, we incur an additional learning error compared to \cite{varma2023dynamic} which prevents us from improving the regret beyond $\tilde{\Theta}(T^{5/6})$. Nonetheless, we significantly improve over \cite{yang2024learning} and achieve a near-optimal trade-off between regret and average queue length. Additionally, our policy also allows us to ensure $\Theta(T^{\gamma})$ maximum queue length.

Our pricing scheme is a novel dynamic probabilistic policy that optimizes the tension between obtaining useful samples for fast learning and maintaining small queue lengths while incurring low regret.
At the time $t \in [T]$, let $p(t)$ be the price prescribed by the learning scheme for a fixed customer type with queue length $Q(t) \in \mathbb{Z}_+$. Then, we are inclined to set a price $p(t)$ to obtain useful samples for learning, but that may increase the queue length. Our policy resolves this difficulty as follows. If $Q(t) = 0$, set the price to be $p(t)$ to prioritize learning when the queue length is empty. Similarly, if $Q(t) \ge q^{\mathrm{th}}$, set the price to be the maximum price to prioritize high negative drift when the queue length is large. For the in-between case of $0 < Q(t) < q^{\mathrm{th}}$, we set the price to be $p(t)$ with probability $1/2$ and $p(t)+\alpha$ otherwise. Such a probabilistic scheme ensures that half of the collected samples is useful in learning. In addition, $\alpha \in \mathbb{R}$ is suitably picked to optimize the trade-off between maintaining low queue lengths and incurring a small regret. In other words, we converge to an optimal two-price policy as $T \rightarrow \infty$, as opposed to a static policy in \cite{yang2024learning} which is known to be sub-optimal.

\subsection{Related Work}

Contrary to the classical literature on online learning \cite{auer2002finite,zinkevich2003online,flaxman2004online,agarwal2010optimal,lattimore2020bandit}, learning in queueing systems poses an additional challenge of strong correlations over time induced by maintaining a queue, in addition to its countable state space invoking the curse of dimensionality. A recent surge of literature \cite{yang2023learning, yang2024learning, Kempen2024Learning, zhong2024learning, Sun2024Inpatient} has emerged to better understand online learning in queueing.

One line of work views the queueing system as a Markov decision process and uses reinforcement learning methods in an attempt to learn a global optimal policy \cite{liu2019reinforcement, murthy2024performance}. These methods are broadly applicable to many types of queueing systems.
Contrary to this, our focused approach allows us to exploit the structural result that a two-price policy is near optimal \cite{varma2023dynamic} to restrict our policy space, resulting in a more efficient and practical learning scheme.

In line with our approach, another line of work exploits the structure of either the model \cite{Weber2024Reinforcement}, the stationary distribution \cite{Comte2023Score}, or the (near) optimal policy \cite{Chen2023Online, zhong2024learning}. Closest to ours is the paper \cite{Chen2023Online} that learns to price for a queueing system with a single server and a single queue, where arrivals follow a Poisson process and service times are independent and identically distributed with a general distribution. However, they assume uniform stability of the queueing system,
circumventing one of the main challenges we tackle in this work. This assumption allows them to learn a static pricing policy and show its optimality \cite{kim2018value}. On the other hand, we focus on learning a dynamic pricing policy and demonstrate its benefits, which we believe is novel in the literature on learning in queues.

\section{Problem Formulation}
\label{sec:model}

\begin{figure}[htb]
    \centering
    \includegraphics[width=0.7\linewidth]{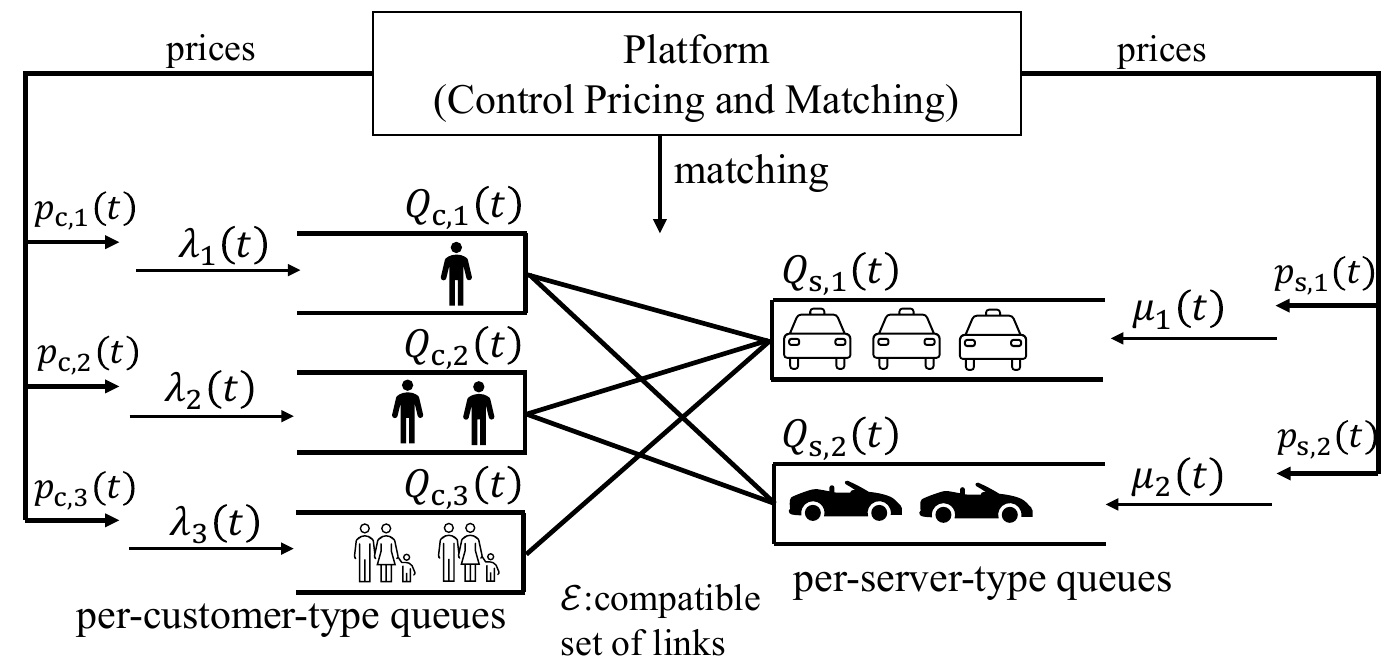}
    \caption{The model, an example with 3 types of customers and 2 types of servers.}
    \label{fig:model}
\end{figure}

We are using the model developed by \cite{yang2024learning}. We present the model in this section for completeness.
We study a discrete-time system that contains two sets (sides) of queues: one for customers and the other for servers. Each side includes multiple queues to represent various types of customers or servers. Customers can be seen as demand, and servers as supply. This system can be modeled using a bipartite graph $G({\cal I} \cup {\cal J}, {\cal E})$, where ${\cal I}=\{1,2,\ldots,I\}$ denotes customer types and ${\cal J}=\{1,2,\ldots,J\}$ denotes server types, with $|{\cal I}|=I$ and $|{\cal J}|=J$. The set ${\cal E}$ includes all compatible links, that is, a type $i$ customer can be served by a type $j$ server if and only if $(i,j)\in {\cal E}$. Figure~\ref{fig:model} illustrates this with three customer types and two server types ($I=3, J=2$).

At the beginning of each time slot $t$, the platform observes $Q_{\mathrm{c}, i}(t)$ and $Q_{\mathrm{s}, j}(t)$ for all $i\in {\cal I}$ and $j\in {\cal J}$, where $Q_{\mathrm{c}, i}(t)$ and $Q_{\mathrm{s}, j}(t)$ represent the queue lengths of type $i$ customers and type $j$ servers, respectively. Assume that all queues are initially empty at $t=1$. 
Then the platform determines the price for each queue, i.e., for each type of customer and server. Let $p_{\mathrm{c}, i}(t)$ represent the price for a type $i$ customer and $p_{\mathrm{s}, j}(t)$ the price for a type $j$ server in time slot $t$. The price $p_{\mathrm{c}, i}(t)$ means that a type $i$ customer will be charged $p_{\mathrm{c}, i}(t)$ units of money by the platform when they join the queue. The price $p_{\mathrm{s}, j}(t)$ means that a type $j$ server will receive $p_{\mathrm{s}, j}(t)$ units of money from the platform when they join the queue. These prices will influence the arrivals of the customers and servers. In time slot $t$, let $A_{\mathrm{c},i}(t)$ represent the number of type $i$ customer arrivals, and $A_{\mathrm{s},j}(t)$ the number of type $j$ server arrivals. Assume $A_{\mathrm{c},i}(t)$ and $A_{\mathrm{s},j}(t)$ are independently distributed according to Bernoulli distributions. Define $\lambda_i(t)\coloneqq \expt[A_{\mathrm{c},i}(t)]$ and $\mu_j(t)\coloneqq \expt[A_{\mathrm{s},j}(t)]$. 
We use a demand function and a supply function to model the relation between the price and the arrival rate for each customer and each server.
For a type $i$ customer, the demand function, $F_i:[0,1] \rightarrow [p_{\mathrm c,i,\min}, p_{\mathrm c,i,\max}]$, models the relation between
the arrival rate $\lambda_i(t) \in [0,1]$ and the price $p_{\mathrm{c}, i}(t) \in [p_{\mathrm c,i,\min}, p_{\mathrm c, i,\max}]$, with $p_{\mathrm c,i,\min}$ and $p_{\mathrm c,i,\max}$ as the minimum and maximum prices, respectively. Similarly, the supply function for a type $j$ server, $G_j: [0,1] \rightarrow [p_{\mathrm s,j,\min}, p_{\mathrm s,j,\max}]$, models the relation between the arrival rate $\mu_j(t) \in [0,1]$ and the price $p_{\mathrm{s}, j}(t) \in [p_{\mathrm s,j,\min}, p_{\mathrm s,j,\max}]$, where $p_{\mathrm s,j,\min}$ and $p_{\mathrm s,j,\max}$ are the minimum and maximum prices for a type $j$ server. We make the following assumption on $F_i$ and $G_j$ as in \cite{yang2024learning}:
\begin{assumption}\label{assum:1}
    For any customer type $i$, $F_i$ is strictly decreasing, bijective, and $L_{F_i}$-Lipschitz. Its inverse function $F_i^{-1}$ is $L_{F_i^{-1}}$-Lipschitz.
    For any server type $j$, $G_j$ is strictly increasing, bijective, and $L_{G_j}$-Lipschitz. Its inverse function $G_j^{-1}$ is $L_{G_j^{-1}}$-Lipschitz.
\end{assumption}
Thus, the price $p_{\mathrm{c}, i}(t)$ determines the arrival rate $\lambda_i(t)$ through $\lambda_i(t) = F_i^{-1} (p_{\mathrm{c}, i}(t))$ for all $i\in{\cal I}$ and the price $p_{\mathrm{s}, j}(t)$ determines the arrival rate $\mu_j(t)$ through $\mu_j(t) = G_j^{-1}(p_{\mathrm{s}, j}(t))$ for all $j\in{\cal J}$.
After arrivals in the time slot $t$, the platform matches customers and servers in the queues via compatible links. Once a customer is matched with a server, both exit the system immediately.

\textbf{Objectives:} Consider a finite time horizon $T$. As in \cite{yang2024learning}, we consider three performance metrics, profit, average queue length, and maximum queue length, as shown as follows:
\begin{align*}
    & \text{Profit} \text{\small $(T)
    = \sum_{t=1}^{T} \expt \left[ \sum_i \lambda_i(t) F_i(\lambda_i(t)) - \sum_j \mu_j(t) G_j(\mu_j(t)) \right]$}\nonumber\\
    & \text{AvgQLen}(T) =  \frac{1}{T}\sum_{t=1}^{T} \expt \biggl[\sum_i Q_{\mathrm{c}, i}(t) + \sum_j Q_{\mathrm{s}, j}(t)\biggr]\nonumber\\
    & \text{MaxQLen}(T) =  \max_{t=1,\ldots,T} \max\Bigl\{\max_i Q_{\mathrm{c}, i}(t), \max_j Q_{\mathrm{s}, j}(t)\Bigr\}.
\end{align*}
Our goal is to design an online pricing and matching algorithms to maximize the profit of the platform \textit{without knowing the demand and supply functions}, while maintaining reasonable maximum and average queue lengths.

\textbf{Baseline:} We compare the profit of an online algorithm with the optimal value of a fluid-based optimization problem~\cite{varma2023dynamic} as follows:
\begin{align}
    \max_{\boldsymbol{\lambda}, \boldsymbol{\mu}, \boldsymbol{x}} & \sum_i \lambda_i F_i(\lambda_i) - \sum_j \mu_j G_j(\mu_j)\label{equ:fluid-opti-obj}\\
    \mathrm{s.t.} \qquad \lambda_i = &  \sum_{j: (i,j)\in {\cal E}} x_{i,j}  \text{  for all } i\in{\cal I}, 
    \quad \mu_j =  \sum_{i: (i,j)\in {\cal E}} x_{i,j} \text{  for all } j\in {\cal J},  \label{equ:fluid-opti-constr2}\\
    x_{i,j} \ge & 0  \text{  for all } (i,j)\in {\cal E},
    \quad    \lambda_i, \mu_j \in  [0,1] \text{  for all } i\in{\cal I}, ~j\in {\cal J},\label{equ:fluid-opti-constr4}
\end{align}
where $\boldsymbol{\lambda}\coloneqq (\lambda_i)_{i\in {\cal I}}$ and $\boldsymbol{\mu}\coloneqq (\mu_j)_{j\in {\cal J}}$ can be viewed as steady-state arrival rates,
and $\boldsymbol{x}\coloneqq (x_{i,j})_{(i,j)\in {\cal E}}$ can be viewed as steady-state matching rates. The constraints \eqref{equ:fluid-opti-constr2} can be viewed as balance equations.
We adopt the following assumptions as in~\cite{yang2024learning,varma2023dynamic}:
\begin{assumption}\label{assum:2}
    The function $\lambda_i F_i(\lambda_i)$ is concave in $\lambda_i$ for all $i$ and the function $\mu_j G_j(\mu_j)$ is convex in $\mu_j$ for all $j$.
\end{assumption}
\begin{assumption}\label{assum:3}
    There exists a known positive number $a_{\min}\in (0,1)$ for which there exists an optimal solution $(\boldsymbol{\lambda}^*, \boldsymbol{\mu}^*, \boldsymbol{x}^*)$ to the optimization problem \eqref{equ:fluid-opti-obj}-\eqref{equ:fluid-opti-constr4}, satisfying $x_{i,j}^* > 0$, $a_{\min} \le \lambda_i^* < 1$ and $a_{\min} \le \mu_j^* < 1$ for all $i$ and $j$.
\end{assumption}
Assumption~\ref{assum:2} implies that the profit function $\lambda_i F_i(\lambda_i) - \mu_j G_j(\mu_j)$ is concave and the optimization problem defined by \eqref{equ:fluid-opti-obj}-\eqref{equ:fluid-opti-constr4} is concave. This concavity assumption follows from the economic law of diminishing marginal return: as the arrival rate increases, the marginal return, that
is, the derivative of the profit function, decreases, which implies that the profit function is concave.
Assumption~\ref{assum:3} means that there exists an optimal solution that resides in the interior of the feasible set. This means that no links or queues are redundant and no queues reach maximum arrival rates in this optimal solution.
This is a mild assumption because:
\begin{itemize}[leftmargin=*]
    \item If any redundant links or queues exist, they can likely be easily identified in practice and removed from the problem.
    \item Suppose each time slot is set to be sufficiently small when modeling real-world arrival rates. Then, an arrival rate of $1$ in this discrete-time model with Bernoulli arrivals corresponds to a sufficiently large actual arrival rate. In this case, no queue would reach the maximum arrival rate in the optimal solution due to the concavity of the profit function w.r.t. arrival rates.
\end{itemize}

We consider policies that make the queue mean rate stable~\cite{neely2022stochastic}, i.e., under the policy, for all $i,j$,
\begin{align}\label{equ:mean-rate-stable}
    \lim_{T\rightarrow \infty}\frac{1}{T} \expt [Q_{\mathrm{c},i}(T)] = 0,
    \qquad \lim_{T\rightarrow \infty}\frac{1}{T} \expt [Q_{\mathrm{s},j}(T)] = 0.
\end{align}
As shown in \cite{yang2024learning}, the optimal value of the fluid optimization problem \eqref{equ:fluid-opti-obj}-\eqref{equ:fluid-opti-constr4} is an upper bound on the asymptotic time-averaged expected profit under any mean rate stable policy.
Therefore, we use the optimal value of the fluid problem as baseline to define expected regret:
\begin{align}\label{equ:regret-def}
    \expt [R(T)] \coloneqq  
     T\biggl(\sum_i \lambda^*_i F_i(\lambda^*_i) - \sum_j \mu^*_j G_j(\mu^*_j) \biggr) 
    -   \sum_{t=1}^{T} \expt \left[  \sum_i \lambda_i(t) F_i(\lambda_i(t)) - \sum_j \mu_j(t) G_j(\mu_j(t))  \right].
\end{align}
\textbf{Challenges: } We face several challenges in addressing this problem:
\begin{itemize}[leftmargin=*]
    \item The first challenge is the \textit{unknown} demand and supply functions $F_i$ and $G_j$. The objective and constraints of the optimization problem~\eqref{equ:fluid-opti-obj}-\eqref{equ:fluid-opti-constr4} are in terms of arrival rates but the platform can only control the prices directly. If we choose to use prices as the control variable of the optimization problem, then the problem becomes nonconcave due to nonlinear equations in the constraint set, and the constraint set becomes unknown. This is the key challenge compared to the case of the single-sided queue~\cite{Chen2023Online}, where balance equations are not needed.  
    \item We are considering an online learning setting where learning and decision making occur simultaneously. It is not acceptable to first estimate the demand and supply functions through running the system with difference prices (e.g., nonparametric regression) and then solve the optimization problem, because the queue lengths can grow linearly over time in the estimation process due to imbalanced arrival rates. An example can be found in Appendix~\ref{app:example-predict-then-optimize}.
    \item  It is challenging to characterize and achieve the optimal tradeoff between regret and queue length. The paper \cite{yang2024learning} obtain $\tilde{\Theta}(T^{1-\gamma})$ regret and $\tilde{\Theta}(T^{\gamma})$ average queue length bounds, but we know from \cite{varma2023dynamic} that this tradeoff is not optimal. It is worth studying whether there exists an algorithm that achieves a near-optimal tradeoff between regret and average queue length in this learning setting.
\end{itemize}

\textbf{Definition of Pareto-optimal tradeoff:}
Let $\mathbb{E}_{\pi,d}[R(T)]$ denote the expected profit regret under policy $\pi$ and problem instance $d$. The problem instance $d$ specifies the bipartite graph $G({\cal I} \cup {\cal J}, {\cal E})$ and the demand and supply functions. Let $\mathrm{AvgQLen}_{\pi,d}(T)$ denote the average queue length under policy $\pi$ and problem instance $d$. 
Let $(x(\pi),y(\pi))$ denote the exponents of the objective values under the worst case, i.e., 
\[
x(\pi)\coloneqq \limsup_{T\rightarrow\infty} \log_T (\sup_d \mathbb{E}_{\pi,d}[R(T)])
\]
and 
\[
y(\pi)\coloneqq \limsup_{T\rightarrow\infty} \log_T (\sup_d \mathrm{AvgQLen}_{\pi,d}(T)).
\]
A point $(x(\pi_1),y(\pi_1))$ is said to be better than another point $(x(\pi_2),y(\pi_2))$ if $x(\pi_1) \le x(\pi_2)$ and $y(\pi_1)\le y(\pi_2)$
with at least one inequality strict.
A point $(x(\pi),y(\pi))$ is Pareto-optimal if there is no other policy $\pi'$ that satisfies $x(\pi') \le x(\pi)$ and $y(\pi')\le y(\pi)$
with at least one inequality strict. The set of all Pareto-optimal points forms the Pareto frontier.

\textbf{Notation:} Bold symbols denote vectors in $\mathbb{R}^{|{\cal E}|}$, $\mathbb{R}^{I}$ or $\mathbb{R}^{J}$, such as $\boldsymbol{x}, \boldsymbol{\lambda}, \boldsymbol{\mu}$.
Subscript $i$ or $j$ refer to individual elements of a vector; for example, $\lambda_i$ represents the $i^{\mathrm{th}}$ element of $\boldsymbol{\lambda}$. Note that $\boldsymbol{x}= (x_{i,j})_{(i,j)\in {\cal E}}$ is treated as a vector in $\mathbb{R}^{|{\cal E}|}$. The order of elements in $\boldsymbol{x}$ can be arbitrary, provided it is consistent. Additionally, we use the subscript $\mathrm{c}$ for the customer side and $\mathrm{s}$ for the server side.

\section{Main Results}
\label{sec:overview-results}
In this section, we provide an overview of the main results in this paper. We use a longest-queue-first matching algorithm and propose a novel learning-based pricing algorithm, which will be described in Section~\ref{sec:alg} in detail. The new algorithm achieves the following profit-regret bound and queue-length bounds:
\begin{align*}
    \expt[R(T)] = \tilde{O}\Bigl(T^{1 - \gamma}\Bigr),
    \quad \text{AvgQLen}(T) = \tilde{O}(T^{\frac{\gamma}{2}}),
    \quad \text{MaxQLen}(T) = O\Bigl(T^{\gamma}\Bigr),
\end{align*}
for any $\gamma\in[0, 1/6]$, where $\gamma$ is a hyperparameter in the proposed algorithm. 
The formal statements of the results can be found in Appendix~\ref{app:formal-theorems-analysis}.
As the allowable queue length increases, the achievable regret bound improves. However, the regret bound cannot be further reduced below $\tilde{O}(T^{5/6})$ in our result. Compared to \cite{yang2024learning}, the average queue length in this tradeoff is significantly improved from $\tilde{O}(T^{\gamma})$ to $\tilde{O}(T^{\gamma / 2})$ while the regret and the maximum queue length remain the same.

\textbf{Near-optimal tradeoff:}
This $\tilde{O}(T^{1 - \gamma})$ versus $\tilde{O}(T^{\gamma / 2})$ tradeoff between regret and average queue length matches the $\Theta(\eta^{1-\gamma})$ regret and $\Theta(\eta^{\gamma/2})$ queue length tradeoff in \cite{varma2023dynamic} up to logarithmic factors, where they show the tradeoff is optimal in their setting. The parameter $\eta$ in \cite{varma2023dynamic} is the scaling of arrival rates in their setting, which plays a similar role as the time horizon $T$ in our setting because both characterize the scaling of the number of arrivals to the system. We now establish that this trade-off between regret and average queue length is nearly optimal. We restrict ourselves to a single-link system, i.e., $I=J=1$ and drop the subscripts $i$ and $j$.
Consider the matching policy that matches all pairs whenever possible. Note that in a single-link system, there is no incentive to delay any possible match, as doing so only results in increased queue lengths.
Similar to $\text{AvgQLen}(T)$, define $\text{AvgQ$^2$Len}(T) \coloneqq \frac{1}{T} \sum_{t=1}^T \expt[Q_{\mathrm{c}}(t)^2+Q_{\mathrm{s}}(t)^2]$. Assuming an upper bound on $\text{AvgQ$^2$Len}(T)$, the lemma below obtains a lower bound on the regret. 

\begin{lemma}\label{lemma:lower-bound}
    Let Assumption~\ref{assum:1} and Assumption~\ref{assum:2} hold. Assume that the functions $-xF(x)$ (negative revenue) and $xG(x)$ (cost) are both strongly convex. Also assume the unique optimal solution to the fluid optimization problem $\lambda^*=\mu^* \notin \{0,1\}$ to avoid the trivial case.

    Fix a $\gamma \in [0, 1/2]$. Then, for sufficiently large $T$, for any pricing policy for which $\sqrt{\text{AvgQ$^2$Len}(T)} \le T^{\gamma/2}$ and $\expt [Q_{\mathrm{c}}(T+1)^2+Q_{\mathrm{s}}(T+1)^2] = o(T)$, we must have $
    \expt[R(T)] = \Omega (T^{1-\gamma})
    $.
\end{lemma}
Proof of Lemma~\ref{lemma:lower-bound} can be found in Appendix~\ref{app:proof-lemma-lower-bound}. Note that the above lemma assumes $\sqrt{\text{AvgQ$^2$Len}(T)}$ $\leq T^{\gamma/2}$, which is slightly stronger than assuming $\text{AvgQLen}(T) \leq T^{\gamma/2}$ as $\sqrt{\text{AvgQ$^2$Len}(T)} \geq \text{AvgQLen}(T)$ by Jensen's inequality and the fact that $Q_{\mathrm{c}}(t)^2+Q_{\mathrm{s}}(t)^2 = (Q_{\mathrm{c}}(t) + Q_{\mathrm{s}}(t))^2$ (by Lemma 1 in \cite{yang2024learning}). In particular, our proof strategy requires a handle on a higher moment of the queue length to obtain the required bound on the regret. In Appendix~\ref{app:proof-two-price-satisfy-cond}, we show that $\sqrt{\text{AvgQ$^2$Len}(T)} = \Theta(\text{AvgQLen}(T))$ for a large class of pricing policies that are a small perturbation of a static pricing policy satisfying a natural negative drift condition. Such a class of policies is ubiquitous in the literature \cite{kim2018value, varma2023dynamic, varma_ht} on dynamic pricing in queues.

To understand the $1-\gamma$ versus $\gamma/2$ trade-off, consider a general pricing policy given by $\lambda(Q_{\mathrm{c}}, Q_{\mathrm{s}})$, $\mu(Q_{\mathrm{c}}, Q_{\mathrm{s}}) \in [\lambda^\star-\alpha, \lambda^\star+\alpha]$ for some $\alpha > 0$. For any such policy, the average queue length is at least $\Theta(1/\alpha)$ as the difference in the arrival rates is $\Theta(\alpha)$ \cite{varma_ht}. Also, denote the regret as a function of the arrival rate for each time slot by $r$ and note that $r(\lambda^\star) = 0$. By Taylor's expansion, the cumulative regret in time $T$ for arrival rate $\lambda^\star \pm \Theta(\alpha)$ is $T\Theta(\alpha r'(\lambda^\star) + \alpha^2 r''(\lambda^\star)/2)$, where the first order term vanishes as $r'(\lambda^\star) \approx 0$ by the first order optimality conditions. Thus, in summary, the queue length is of the order $1/\alpha$ and the regret is $T\alpha^2$. By substituting $\alpha = T^{-\gamma/2}$, we obtain that the $1-\gamma$ versus $\gamma/2$ trade-off is fundamental.

\textbf{Comparison with \cite{yang2024learning}:} The paper \cite{yang2024learning} employs a pricing policy with an admission control threshold $q^{\mathrm{th}}$. In such a setting, the queue length is of the order $\Theta(q^{\mathrm{th}})$ as the underlying system dynamics is a symmetric random walk between $[0, q^{\mathrm{th}}]$. In addition, the regret is equal to $\sum_{t=1}^{T}\Pr[Q(t)=q^{\mathrm{th}}]$, i.e., the proportion of the time we hit the boundary, which is approximately equal to $T/q^{\mathrm{th}}$. Thus, by picking $q^{\mathrm{th}}=T^{\gamma}$, it will result in a $1-\gamma$ versus $\gamma$ trade-off. 
\emph{This intuition shows that the approach of \cite{yang2024learning} will not be able to result in the optimal $1-\gamma$ versus $\gamma/2$ trade-off. }

\textbf{Scaling with the number of customer/server types:}
The regret upper bound scales with the number of customer types $I$ and server types $J$ as $\tilde{O}(I^4J^4(I+J)T^{1-\gamma})$. 
The average queue length upper bound scales with $I$ and $J$ as $\tilde{O}(IJ(I+J)T^{\gamma/2})$.
The maximum queue length does not scale with $I$ or $J$. We believe the dependencies on $I$ and $J$ can be improved, as our analysis mainly focuses on obtaining the correct dependence in terms of the time horizon $T$.

\section{Algorithm}
\label{sec:alg}

In this section, we will describe the matching algorithm we use and propose a novel pricing algorithm, under which we can achieve the results in the previous section.

\textbf{Matching algorithm:} We use the same matching algorithm proposed in \cite{yang2024learning}, which is a discretized version of the MaxWeight algorithm in \cite{varma2023dynamic}.
At each time slot, we iterate through all queues on both the customer and server sides. For each queue, we first check for new arrivals. If a new arrival is detected (either a customer or a server), the algorithm matches it with an entity from the longest compatible queue on the opposite side.

\textbf{Pricing algorithm:} We propose a novel pricing algorithm that aims to achieve the optimal tradeoff between regret and average queue length, while having a guarantee of anytime maximum queue length. Same as \cite{yang2024learning}, we use the techniques of zero-order stochastic projected gradient ascent~\cite{agarwal2010optimal} and bisection search in the pricing algorithm. The key innovation lies in how we implement the $\alpha$-perturbed two-price policy mentioned in the intuition in Section~\ref{sec:overview-results} in the \emph{learning} setting.
We propose a novel \textit{probabilistic two-price policy}, where we reduce the arrival rate by perturbing the price with a constant probability for each queue when the queue is nonempty and the length is less than a predefined threshold. This \textit{probabilistic two-price policy} achieves a near-optimal tradeoff between regret and average queue length, while having a guarantee of anytime maximum queue length.
The rest of this section provides a detailed explanation of our pricing algorithm.

\subsection{Two-Point Zero-Order Method}
Note that under Assumption~\ref{assum:3}, the fluid problem \eqref{equ:fluid-opti-obj}-\eqref{equ:fluid-opti-constr4} can be equivalently rewritten as
\begin{align}
    \max_{\boldsymbol{x}} f(\boldsymbol{x}) \coloneqq & \sum_i  \Biggl(\sum_{j\in {\cal E}_{\mathrm{c},i}} x_{i,j} \Biggr) F_i\Biggl(\sum_{j\in {\cal E}_{\mathrm{c},i}} x_{i,j}\Biggr) 
    - \sum_j \Biggl(\sum_{i\in {\cal E}_{\mathrm{s},j}} x_{i,j}\Biggr) G_j\Biggl(\sum_{i\in {\cal E}_{\mathrm{s},j}} x_{i,j}\Biggr) \label{equ:eqv-opti-obj}\\
    \mathrm{s.t.}  \sum_{j\in {\cal E}_{\mathrm{c},i}} x_{i,j} \in & [a_{\min}, 1] \text{  for all } i\in{\cal I},
    \quad 
    \sum_{i\in {\cal E}_{\mathrm{s},j}} x_{i,j} \in [a_{\min}, 1] \text{  for all } j\in{\cal J}, \label{equ:eqv-opti-constr2}\\
    x_{i,j} \ge & 0  \text{  for all } (i,j)\in {\cal E},\label{equ:eqv-opti-constr3}
\end{align}
where ${\cal E}_{\mathrm{c},i}\coloneqq \{j | (i,j)\in {\cal E}\}$ and ${\cal E}_{\mathrm{s},j}\coloneqq \{i | (i,j)\in {\cal E}\}$.
Given that the demand and supply functions $F_i$ and $G_j$ are unknown, we cannot solve the problem \eqref{equ:eqv-opti-obj}-\eqref{equ:eqv-opti-constr3} directly.
Suppose we have zero-order access to the objective function \eqref{equ:eqv-opti-obj}, i.e., the value of the objective function can be evaluated given any input variable $\boldsymbol{x}$. Then, we can use the two-point zero-order method~\cite{agarwal2010optimal}. We will present the main idea here. In the method, we begin with an initial feasible solution and perform projected gradient ascent, estimating the gradient at each iteration using two points that are perturbed from the current solution. Specifically, it can be described in the following steps:

\textbf{(1) Initial feasible point:} 
Let ${\cal D}$ denote the feasible set of the problem \eqref{equ:eqv-opti-obj}-\eqref{equ:eqv-opti-constr3}. Let ${\cal D}'$ denote a shrunk set of ${\cal D}$ with a parameter $\delta>0$. The word ``shrunk'' means that $\boldsymbol{x} + \delta \boldsymbol{u} \in {\cal D}$ for any $\boldsymbol{x}\in {\cal D}'$ and any vector $\boldsymbol{u}$ in the unit ball, which was proved in \cite{yang2024learning} with the definition of ${\cal D}'$ under some mild assumptions. The details about ${\cal D}'$ are presented in Appendix~\ref{app:def-shrunk-set} for completeness. The algorithm begins with an initial feasible solution $x(1)\in{\cal D}'$.

\textbf{(2) Generate a random direction and two points: }
In each iteration $k$, we begin by sampling a unit vector $\boldsymbol{u}(k)$ in a uniformly random direction. Next, we perturb the current solution $\boldsymbol{x}(k)$ in the direction of $\boldsymbol{u}(k)$ and the opposite direction of $\boldsymbol{u}(k)$, generating two points, $\boldsymbol{x}(k)+ \delta \boldsymbol{u}(k)$ and $\boldsymbol{x}(k)- \delta \boldsymbol{u}(k)$. We know that if the current solution $\boldsymbol{x}(k)$ is in the shrunk set ${\cal D}'$, then the two points will be in the feasible set ${\cal D}$ and hence are feasible.

\textbf{(3) Estimate the function values in the two points: } If we have zero-order access of the objective function $f$, we can obtain the function values $f(\boldsymbol{x}(k)+ \delta \boldsymbol{u}(k))$ and $f(\boldsymbol{x}(k) - \delta \boldsymbol{u}(k))$. However, we do not have zero-order access to $f$ since we do not even have zero-order access to the demand and supply functions $F_i$ and $G_j$. In this case, we need to estimate the values of $F_i$ and $G_j$ given some known arrival rates as inputs. We will use a bisection search method similar to \cite{yang2024learning} to estimate the values, which will be described later in Section~\ref{sec:bisection}. 

\textbf{(4) Gradient Calculation: }
With the estimation of the two function values $\hat{f}(\boldsymbol{x}(k)+ \delta \boldsymbol{u}(k))$ and $\hat{f}(\boldsymbol{x}(k) - \delta \boldsymbol{u}(k))$, the gradient
$
    \hat{\boldsymbol{g}}(k) = \frac{|{\cal E}|}{2\delta} [\hat{f}(\boldsymbol{x}(k)+ \delta \boldsymbol{u}(k)) - \hat{f}(\boldsymbol{x}(k) - \delta \boldsymbol{u}(k)) ] \boldsymbol{u}(k).
$

\textbf{(5) Projected gradient ascent: }
With the estimated gradient $\hat{\boldsymbol{g}}(k)$, we perform a step of projected gradient ascent with a step size $\eta$:
$
    \boldsymbol{x}(k+1) = \Pi_{{\cal D}'}(\boldsymbol{x}(k)+\eta \hat{\boldsymbol{g}}(k) ).
$

Then the algorithm repeats Step (2)-(5) for the next iteration. This part of the pricing algorithm is the same as the balanced pricing algorithm in \cite{yang2024learning} and the details of the algorithm can be found in Appendix~\ref{app:alg-pricing} for completeness.

\subsection{Bisection Search Method}
\label{sec:bisection}

In this section, we will illustrate the bisection search method that we mentioned in the previous section. The bisection search takes the target arrival rates and the price searching intervals for all queues as inputs and outputs the estimated prices corresponding to the input arrival rates with accuracy $\epsilon$.
The bisection search can be illustrated in the following steps:

\textbf{(1) Calculate the midpoints:} In each bisection iteration $m$ ($m=1,2,\ldots,M$, $M=\lceil \log_2 (1/\epsilon)\rceil$), we first calculate the midpoint of the price searching interval for each queue. Note that the initial price searching intervals when $k=1$ and $m=1$ are predetermined at the beginning of the pricing algorithm, and the initial price searching intervals when $k>1$ and $m=1$ are calculated from the output of the bisection search in the previous outer iteration $k-1$, which can be found in Line~\ref{line:alg-pricing-binary-start}-\ref{line:alg-pricing-set-interval-end} in Algorithm~\ref{alg:pricing}. The underlying idea is that the price will not change too much since the corresponding target arrival rate will not change too much due to the small step size $\eta$.

\textbf{(2) Run the system to collect samples:} Next, we run the system with the midpoints for a certain number of time slots to obtain $N$ samples of each arrival rate corresponding to each midpoint, where $N=\lceil(\beta/\epsilon^2)\ln (1/\epsilon)\rceil$ and $\beta>0$ is a constant. How we run the system to collect samples is the key difference between our proposed algorithm and that of \cite{yang2024learning}. We propose a \textit{probabilistic two-price policy} and will present it later in Section~\ref{sec:alg-rand-two-price}. 

\textbf{(3) Estimate the arrival rates and update the price searching intervals: } Then, we can obtain an estimate of the arrival rate for each queue by taking the average of $N$ samples.
For each queue on the customer side, if the estimated arrival rate is greater than the input arrival rate, we should increase the price to reduce the arrival rate because the demand function is decreasing, so the price searching interval should be updated to the upper half of the previous price searching interval. Otherwise, the price searching interval should be updated to the lower half of the previous price searching interval. For each queue on the server side, we should do the opposite since the supply function is increasing.

Then the algorithm repeats Step (1)-(3) for the next bisection iteration. The details of the algorithm can be found in Appendix~\ref{app:bisection}.

\subsection{Probabilistic Two-Price Policy}
\label{sec:alg-rand-two-price}

In this section, we propose a novel pricing policy called \textit{probabilistic two-price policy} for Step (2) in Section~\ref{sec:bisection}.
Recall that the goal of Step (2) is to collect samples to estimate the arrival rates corresponding to the prices of the midpoints. For an estimate of accuracy $\epsilon$, we need $N=\tilde{\Theta}(1/\epsilon^2)$ number of samples according to the central limit theorem.

\textbf{Challenges:}
The key challenge in setting prices in this sample collection process is that we need to consider and balance three objectives. First, we need to set prices to be the midpoints to collect samples for learning. Second, we need to control the average queue length and maximum queue length during this sample collection process. Third, we need to minimize the regret.
For fast learning, we want to set the prices to be the midpoints as frequently as possible. However, simply using the midpoints to run the system for $N$ time slots will not give us any way to control the queue lengths. To control the maximum queue length, we use a threshold $q^{\mathrm{th}}$ as in \cite{yang2024learning} -- when the queue length exceeds $q^{\mathrm{th}}$, we reject arrivals by setting the highest price $p_{\mathrm{c},i,\max}$ for customer queue $i$ or the lowest price $p_{\mathrm{s},j,\min}$ for server queue $j$. However, this threshold policy cannot achieve an optimal tradeoff between \textit{average queue length} and regret. 
A challenging question is whether we can achieve a near-optimal tradeoff.

We propose the \textit{probabilistic two-price policy}, which resolves the tension between obtaining useful samples for fast learning and maintaining
small queue lengths, providing a positive answer to the above question.
The pseudo-code of the proposed policy can be found in Algorithm~\ref{alg:rand-two-price} in Appendix~\ref{app:alg-prob-two-price}. The proposed policy can be explained in the following steps:
\begin{itemize}[leftmargin=16pt]
    \item[(1)] At the beginning of the algorithm, we initialize counters to track the number of useful arrival rate samples for each queue, as shown in Line~\ref{line:alg-rand-two-price-counters} in Algorithm~\ref{alg:rand-two-price}.
    \item[(2)] Control the maximum queue length: As in \cite{yang2024learning}, for each queue, we use a threshold $q^{\mathrm{th}}$ to control the maximum queue length.
    If the queue length exceeds or equals the threshold, we will set the price to the maximum (for customer queues) or the minimum (for server queues) to reject new arrivals, as shown in Line~\ref{line:alg-rand-two-price-maximum-customer} and Line~\ref{line:alg-rand-two-price-maximum-server} in Algorithm~\ref{alg:rand-two-price}.
    \item[(3)] \textbf{Control the average queue length with a probabilistic approach:} To control the average queue length, we propose a probabilistic approach of adjusting the price, to reduce the arrival rate for each queue when the queue is nonempty and the length is less than the threshold $q^{\mathrm{th}}$. Specifically, for each customer queue, with a constant probability (set to $1/2$ for simplicity), we increase the price by a small amount $\alpha$ to reduce the arrival rate; for each server queue, with a constant probability, we decrease the price by $\alpha$ to reduce the arrival rate, as shown in Line~\ref{line:alg-rand-two-price-average-customer} and Line~\ref{line:alg-rand-two-price-average-server} in Algorithm~\ref{alg:rand-two-price}. When the queue is empty, we keep the original prices of the midpoints, as shown in Line~\ref{line:alg-rand-midpt-customer} and Line~\ref{line:alg-rand-midpt-server}.
    \item[(4)] With the above prices, we run the system for one time slot. For each queue, we keep the arrival sample only when the original price is used, discarding the sample when the price is adjusted, as shown in Line~\ref{line:alg-rand-useful-customer} and Line~\ref{line:alg-rand-useful-server} in Algorithm~\ref{alg:rand-two-price}. We repeat Step (2) and Step (3) and run the system until the number of samples we keep is no less than $N$ for every queue. Finally, the algorithm returns the samples we keep.
\end{itemize}
\textbf{Difficulties introduced by learning and algorithmic innovations:}
The paper \cite{varma2023dynamic} proposed a two-price policy (the $\alpha$-perturbed policy mentioned in Section~\ref{sec:overview-results}) -- when the queue length exceeds a predetermined threshold, they reduce the arrival rate by a small amount $\alpha$. 
The idea was to ensure that we always have an $\alpha$ negative drift that reduces the queue lengths in expectation. However, in the learning setting where the demand and supply functions are unknown, the objective of setting prices is not only to induce a negative drift to control queue lengths, but also to yield useful samples for learning.
Note that every time we adjust the price by $\alpha$ to induce a negative drift, these arrival samples become biased. These biases influence the value estimates of the two points, which in turn affect the gradient estimates in the two-point gradient ascent method, ultimately impacting the regret.
Using these biased samples will lead to a regret of at least order $\Theta(T\alpha)$. To achieve the optimal trade-off between average queue length and regret, the regret must be $\Theta(T\alpha^2)$ as mentioned in the intuition in Section~\ref{sec:overview-results}. Therefore, in our policy, we discard these biased samples, making them unusable for learning.
Hence, it is important to ensure that negative drift is enforced while not wasting too many samples.
The novel probabilistic approach enables us to obtain useful samples with a constant fraction of time for fast learning, while also ensuring negative drift through reducing arrival rates by $\Theta(\alpha)$ with a constant probability when the queue is nonempty . By setting $\alpha$ we are able to achieve a near-optimal tradeoff between average queue length and regret.

\textbf{Technical innovations:} The analysis related to the proposed probabilistic two-price policy is novel.
Specifically, the new technical challenges and the corresponding innovations include:
\begin{itemize}[leftmargin=*]
    \item One challenge in the analysis is to bound the number of useless samples that are collected at the time of adjusting prices to reduce the arrival rate for queue-length control. The proposed probabilistic approach controls the fraction of time spent adjusting prices, thereby bounding the number of useless samples. This can be formally established using Wald's lemma, as shown in Appendix~\ref{app:sec:bound-num-slots-bisection}.

    \item Another challenge in the analysis is to bound the regret induced by adjusting prices by $\alpha$. The regret induced by adjusting prices can be bounded by the first-order error and the second-order error. The difficulty is to prove that the first-order error is no greater than the second-order error in the learning setting. To show this, we combine the KKT condition (Appendix~\ref{sec:bound-regret-reducing-arr}) and the Lyapunov drift method (Lemma~\ref{lemma:avg-rate-balanced}, proof in Appendix~\ref{app:proof-lemma-avg-rate-balanced}).
\end{itemize}

\subsection{Computational Complexity}
\label{sec:comp-compl}

The proposed pricing algorithm has computational complexity $\tilde{O}((I+J)T + IJT^{1-4\gamma})$, along with $T^{1-4\gamma}$ calls to a projection oracle. The projection can be implemented by solving a convex quadratic program using interior point methods, which have a computational complexity of $\tilde{O}((IJ)^{3.5})$ \cite{nesterov1994interior}. Hence, the overall computational complexity is $\tilde{O}((I+J)T + (IJ)^{3.5} T^{1-4\gamma})$.

\section{Numerical Results}
\label{sec:simu}

In this section, we present simulation results for the proposed algorithm and compare its performance with the \textit{two-price policy} (with known demand and supply functions) from \cite{varma2023dynamic} and the \textit{threshold policy} from \cite{yang2024learning}.

To compare the performance of these algorithms, we consider the following objective function:
\begin{align}\label{equ:simu-obj}
    \sum_{\tau=1}^{t}\biggl( \sum_i \lambda_i(\tau)\left(p_{\mathrm{c},i}(\tau) -w\expt[W_{\mathrm{c},i}(\tau)]\right)
    - \sum_j \mu_j(\tau)\left(p_{\mathrm{s},j}(\tau) + w\expt[W_{\mathrm{s},j}(\tau)]\right)\biggr),
\end{align}
where $W_{\mathrm{c},i}(\tau)$ denotes the waiting time of the customer arriving at queue $i$ at time slot $\tau$ and $W_{\mathrm{s},j}(\tau)$ denotes the waiting time of the server arriving at queue $j$ at time slot $\tau$. The constant $w>0$ can be viewed as the compensation the platform pays for per unit time the customer/server wait.
This objective can be shown to be equivalent to maximizing the originally defined profit minus the sum of the queue lengths multiplied by $w$ (see Appendix~\ref{app:simu}). Therefore, the regret of the objective \eqref{equ:simu-obj} can be calculated by $\expt[R(t)] + w t \text{AvgQLen}(t)$.

\begin{figure}[htbp]
    \centering
    \subfloat[\texorpdfstring{$\expt[R(t)] + w t \text{AvgQLen}(t)$}{E[R(t)] + w t AvgQLen(t)}.]{
    \includegraphics[width=0.44\linewidth]{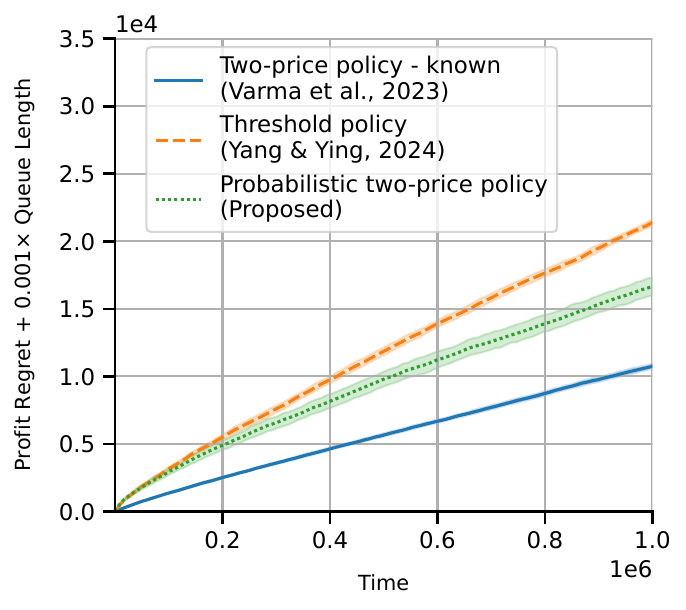}
    }
    \subfloat[Percentage improvement over the \emph{threshold policy} \cite{yang2024learning}.]{
    \includegraphics[width=0.53\linewidth]{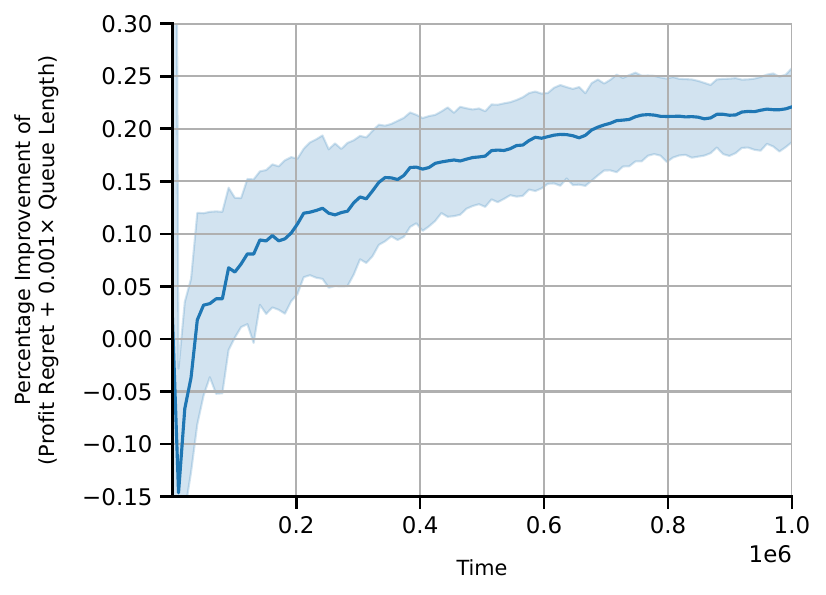}
    }
    \caption{Comparison among the \emph{two-price policy} (no learning, known demand and supply functions) \cite{varma2023dynamic}, the \emph{threshold policy} \cite{yang2024learning}, and the proposed \emph{probabilistic two-price policy} with $w=0.001$. The shaded area is $95\%$ confidence interval with 10 independent runs.}
    \label{fig:comp-0.001}
\end{figure}

Figure~\ref{fig:comp-0.001} presents the performance of the algorithms in a single-link system with $w=0.001$ and $\gamma=1/6$.
We can see that the proposed algorithm significantly improves upon the algorithm in \cite{yang2024learning} by up to $22\%$. The \textit{two-price policy} from \cite{varma2023dynamic} outperforms the other two algorithms because it is a genie-aided policy that assumes full knowledge of the true demand and supply functions, thereby bypassing the challenges associated with learning.

Details of the simulation are provided in Appendix~\ref{app:simu}, where we report the regret, average queue length, and maximum queue length of these algorithms, and also compare their performance under a different weight value $w=0.01$. We also test sensitivity of the performance of the proposed algorithm to parameter tuning. We conduct simulation for a multi-link system as well.

\textbf{Comparison with recent methods for queueing control:} Most of the work on reinforcement learning for queueing in the literature \cite{raeis2021queue,dai2022queueing,liu2022rl,chen2024qgym,murthy2024performance} does not utilize the special structure of our pricing and matching problem, and it is difficult to trade off between profit and queue length. When facing the curse of dimensionality of queueing systems, the works \cite{raeis2021queue,dai2022queueing,chen2024qgym} use deep reinforcement learning, which does not have any theoretical guarantee on the performance. On the other hand, our approach circumvents the curse of dimensionality by restricting to the set of probabilistic two-price policies as opposed to a fully dynamic policy, as we know that a two-price policy is near-optimal.
Other adaptive optimization approaches, such as those that discretize the price space and apply UCB-based methods, fail to balance matching rates and arrival rates on both sides, making it difficult to control queue lengths.


\clearpage
\newpage
\bibliographystyle{unsrt}
\bibliography{refs}

\begin{thebibliography}{10}

\bibitem{yang2024learning}
Zixian Yang and Lei Ying.
\newblock Learning-based pricing and matching for two-sided queues.
\newblock {\em arXiv preprint arXiv:2403.11093}, 2024.

\bibitem{varma2023dynamic}
Sushil~Mahavir Varma, Pornpawee Bumpensanti, Siva~Theja Maguluri, and He~Wang.
\newblock Dynamic pricing and matching for two-sided queues.
\newblock {\em Operations Research}, 71(1):83--100, 2023.

\bibitem{gurvich2015dynamic}
Itai Gurvich and Amy Ward.
\newblock On the dynamic control of matching queues.
\newblock {\em Stochastic Systems}, 4(2):479--523, 2015.

\bibitem{varma2021throughput}
Sushil~Mahavir Varma and Siva~Theja Maguluri.
\newblock Throughput optimal routing in blockchain-based payment systems.
\newblock {\em IEEE Transactions on Control of Network Systems}, 8(4):1859--1868, 2021.

\bibitem{zubeldia_quantum}
Martin Zubeldia, Prakirt~R. Jhunjhunwala, and Siva~Theja Maguluri.
\newblock Matching queues with abandonments in quantum switches: Stability and throughput analysis, 2022.

\bibitem{auer2002finite}
P~Auer.
\newblock Finite-time analysis of the multiarmed bandit problem, 2002.

\bibitem{zinkevich2003online}
Martin Zinkevich.
\newblock Online convex programming and generalized infinitesimal gradient ascent.
\newblock In {\em Proceedings of the 20th international conference on machine learning (ICML-03)}, pages 928--936, 2003.

\bibitem{flaxman2004online}
Abraham~D Flaxman, Adam~Tauman Kalai, and H~Brendan McMahan.
\newblock Online convex optimization in the bandit setting: gradient descent without a gradient.
\newblock {\em arXiv preprint cs/0408007}, 2004.

\bibitem{agarwal2010optimal}
Alekh Agarwal, Ofer Dekel, and Lin Xiao.
\newblock Optimal algorithms for online convex optimization with multi-point bandit feedback.
\newblock In {\em Colt}, pages 28--40. Citeseer, 2010.

\bibitem{lattimore2020bandit}
Tor Lattimore and Csaba Szepesv{\'a}ri.
\newblock {\em Bandit algorithms}.
\newblock Cambridge University Press, 2020.

\bibitem{yang2023learning}
Zixian Yang, R~Srikant, and Lei Ying.
\newblock Learning while scheduling in multi-server systems with unknown statistics: Maxweight with discounted ucb.
\newblock In {\em International Conference on Artificial Intelligence and Statistics}, pages 4275--4312. PMLR, 2023.

\bibitem{Kempen2024Learning}
Sanne van Kempen, Jaron Sanders, Fiona Sloothaak, and Maarten~G. Wolf.
\newblock Learning payoffs while routing in skill-based queues, 2024.

\bibitem{zhong2024learning}
Yueyang Zhong, John~R Birge, and Amy~R Ward.
\newblock Learning to schedule in multiclass many-server queues with abandonment.
\newblock {\em Operations Research}, 2024.

\bibitem{Sun2024Inpatient}
Jingjing Sun, Jim Dai, and Pengyi Shi.
\newblock Inpatient overflow management with proximal policy optimization, 2024.

\bibitem{liu2019reinforcement}
Bai Liu, Qiaomin Xie, and Eytan Modiano.
\newblock Reinforcement learning for optimal control of queueing systems.
\newblock In {\em 2019 57th annual allerton conference on communication, control, and computing (allerton)}, pages 663--670. IEEE, 2019.

\bibitem{murthy2024performance}
Yashaswini Murthy, Isaac Grosof, Siva~Theja Maguluri, and R.~Srikant.
\newblock Performance of npg in countable state-space average-cost rl, 2024.

\bibitem{Weber2024Reinforcement}
Lucas Weber, Ana Bušić, and Jiamin Zhu.
\newblock Reinforcement learning and regret bounds for admission control, 2024.

\bibitem{Comte2023Score}
Céline Comte, Matthieu Jonckheere, Jaron Sanders, and Albert Senen-Cerda.
\newblock Score-aware policy-gradient methods and performance guarantees using local lyapunov conditions: Applications to product-form stochastic networks and queueing systems, 2023.

\bibitem{Chen2023Online}
Xinyun Chen, Yunan Liu, and Guiyu Hong.
\newblock Online learning and optimization for queues with unknown demand curve and service distribution, 2023.

\bibitem{kim2018value}
Jeunghyun Kim and Ramandeep~S Randhawa.
\newblock The value of dynamic pricing in large queueing systems.
\newblock {\em Operations Research}, 66(2):409--425, 2018.

\bibitem{neely2022stochastic}
Michael Neely.
\newblock {\em Stochastic network optimization with application to communication and queueing systems}.
\newblock Springer Nature, 2022.

\bibitem{varma_ht}
Sushil~Mahavir Varma and Siva~Theja Maguluri.
\newblock A heavy traffic theory of matching queues, 2021.

\bibitem{nesterov1994interior}
Yurii Nesterov and Arkadii Nemirovskii.
\newblock {\em Interior-point polynomial algorithms in convex programming}.
\newblock SIAM, 1994.

\bibitem{raeis2021queue}
Majid Raeis, Ali Tizghadam, and Alberto Leon-Garcia.
\newblock Queue-learning: A reinforcement learning approach for providing quality of service.
\newblock In {\em Proceedings of the AAAI Conference on Artificial Intelligence}, volume~35, pages 461--468, 2021.

\bibitem{dai2022queueing}
Jim~G Dai and Mark Gluzman.
\newblock Queueing network controls via deep reinforcement learning.
\newblock {\em Stochastic Systems}, 12(1):30--67, 2022.

\bibitem{liu2022rl}
Bai Liu, Qiaomin Xie, and Eytan Modiano.
\newblock Rl-qn: A reinforcement learning framework for optimal control of queueing systems.
\newblock {\em ACM Transactions on Modeling and Performance Evaluation of Computing Systems}, 7(1):1--35, 2022.

\bibitem{chen2024qgym}
Haozhe Chen, Ang Li, Ethan Che, Jing Dong, Tianyi Peng, and Hongseok Namkoong.
\newblock Qgym: Scalable simulation and benchmarking of queuing network controllers.
\newblock {\em Advances in Neural Information Processing Systems}, 37:92401--92419, 2024.

\bibitem{dani2008stochastic}
Varsha Dani, Thomas~P Hayes, and Sham~M Kakade.
\newblock Stochastic linear optimization under bandit feedback.
\newblock In {\em 21st Annual Conference on Learning Theory}, number 101, pages 355--366, 2008.

\bibitem{boyd2004convex}
Stephen Boyd and Lieven Vandenberghe.
\newblock {\em Convex optimization}.
\newblock Cambridge university press, 2004.

\bibitem{bertsekas1999nonlinear}
Dimitri Bertsekas.
\newblock {\em Nonlinear Programming}.
\newblock Athena Scientific, 1999.

\end{thebibliography}


\clearpage
\newpage
\appendix

\clearpage

\section{An Estimate-Then-Optimize Algorithm}
\label{app:example-predict-then-optimize}

Consider the following estimate-then-optimize algorithm.
First, we try prices uniformly in the price space. For each queue, to obtain an accuracy of $\zeta$ for the estimation of the demand/supply function, we need to discretize the space into $\Theta(1/\zeta)$ points, and for each point, we need at least $\Theta(1/\zeta^2)$ samples according to the central limit theorem. In this process, the queue length will increase linearly over time, up to $\Theta(1/\zeta^3)$. Using the estimated functions, we can obtain a solution with accuracy $\zeta$, and we can use this solution until the time horizon $T$. Hence, the worst-case average queue length will be $\Theta(1/\zeta^3 + (T - 1/\zeta^3)\zeta)$. The lowest worst-case average queue length that this approach can achieve is $\Theta(T^{3/4})$ with $\zeta=T^{-1/4}$, while our approach can achieve a much smaller average queue length $\tilde{\Theta}(T^{1/12})$ with $\gamma=1/6$ according to the results in Section~\ref{sec:overview-results}.

\section{Formal Statements of Theoretical Results}
\label{app:formal-theorems-analysis}

In this section, we will present formal theoretical results for the proposed algorithm described in Section~\ref{sec:alg}, and discuss the difficulties in the analysis.

\subsection{Assumptions}
To derive theoretical results, we impose the following additional assumptions.

We make the following assumption on the knowledge of the initial feasible point and the initial price searching intervals as in \cite{yang2024learning}.
\begin{assumption}[Initial Balanced Arrival Rates]\label{assum:5}
    Assume that the initial point $\boldsymbol{x}(1)\in \mathcal{D}'$ and the initial pricing searching intervals
    $\{[\underline{p}_{\mathrm{c}, i},\bar{p}_{\mathrm{c},i}], [\underline{p}_{\mathrm{s}, j}, \bar{p}_{\mathrm{s},j}]\}
    _{i\in {\cal I}, j\in {\cal J}}$
    satisfy
    \begin{align*}
        \sum_{j\in {\cal E}_{\mathrm{c}, i}} x_{i,j}(1) \in & \biggl[F^{-1}_i(\bar{p}_{\mathrm{c},i})-\epsilon + \sqrt{|{\cal E}_{\mathrm{c}, i}|} \delta,
        F^{-1}_i(\underline{p}_{\mathrm{c},i}) + \epsilon - \sqrt{|{\cal E}_{\mathrm{c}, i}|} \delta\biggr], \mbox{for all } i\nonumber\\
        \sum_{i\in {\cal E}_{\mathrm{s}, j}} x_{i,j}(1) \in & \biggl[G^{-1}_j(\underline{p}_{\mathrm{s},j})-\epsilon + \sqrt{|{\cal E}_{\mathrm{s}, j}|} \delta,
        G^{-1}_j(\bar{p}_{\mathrm{s},j}) + \epsilon - \sqrt{|{\cal E}_{\mathrm{s}, j}|} \delta \biggr], \mbox{for all } j,\nonumber\\
        \bar{p}_{\mathrm{c},i} - \underline{p}_{\mathrm{c},i} \le & 2e_{\mathrm{c}, i}, \quad
        \bar{p}_{\mathrm{s},j} - \underline{p}_{\mathrm{s},j} \le 2 e_{\mathrm{s}, j}, \quad \mbox{for all } i, j.
    \end{align*}
\end{assumption}
The small numbers $e_{\mathrm{c}, i}$ and $e_{\mathrm{s}, j}$ can be found in Appendix~\ref{app:exact-expression}.

\textbf{Discussion on Assumption~\ref{assum:5}:}
As in \cite{yang2024learning}, we assume that a feasible point $\boldsymbol{x}(1)\in \mathcal{D}'$ and the corresponding price bounds are known. We set the initial feasible point to be this point and use these bounds as the initial price searching intervals. Note that this point $\boldsymbol{x}(1)$ and the prices are usually not optimal. 
This assumption is mild, because in real application, we usually have some existing pricing strategy that may not be optimal but usually has approximately balanced arrival rates.

We make the following assumption about the objective function $f$ and an optimal solution.
\begin{assumption}[Weaker Strong-Convexity]\label{assum:6}
    There exists a positive constant $\nu$ such that there is an optimal solution $(\boldsymbol{\lambda}^*, \boldsymbol{\mu}^*, \boldsymbol{x}^*)$ satisfying Assumption~\ref{assum:3} and the following:
    \begin{align}\label{equ:strong-convexity}
        f(\boldsymbol{x}^*) - f(\boldsymbol{x}) \ge \frac{\nu}{2} \| \boldsymbol{x} - \boldsymbol{x}^* \|_2^2,
    \end{align}
    for all $\boldsymbol{x}\in \mathcal{D}'$.
\end{assumption}

\textbf{Discussion on Assumption~\ref{assum:6}:}
The equation \eqref{equ:strong-convexity} in Assumption~\ref{assum:6} means that the optimal solution satisfies a property that is weaker than strong convexity. In fact, it is a necessary condition for strong convexity.

Under Assumption~\ref{assum:3} and Assumption~\ref{assum:6}, without loss of generality, we can assume $a_{\min}$ satisfies
\begin{align}
    \sum_{j\in {\cal E}_{\mathrm{c},i}} \frac{a_{\min}+1}{2N_{i,j}} - a_{\min} > 0 \quad \text{for all } i \label{equ:cond-a-min-1}\\
    \sum_{i\in {\cal E}_{\mathrm{s},j}} \frac{a_{\min}+1}{2N_{i,j}} - a_{\min} > 0
    \quad \text{for all } j \label{equ:cond-a-min-2}
\end{align}
To see this, notice that \eqref{equ:cond-a-min-1} and \eqref{equ:cond-a-min-2} hold as long as the constant $a_{\min}$ is sufficiently small. If \eqref{equ:cond-a-min-1} or \eqref{equ:cond-a-min-2} does not hold, we can always find another constant $a'_{\min}$ such that $0 < a'_{\min} \le a_{\min}$ and $a'_{\min}$ satisfies \eqref{equ:cond-a-min-1} and \eqref{equ:cond-a-min-2}. Then, we can use this $a'_{\min}$ and all the results in this paper still hold. 

We make the following assumption about the functions $F_i^{-1}$, $G_j^{-1}$, $F_i$, and $G_j$.
\begin{assumption}[Smoothness \& Nonzero Derivatives]
\label{assum:7}
We make the following assumptions on the demand and supply functions.
\begin{itemize}[leftmargin=15pt]
    \item[(1)] Assume for all customer type $i$ and all server type $j$, the function $F^{-1}_i$ is $\beta_{F_i^{-1}}$-smooth, the function $G^{-1}_j$ is $\beta_{G_j^{-1}}$-smooth, the function $F_i$ is $\beta_{F_i}$-smooth, and the function $G_j$ is $\beta_{G_j}$-smooth, i.e.,
    \begin{align*}
        & \text{\small $\left| \diff*{F_i^{-1}(p)}{p}{p=p_1} -  \diff*{F_i^{-1}(p)}{p}{p=p_2} \right| 
        \le \beta_{F_i^{-1}} | p_1 - p_2 |,$}\nonumber\\
        & \text{\small $\left| \diff*{G_j^{-1}(p)}{p}{p=p_1} -  \diff*{G_j^{-1}(p)}{p}{p=p_2} \right| 
        \le \beta_{G_j^{-1}} | p_1 - p_2 |,$} \nonumber\\
        & \text{\small $\left| \diff*{F_i(\lambda)}{\lambda}{\lambda=\lambda_1} -  \diff*{F_i(\lambda)}{\lambda}{\lambda=\lambda_2} \right| 
        \le \beta_{F_i} | \lambda_1 - \lambda_2 |,$}\nonumber\\
        & \text{\small $\left| \diff*{G_j(\mu)}{\mu}{\mu=\mu_1} -  \diff*{G_j(\mu)}{\mu}{\mu=\mu_2} \right| 
        \le \beta_{G_j} | \mu_1 - \mu_2 |$}
    \end{align*}
    for all prices $p_1,p_2$ and all arrival rates $\lambda_1, \lambda_2, \mu_1, \mu_2$.
    \item[(2)] Assume for all customer type $i$ and all server type $j$, the derivatives of $F^{-1}_i$ and $G^{-1}_j$ are lower bounded by a constant, i.e.,
    \begin{align*}
        \left|\diff{F_i^{-1}(p)}{p}\right| \ge & C_{\mathrm{L}} \mbox{ for all } p \in [p_{\mathrm c,i,\min}, p_{\mathrm c, i,\max}],\nonumber\\
        \, \left|\diff{G_j^{-1}(p)}{p}\right| \ge & C_{\mathrm{L}} \mbox{ for all } p \in [p_{\mathrm s,j,\min}, p_{\mathrm s, j,\max}],
    \end{align*}
    where $C_{\mathrm{L}}$ is a constant independent of $T$.
\end{itemize}
\end{assumption}

\textbf{Discussion on Assumption~\ref{assum:7}:}
Assumption~\ref{assum:7}(1) means that the demand and supply functions $F_i^{-1}$, $G_j^{-1}$, $F_i$, and $G_j$ are smooth. The smoothness assumption is common in the optimization literature.
Assumption~\ref{assum:7}(2) means that the derivatives of these functions are bounded away from zero, which is necessary to ensure that the arrival rate can be reduced through adjusting the price. Nonzero derivatives is usually true in practice.

\subsection{Theoretical Bounds}
Suppose we use the proposed algorithm described in Section~\ref{sec:alg}. Let the parameters $\eta$, $\epsilon$, $\delta$, $\alpha$, $q^{\mathrm{th}}$ be functions of $T$. Let $\eta$, $\epsilon$, $\delta$, $\alpha$ be nonincreasing in $T$ and $q^{\mathrm{th}}$ be nondecreasing in $T$.
We have the following theoretical bounds.
\begin{theorem}\label{theo:3}
    Let Assumption~\ref{assum:1}, \ref{assum:2}, \ref{assum:3}, \ref{assum:5}, \ref{assum:6}, \ref{assum:7} hold.
    Suppose $\epsilon < \delta$, $T \ge 2MN$, $\alpha$ is orderwise greater than $\left(\frac{\eta \epsilon}{\delta} + \eta + \delta + \epsilon\right)$, i.e.,
    $
    \lim_{T\rightarrow \infty} (\frac{\eta \epsilon}{\delta} + \eta + \delta + \epsilon)/\alpha = 0,
    $
    then for sufficiently large $T$, we have
    \begin{align}\label{equ:regret-bound-3}
        & \expt[R(T)]
        =  O \Biggl( 
        \frac{T}{q^{\mathrm{th}}}  + \frac{\log^2 (1/\epsilon) }{\eta \epsilon^2 } 
        + T \left( \frac{\epsilon}{\delta} + \eta + \delta \right) 
        + \frac{\alpha \sqrt{T} \log (1/\epsilon)}{\epsilon \sqrt{\eta}} + T \alpha \left( \sqrt{\eta} + \sqrt{\frac{\epsilon}{\delta}} + \sqrt{\delta}\right) \nonumber\\
        & \qquad \qquad \qquad + T\alpha^2 
        + q^{\mathrm{th}} + T^2\epsilon^{\frac{\beta}{2}+1} + T\epsilon^{\frac{\beta}{2}-1} \log (1/\epsilon) 
        \Biggr),
    \end{align}
    and the queue lengths
    \begin{align}
        \text{AvgQLen}(T) \le \Theta\left( \frac{1}{\alpha} + \frac{q^{\mathrm{th}}}{\alpha} \left(T\epsilon^{\frac{\beta}{2}+1} + \epsilon^{\frac{\beta}{2}-1} \log (1/\epsilon) 
        \right)  \right),
        \quad \text{MaxQLen}(T) \le q^{\mathrm{th}}.\label{equ:avg-queue-length-bound-3}
    \end{align}
\end{theorem}
Proof of Theorem~\ref{theo:3} can be found in Appendix~\ref{app:theo:3}.
In the regret bound \eqref{equ:regret-bound-3}, the term $T/q^{\mathrm{th}}$ is caused by rejecting arrivals when queue lengths exceed $q^{\mathrm{th}}$.
The term $\log^2 (1/\epsilon) / (\eta \epsilon^2) 
+ T ( \epsilon/\delta + \eta + \delta )$ is caused by the zero-order projected stochastic gradient ascent with biased gradient estimation. The term $\alpha \sqrt{T} \log (1/\epsilon)/(\epsilon \sqrt{\eta}) + T \alpha ( \sqrt{\eta} + \sqrt{\frac{\epsilon}{\delta}} + \sqrt{\delta}) + T\alpha^2 + q^{\mathrm{th}}$ is caused by reducing arrival rates through adjusting the prices by $\alpha$. The term $T^2\epsilon^{\frac{\beta}{2}+1} + T\epsilon^{\frac{\beta}{2}-1} \log (1/\epsilon)$ is caused by the estimation error of the arrival rates. 

In order to obtain the tradeoff among regret, average queue length, and maximum queue length, we first set $q^{\mathrm{th}}=T^{\gamma}$ for a fixed $\gamma$. By first optimizing the order of the regret bound \eqref{equ:regret-bound-3} with respect to the parameters $\epsilon,\eta,\delta,\alpha$, and then optimizing the order of the average queue length bound \eqref{equ:avg-queue-length-bound-3} over $\alpha$ without compromising the order of the regret bound, we can obtain the following corollary.
\begin{corollary}\label{cor:7}
    Let all the assumptions in Theorem~\ref{theo:3} hold. For any $\gamma \in (0, \frac{1}{6}]$, setting parameters $q^{\mathrm{th}}= T^{\gamma}$, $\epsilon=T^{-2\gamma}$, $\eta=\delta=T^{-\gamma}$, $\alpha=T^{-\gamma/2}$, $\beta=1/\gamma-1$, for sufficiently large $T$, we can achieve a sublinear regret as
    \begin{align*}
        \expt[R(T)] = \tilde{O}(T^{1-\gamma}),
    \end{align*}
    and the queue lengths
    \begin{align*}
        \text{AvgQLen}(T) =  \tilde{O} ( T^{\frac{\gamma}{2}} ), 
        \quad \text{MaxQLen}(T) \le  T^{\gamma}.
    \end{align*}
    For any $\gamma \in (\frac{1}{6}, 1]$, setting parameters $q^{\mathrm{th}}= T^{\gamma}$, $\epsilon=T^{-1/3}$, $\eta=\delta=T^{-1/6}$, $\alpha=T^{-1/12}$, $\beta=5$, for sufficiently large $T$, we can achieve a sublinear regret as
    \begin{align*}
        \expt[R(T)] = \tilde{O}(T^{5/6}),
    \end{align*}
    and the queue lengths
    \begin{align*}
        \text{AvgQLen}(T) =  \tilde{O} ( T^{1/12} ), 
        \quad \text{MaxQLen}(T) \le  T^{\gamma}.
    \end{align*}
\end{corollary}
We can see from Corollary~\ref{cor:7} that compared to \cite{yang2024learning}, the average queue length bound is significantly improved from $\tilde{O} ( T^{\gamma} )$ to $\tilde{O} ( T^{\frac{\gamma}{2}} )$ with the use of the \textit{probabilistic two-price policy} while the regret bound remains the same order, .
We remark that this tradeoff between regret and average queue length, $\tilde{O}(T^{1-\gamma})$ versus $\tilde{O} ( T^{\frac{\gamma}{2}} )$, matches the optimal tradeoff in \cite{varma2023dynamic} up to logarithmic order, although they do not consider learning in their setting.
As the allowable maximum queue length increases (up to $T^{1/6}$), the achievable regret bound improves. However, increasing the maximum queue length beyond $T^{1/6}$ does not further enhance the regret bound, which remains at $\tilde{O}(T^{5/6})$ (after parameter optimization). Hence, we will not let $\gamma$ increase over $1/6$.

\subsection{Discussion}

\textbf{Reason for choosing a probabilistic approach:} 
The probabilistic pricing scheme adjusts the price to reduce the arrival rate with a certain probability for each nonempty queue. Therefore, for each nonempty queue, with a certain probability, the arrival rate will be strictly smaller than the sum of the average matching rates over all the connected links. So the lengths of nonempty queues will decrease, making the average queue length bounded.
Other approaches, such as threshold policies in \cite{varma2023dynamic}, can also control the average queue length, but they will induce a larger regret or do not have any theoretical guarantee for the induced regret.
The reason is that these threshold approaches introduce strong correlation among price adjustment decisions across different time slots, making it difficult to bound the number of useless samples. 
The proposed probabilistic pricing decouples this correlation, providing theoretically guaranteed regret and average queue length,
as well as a near-optimal tradeoff between them.

\textbf{Range of $\gamma$:} One (possible) limitation of our result is the restricted range of $\gamma \in (0, 1/6]$. First, note that, in the learning setting, we cannot expect to have a better regret than $\sqrt{T}$ as shown in \cite{dani2008stochastic}. Thus, we should expect to have at best $\gamma \in (0, 1/2]$. In our setting, we incur an even larger regret of $T^{5/6}$ as the function we are maximizing is a concave function of the arrival rates but we cannot directly set the arrival rates; we are only allowed to set the prices which in turn determines the arrival rates. This extra dependency forces us to implement a line search to first try several different prices to set the correct arrival rate, which in turn allows us to implement one iteration of the gradient algorithm. This additional step seems to be necessary due to the constraints imposed by the system and results in an additional regret. Having said that, we are not sure what the best possible regret is (somewhere between $\sqrt{T}$ and $T^{5/6}$), and it is an interesting future direction. We believe such a lower bound is non-trivial due to the dual learning setup (using prices to learn the arrival rate and then using the arrival rate to learn the optimal profit).

\textbf{Possible Extension of the Probabilistic Approach:}
In problems of joint queueing, learning, and optimization, the correlation among them is usually
complicated. One of the fundamental trade-offs in such problems is to optimize an objective while
ensuring sustained low queue lengths. Thus, making mistakes during online learning in queues has a
sustained effect as queues build up, and so, one needs to carefully select the actions (prices in our
case) to determine the optimal action while keeping low queue lengths. The idea of a probabilistic
algorithm partially decouples these two phenomena: it allows to exploration of the prices to optimize
the profit while ensuring low queue lengths. This decoupling allows for a tight analysis, resulting in
characterizing the optimal trade-off between profit and queue length. Thus, we believe that the idea
of using such a probabilistic algorithm could be widely applicable in the context of online learning in
queues.

\clearpage
\newpage

\section{
\texorpdfstring{Definitions of the Shrunk Feasible Set ${\cal D}'$}{Definitions of the Shrunk Feasible Set}
}
\label{app:def-shrunk-set}
The definition of the shrunk set ${\cal D}'$ is as follows.
\begin{align}\label{equ:def-D'}
    & {\cal D}' \coloneqq \Biggl\{ \boldsymbol{x} \Biggl|
    \text{ for all } i,j,
    x_{i,j}-\frac{a_{\min}+1}{2N_{i,j}} \ge - \biggl(1-\frac{\delta}{r}\biggr) \frac{a_{\min}+1}{2N_{i,j}},\nonumber\\
    & \sum_{j'\in {\cal E}_{\mathrm{c},i}} \biggl(x_{i,j'} - \frac{a_{\min}+1}{2N_{i,j'}}\biggr)\in 
    \biggl[ -\biggl(1-\frac{\delta}{r}\biggr)\biggl( \sum_{j'\in {\cal E}_{\mathrm{c},i}} \frac{a_{\min}+1}{2N_{i,j'}} - a_{\min} \biggr),\nonumber\\ 
    & \qquad  \qquad \qquad \qquad \qquad \qquad \biggl(1-\frac{\delta}{r}\biggr)  \biggl(  1 - \sum_{j'\in {\cal E}_{\mathrm{c},i}} \frac{a_{\min}+1}{2N_{i,j'}} \biggr) \biggr],\nonumber\\
    & \sum_{i'\in {\cal E}_{\mathrm{s},j}} \biggl(x_{i',j} - \frac{a_{\min}+1}{2N_{i',j}}\biggr)\in 
    \biggl[ -\biggl(1-\frac{\delta}{r}\biggr)\biggl( \sum_{i'\in {\cal E}_{\mathrm{s},j}} \frac{a_{\min}+1}{2N_{i',j}} - a_{\min} \biggr), \nonumber\\ 
    & \qquad  \qquad \qquad \qquad \qquad \qquad
    \biggl(1-\frac{\delta}{r}\biggr)  \biggl(  1 - \sum_{i'\in {\cal E}_{\mathrm{s},j}} \frac{a_{\min}+1}{2N_{i',j}} \biggr) \biggr]
    \Biggr.\Biggr\},
\end{align}
where $N_{i,j}\coloneqq \max\{|{\cal E}_{\mathrm{c},i}|, |{\cal E}_{\mathrm{s},j}|\}$ denotes the maximum cardinality of the sets ${\cal E}_{\mathrm{c},i}$ and ${\cal E}_{\mathrm{s},j}$,
\begin{align}\label{equ:def-r}
    r \coloneqq & \min_{i,j} \Biggl\{
    \frac{1+a_{\min}}{2N_{i,j}},
    \frac{1}{|{\cal E}_{\mathrm{c}, i}|}\biggl( 1 -  \sum_{j'\in {\cal E}_{\mathrm{c},i}} \frac{a_{\min}+1}{2N_{i,j'}} \biggr), \frac{1}{|{\cal E}_{\mathrm{s}, j}|}\biggl( 1 -  \sum_{i'\in {\cal E}_{\mathrm{s},j}} \frac{a_{\min}+1}{2N_{i',j}} \biggr), \nonumber\\
    & \qquad \frac{1}{|{\cal E}_{\mathrm{c}, i}|} \biggl( \sum_{j'\in {\cal E}_{\mathrm{c},i}} \frac{a_{\min}+1}{2N_{i,j'}} - a_{\min} \biggr),
    \frac{1}{|{\cal E}_{\mathrm{s}, j}|} \biggl( \sum_{i'\in {\cal E}_{\mathrm{s},j}} \frac{a_{\min}+1}{2N_{i',j}} - a_{\min} \biggr)
    \Biggr\},
\end{align}
and $\delta \in (0 ,r)$. 

Under the conditions \eqref{equ:cond-a-min-1} and \eqref{equ:cond-a-min-2}, it can be shown that $\boldsymbol{x} + \delta \boldsymbol{u} \in {\cal D}$ for any $\boldsymbol{x}\in {\cal D}'$ and any vector $\boldsymbol{u}$ in the unit ball~\cite{yang2024learning}.

\clearpage
\newpage

\section{Pricing Algorithm -- Two-Point Zero-Order Projected Gradient Ascent}
\label{app:alg-pricing}

We present the details of the pricing algorithm -- two-point zero-order projected gradient ascent in Algorithm~\ref{alg:pricing}, which is the same as the balanced pricing algorithm in \cite{yang2024learning}.
The index $(k)$ denotes the $k^{\mathrm{th}}$ outer iteration (iteration of the gradient ascent) and $(k,m)$ denotes the $m^{\mathrm{th}}$ bisection iteration in the $k^{\mathrm{th}}$ outer iteration.

\clearpage

\begin{algorithm}[ht]
\caption{Pricing algorithm}\label{alg:pricing}
\begin{algorithmic}[1]
\STATE \textbf{Initialize:} Choose an exploration parameter $\delta\in(0,r)$.
Choose a step size $\eta\in (0,1)$. 
Choose an accuracy parameter $\epsilon\in (0, 1/e)$.
Choose a queue length threshold $q^{\mathrm{th}}$.
Choose a two-price parameter $\alpha$.
Choose $\boldsymbol{x}(1) \in {\cal D}'$ as the initial point.
Choose some initial price searching intervals $[\underline{p}_{\mathrm{c}, i}, \bar{p}_{\mathrm{c},i}]$ for customer type $i$, $[\underline{p}_{\mathrm{s}, j}, \bar{p}_{\mathrm{s},j}]$ for server type $j$.
Define $e_{\mathrm{c},i}$ and $e_{\mathrm{s},j}$ according to \eqref{equ:def-e-c} and \eqref{equ:def-e-s} in Appendix~\ref{app:exact-expression}.
Define $N\coloneqq \left\lceil\frac{\beta\ln (1/\epsilon)}{\epsilon^2}\right\rceil$ and $M \coloneqq \left\lceil \log_2 \frac{1}{\epsilon} \right\rceil$, where $\beta>0$ is a constant.
\STATE Time step counter $t\gets 1$
\STATE Outer iteration counter $k\gets 1$
\REPEAT
    \STATE \blue{// \textbf{generate a random direction and two points}}
    \STATE Choose a unit vector $\boldsymbol{u}(k)\in \mathbb{R}^{|\cal E|}$ uniformly at random, i.e., $\|\boldsymbol{u}(k)\|_2 = 1$
    \STATE Let $x^{+}_{i,j}(k)\coloneqq (\boldsymbol{x}(k) + \delta \boldsymbol{u}(k))_{i,j}$ and $x^{-}_{i,j}(k)\coloneqq (\boldsymbol{x}(k) - \delta \boldsymbol{u}(k))_{i,j}$\label{line:alg-pricing-x}
    \STATE Let $\lambda_i^{+}(k)\coloneqq \sum_{j\in {\cal E}_{\mathrm{c},i}} x^{+}_{i,j}(k)$,
    $\lambda_i^{-}(k)\coloneqq \sum_{j\in {\cal E}_{\mathrm{c},i}} x^{-}_{i,j}(k)$,
    $\mu_j^{+}(k)\coloneqq \sum_{i\in {\cal E}_{\mathrm{s},j}} x^{+}_{i,j}(k)$,
    $\mu_j^{-}(k)\coloneqq \sum_{i\in {\cal E}_{\mathrm{s},j}} x^{-}_{i,j}(k)$\label{line:alg-pricing-lambda-mu}
    \STATE Let $\boldsymbol{\lambda}^{+}(k)$ be a vector of $\lambda_i^{+}(k),i=1\ldots,I$. Define similarly $\boldsymbol{\lambda}^{-}(k)$, $\boldsymbol{\mu}^{+}(k)$, and $\boldsymbol{\mu}^{-}(k)$
    \STATE \blue{// \textbf{bisection search to approximate $F_i(\lambda_i^{+}(k)), F_i(\lambda_i^{-}(k)), G_j(\mu_j^{+}(k)), G_j(\mu_j^{-}(k))$ to estimate the profits in the two points.}}
    \IF{$k=1$}\label{line:alg-pricing-binary-start}
        \STATE For all $i$, let $\underline{p}^{+}_{\mathrm{c}, i}(k,1)=\underline{p}_{\mathrm{c}, i}$,
        $\bar{p}^{+}_{\mathrm{c},i}(k,1)=\bar{p}_{\mathrm{c},i}$\label{line:alg-pricing-bound-bisection-1-i}
        \STATE For all $j$, let $\underline{p}^{+}_{\mathrm{s}, j}(k,1)=\underline{p}_{\mathrm{s}, j}$,
        $\bar{p}^{+}_{\mathrm{s},j}(k,1)=\bar{p}_{\mathrm{s},j}$\label{line:alg-pricing-bound-bisection-1-j}
    \ELSE
        \STATE For all $i$, let $\underline{p}^{+}_{\mathrm{c}, i}(k,1)=p_{\mathrm{c},i}^+(k-1,M) - e_{\mathrm{c},i}$,
        $\bar{p}^{+}_{\mathrm{c},i}(k,1)=p_{\mathrm{c},i}^+(k-1, M) + e_{\mathrm{c},i}$\label{line:alg-pricing-bound-bisection-i}
        \STATE For all $j$, let $\underline{p}^{+}_{\mathrm{s}, j}(k,1)=p_{\mathrm{s},j}^+(k-1, M) - e_{\mathrm{s},j}$,
        $\bar{p}^{+}_{\mathrm{s},j}(k,1)=p_{\mathrm{s},j}^+(k-1, M) + e_{\mathrm{s},j}$\label{line:alg-pricing-bound-bisection-j}
    \ENDIF\label{line:alg-pricing-set-interval-end}
    \STATE Let $\underline{\boldsymbol{p}}^{+}_{\mathrm{c}}(k,1)$ be a vector of $\underline{p}^{+}_{\mathrm{c}, i}(k,1), i=1\ldots,I$. Define similarly $\bar{\boldsymbol{p}}^{+}_{\mathrm{c}}(k,1)$, $\underline{\boldsymbol{p}}^{+}_{\mathrm{s}}(k,1)$, $\bar{\boldsymbol{p}}^{+}_{\mathrm{s}}(k,1)$.
    \STATE $t$, $\boldsymbol{p}_{\mathrm{c}}^{+} (k,M)$, $\boldsymbol{p}_{\mathrm{s}}^{+} (k,M)=$
    \STATE \qquad \quad Bisection$\left(\boldsymbol{\lambda}^{+}(k), \boldsymbol{\mu}^{+}(k), \underline{\boldsymbol{p}}^{+}_{\mathrm{c}}(k,1), \bar{\boldsymbol{p}}^{+}_{\mathrm{c}}(k,1), \underline{\boldsymbol{p}}^{+}_{\mathrm{s}}(k,1), \bar{\boldsymbol{p}}^{+}_{\mathrm{s}}(k,1),
    \alpha,
    q^{\mathrm{th}}, t, M, N, \epsilon\right)$ \label{line:alg-pricing-binary-end}
    \STATE Do Line~\ref{line:alg-pricing-binary-start}-\ref{line:alg-pricing-binary-end} for $\boldsymbol{\lambda}^{-}(k), \boldsymbol{\mu}^{-}(k)$. Denote the counterparts of the prices by $\underline{\boldsymbol{p}}^{-}_{\mathrm{c}}(k,1)$, $\bar{\boldsymbol{p}}^{-}_{\mathrm{c}}(k,1)$,
    $\boldsymbol{p}^{-}_{\mathrm{c}}(k,M)$, $\underline{\boldsymbol{p}}^{-}_{\mathrm{s}}(k,1)$, $\bar{\boldsymbol{p}}^{-}_{\mathrm{s}}(k,1)$, $\boldsymbol{p}^{-}_{\mathrm{s}}(k,M)$\label{line:alg-pricing-bisection}.
    \STATE \blue{// \textbf{gradient calculation}}
    \STATE Let $\hat{\boldsymbol{g}}(k)=\frac{|{\cal E}|}{2\delta} \biggl[ \left( \sum_{i=1}^{I} \lambda_i^+(k) p_{\mathrm{c},i}^+(k, M) - \sum_{j=1}^{J} \mu_j^+(k) p_{\mathrm{s},j}^+(k, M) \right)$\label{line:alg-pricing-gradient}
    \STATE \qquad \qquad \qquad \quad$ - \left( \sum_{i=1}^{I} \lambda_i^-(k) p_{\mathrm{c},i}^-(k, M) - \sum_{j=1}^{J} \mu_j^-(k) p_{\mathrm{s},j}^-(k, M)
    \right) \biggr] \boldsymbol{u}(k)$
    \STATE \blue{// \textbf{gradient ascent update}}
    \STATE Projected Gradient Ascent: $\boldsymbol{x}(k+1) = \Pi_{{\cal D}'}(\boldsymbol{x}(k)+\eta \hat{\boldsymbol{g}}(k) )$;\label{line:alg-pricing-pga}
    \STATE $k\gets k + 1$;
\UNTIL{$t > T$}
\end{algorithmic}
\end{algorithm}

\clearpage
\newpage

\section{Definitions of \texorpdfstring{$e_{\mathrm{c},i}$ and $e_{\mathrm{s},j}$}{e\_\{c,i\} and e\_\{s,j\}}}
\label{app:exact-expression}

The definitions of $e_{\mathrm{c},i}$ and $e_{\mathrm{s},j}$ are shown as follows:
\begin{align}
    e_{\mathrm{c},i}
    =&\frac{2 \eta \epsilon |{\cal E}|^{3/2} L_{F_i} }{\delta} 
     \Biggl[ \sum_{i'=1}^{I} L_{F_{i'}}
     \left(1 + L_{F^{-1}_{i'}} \left(p_{\mathrm{c},i',\max} - p_{\mathrm{c},i',\min}\right)\right)\nonumber\\
     & + \sum_{j'=1}^{J} L_{G_{j'}}
    \left( 1 + L_{G^{-1}_{j'}}
    \left(p_{\mathrm{s},j',\max} - p_{\mathrm{s},j',\min}\right)\right)\Biggr]
    + 2\epsilon L_{F_i} \left(1 + L_{F^{-1}_i} \left(p_{\mathrm{c},i,\max} - p_{\mathrm{c},i,\min}\right)\right)\nonumber\\
    & + \eta |{\cal E}|^{3/2} L_{F_i} \left( \sum_{i'=1}^{I} |{\cal E}_{\mathrm{c},i'}| ( L_{F_{i'}} + p_{\mathrm{c}, i', \max} ) + \sum_{j'=1}^{J}  |{\cal E}_{\mathrm{s},j'}| ( L_{G_{j'}} + p_{\mathrm{s}, j', \max} ) \right) + 2 \delta |{\cal E}|^{1/2} L_{F_i}\label{equ:def-e-c}\\
    = & \Theta\left(\frac{\eta \epsilon}{\delta} + \eta + \delta + \epsilon\right),\nonumber
\end{align}

\begin{align}
    e_{\mathrm{s},j}
    =&\frac{2 \eta \epsilon |{\cal E}|^{3/2} L_{G_j} }{\delta} 
     \Biggl[ \sum_{i'=1}^{I} L_{F_{i'}}
     \left(1 + L_{F^{-1}_{i'}} \left(p_{\mathrm{c},i',\max} - p_{\mathrm{c},i',\min}\right)\right)\nonumber\\
    & + \sum_{j'=1}^{J} L_{G_{j'}}
    \left( 1 + L_{G^{-1}_{j'}}
    \left(p_{\mathrm{s},j',\max} - p_{\mathrm{s},j',\min}\right)\right)\Biggr]
    + 2\epsilon L_{G_j} \left(1 + L_{G^{-1}_j} \left( p_{\mathrm{s},j,\max} - p_{\mathrm{s},j,\min}\right)\right)\nonumber\\
    & + \eta |{\cal E}|^{3/2} L_{G_j} \left( \sum_{i'=1}^{I} |{\cal E}_{\mathrm{c},i'}| ( L_{F_{i'}} + p_{\mathrm{c}, i', \max} ) + \sum_{j'=1}^{J}  |{\cal E}_{\mathrm{s},j'}| ( L_{G_{j'}} + p_{\mathrm{s}, j', \max} ) \right) + 2 \delta |{\cal E}|^{1/2} L_{G_j}\label{equ:def-e-s}\\
    = & \Theta\left(\frac{\eta \epsilon}{\delta} + \eta + \delta + \epsilon\right)\nonumber.
\end{align}

\clearpage
\newpage

\section{Bisection Search}
\label{app:bisection}

We present the details of the bisection search method in Algorithm~\ref{alg:bisection}, which is similar to that in \cite{yang2024learning}. The key difference over \cite{yang2024learning} is how we run the system to collect samples. The inputs to the bisection search include the arrival rates $\boldsymbol{\lambda}^{+/-}(k)$ and $\boldsymbol{\mu}^{+/-}(k)$, the price searching intervals $\underline{\boldsymbol{p}}^{+/-}_{\mathrm{c}}(k,1), \bar{\boldsymbol{p}}^{+/-}_{\mathrm{c}}(k,1), \underline{\boldsymbol{p}}^{+/-}_{\mathrm{s}}(k,1), \bar{\boldsymbol{p}}^{+/-}_{\mathrm{s}}(k,1)$, 
the two-price parameter $\alpha$,
the threshold parameter $q^{\mathrm{th}}$, the number of bisection iterations $M$, and the number of samples for each queue in each bisection iteration $N$. The bisection search will output the prices $\boldsymbol{p}^{+/-}_{\mathrm{c}}(k,M)$ and $\boldsymbol{p}^{+/-}_{\mathrm{s}}(k,M)$ such that $\lambda_i^{+/-}(k)\approx F_i^{-1}(p^{+/-}_{\mathrm{c},i}(k,M))$ and $\mu_j^{+/-}(k) \approx G_j^{-1}(p^{+/-}_{\mathrm{s},j}(k,M))$ for all $i,j$.

\clearpage

\begin{algorithm}[ht]
\caption{Bisection {\scriptsize $\left(\boldsymbol{\lambda}^{+/-}(k), \boldsymbol{\mu}^{+/-}(k), \underline{\boldsymbol{p}}^{+/-}_{\mathrm{c}}(k,1), \bar{\boldsymbol{p}}^{+/-}_{\mathrm{c}}(k,1), \underline{\boldsymbol{p}}^{+/-}_{\mathrm{s}}(k,1), \bar{\boldsymbol{p}}^{+/-}_{\mathrm{s}}(k,1), 
\alpha,
q^{\mathrm{th}}, t, M, N, \epsilon\right)$ }
}\label{alg:bisection}
\begin{algorithmic}[1]
\FOR{\textnormal{$m=1$ to $M$}}
    \STATE \blue{// \textbf{Calculate the midpoints}}
    \STATE Let $p^{+/-}_{\mathrm{c}, i} (k,m) = \frac{1}{2} \left(\underline{p}^{+/-}_{\mathrm{c}, i}(k,m) + \bar{p}^{+/-}_{\mathrm{c}, i}(k,m)\right)$ for all $i$
    \label{line:alg-bisection-middle-i}
    \STATE Let $p^{+/-}_{\mathrm{s}, j} (k,m) = \frac{1}{2} \left(\underline{p}^{+/-}_{\mathrm{s}, j}(k,m) + \bar{p}^{+/-}_{\mathrm{s}, j}(k,m)\right)$ for all $j$
    \label{line:alg-bisection-middle-j}
    \STATE \blue{// \textbf{Run the system to collect $N$ samples with the midpoints}}
    \STATE Let $t^{+/-}_{\mathrm{c}, i}(k,m,n)$ denote the time slot when the price $p^{+/-}_{\mathrm{c}, i} (k,m)$ is run for the $n^{\mathrm{th}}$ time for the customer queue $i$; Let $t^{+/-}_{\mathrm{s}, j}(k,m,n)$ denote the time slot when the price $p^{+/-}_{\mathrm{s}, j} (k,m)$ is run for the $n^{\mathrm{th}}$ time for the server queue $j$
    \STATE $t$, $(A_{\mathrm{c},i}(t^{+/-}_{\mathrm{c}, i}(k,m,n)))_{n\in [N], i\in{\cal I}}$, $(A_{\mathrm{s},j}(t^{+/-}_{\mathrm{s}, j}(k,m,n)))_{n\in[N], j\in {\cal J}}$\\
    = ProbTwoPrice$\left(\boldsymbol{p}^{+/-}_{\mathrm{c}} (k,m), \boldsymbol{p}^{+/-}_{\mathrm{s}} (k,m), \alpha, q^{\mathrm{th}}, N, t\right)$,\blue{~// Key innovation over \cite{yang2024learning}}\\
    where $\boldsymbol{p}^{+/-}_{\mathrm{c}} (k,m)$ denotes a vector of $p^{+/-}_{\mathrm{c}, i} (k,M),i\in {\cal I}$ and similarly for $\boldsymbol{p}^{+/-}_{\mathrm{s}} (k,m)$.
    \FOR{\textnormal{$i=1$ to $I$}}
        \STATE \blue{// \textbf{Estimate the arrival rates}}
        \STATE Let $\hat{\lambda}^{+/-}_i  (k,m)=\frac{1}{N}\sum_{n=1}^{N}A_{\mathrm{c},i}(t^{+/-}_{\mathrm{c}, i}(k,m,n))$
        \blue{// sample average}\label{line:alg-bisection-estimate-i}
        \STATE \blue{// \textbf{Update the price searching intervals}}
        \IF{$\hat{\lambda}^{+/-}_i  (k,m) > \lambda^{+/-}_{i} (k)$\label{line:alg-bisection-update-i-start}}
        \STATE $\underline{p}^{+/-}_{\mathrm{c}, i}(k,m+1) = p^{+/-}_{\mathrm{c}, i}  (k,m), 
        ~ \bar{p}^{+/-}_{\mathrm{c},i}(k,m+1) = \bar{p}^{+/-}_{\mathrm{c},i}(k,m)$
        \ELSE
        \STATE $\underline{p}^{+/-}_{\mathrm{c}, i}(k,m+1) = \underline{p}  ^{+/-}_{\mathrm{c}, i}(k,m),
        ~ \bar{p}^{+/-}_{\mathrm{c},i}(k,m+1) = p^{+/-}_{\mathrm{c}, i}  (k,m)$
        \label{line:alg-bisection-update-i-end}
        \ENDIF
    \ENDFOR
    \FOR{\textnormal{$j=1$ to $J$}}
        \STATE \blue{// \textbf{Estimate the arrival rates}}
        \STATE Let $\hat{\mu}^{+/-}_j  (k,m)=\frac{1}{N}\sum_{n=1}^{N} A_{\mathrm{s},j}(t^{+/-}_{\mathrm{s}, j}(k,m,n))$
        \blue{// sample average}\label{line:alg-bisection-estimate-j}
        \STATE \blue{// \textbf{Update the price searching intervals}}
        \IF{$\hat{\mu}^{+/-}_j  (k,m) > \mu^{+/-}_{j} (k)$\label{line:alg-bisection-update-j-start}}
        \STATE $\underline{p}^{+/-}_{\mathrm{s},j}(k,m+1) = \underline{p}  ^{+/-}_{\mathrm{s},j}(k,m), 
        ~ \bar{p}^{+/-}_{\mathrm{s}, j}(k,m+1) = p^{+/-}_{\mathrm{s}, j}  (k,m)$
        \ELSE
        \STATE $\underline{p}^{+/-}_{\mathrm{s}, j}(k,m+1) = p^{+/-}_{\mathrm{s}, j}  (k,m),  
        ~ \bar{p}^{+/-}_{\mathrm{s},j}(k,m+1) = \bar{p}^{+/-}_{\mathrm{s}, j}(k,m)$
        \label{line:alg-bisection-update-j-end}
        \ENDIF
    \ENDFOR   
\ENDFOR
\STATE \textbf{Return} $t$, $\boldsymbol{p}^{+/-}_{\mathrm{c}} (k,M)$, $\boldsymbol{p}^{+/-}_{\mathrm{s}} (k,M)$ 
\end{algorithmic}
\end{algorithm}

\clearpage
\newpage

\section{Probabilistic Two-Price Policy}
\label{app:alg-prob-two-price}

The pseudo-code of the proposed probabilistic two-price policy is presented in Algorithm~\ref{alg:rand-two-price}, where $t^{+/-}_{\mathrm{c}, i}(k,m,n)$ denotes the time slot in which the price of the midpoint is run for the $n^{\mathrm{th}}$ time for the customer queue $i$ and $t^{+/-}_{\mathrm{s}, j}(k,m,n)$ for server queue $j$.

\begin{algorithm}[ht]
\caption{\footnotesize ProbTwoPrice$\bigl(\boldsymbol{p}^{+/-}_{\mathrm{c}} (k,m), \boldsymbol{p}^{+/-}_{\mathrm{s}} (k,m), \alpha, q^{\mathrm{th}}, N, t\bigr)$}
\label{alg:rand-two-price}
\begin{algorithmic}[1]
\STATE Let $n_{\mathrm{c},i}(k,m)=0$ and $n_{\mathrm{s},j}(k,m)=0$ for all $i,j$.\label{line:alg-rand-two-price-counters}
\REPEAT
\FOR{\textnormal{$i=1$ to $I$}}
    \lIF{$Q_{\mathrm{c},i}(t)\ge q^{\mathrm{th}}$}
    {set price $p_{\mathrm{c},i,\max}$ for queue $i$
    \blue{ // to control the maximum queue length}}
    \label{line:alg-rand-two-price-maximum-customer}
    \lELSIF{$0 < Q_{\mathrm{c},i}(t)< q^{\mathrm{th}}$}
    {for queue $i$, set price: \blue{// to control the average queue length}
    \begin{align*}
    \left\{
    \begin{array}{ll}
         \min\{p^{+/-}_{\mathrm{c}, i} (k,m) + \alpha, p_{\mathrm{c},i,\max}\}, & \text{w.p. $1/2$}\nonumber\\
         p^{+/-}_{\mathrm{c}, i} (k,m), & \text{w.p. $1/2$}
    \end{array}
    \right.
    \end{align*}
    }
    \label{line:alg-rand-two-price-average-customer}
    \lELSE{set price $p^{+/-}_{\mathrm{c}, i} (k,m)$ for queue $i$.}
    \label{line:alg-rand-midpt-customer}
    \lIF{price=$p^{+/-}_{\mathrm{c}, i} (k,m)$} 
    {$n_{\mathrm{c},i}(k,m) \gets n_{\mathrm{c},i}(k,m) + 1$
    \blue{// the number of useful samples}}
    \label{line:alg-rand-useful-customer}
\ENDFOR
\FOR{\textnormal{$j=1$ to $J$}}
    \lIF{$Q_{\mathrm{s},j}(t)\ge q^{\mathrm{th}}$}
    {set price $p_{\mathrm{s},j,\min}$ for queue $j$
    \blue{// to control the maximum queue length}} 
    \label{line:alg-rand-two-price-maximum-server}
    \lELSIF{$0 < Q_{\mathrm{s},j}(t) < q^{\mathrm{th}}$}
    {for queue $j$, set price: \blue{// to control the average queue length}
    \begin{align*}
    \left\{
    \begin{array}{ll}
         \max\{p^{+/-}_{\mathrm{s}, j} (k,m) - \alpha, p_{\mathrm{s},j,\min}\}, & \text{w.p. $1/2$}\nonumber\\
         p^{+/-}_{\mathrm{s}, j} (k,m), & \text{w.p. $1/2$}
    \end{array}
    \right.
    \end{align*}
    }
    \label{line:alg-rand-two-price-average-server}
    \lELSE
    {set price $p^{+/-}_{\mathrm{s}, j} (k,m)$ for queue $j$.}
    \label{line:alg-rand-midpt-server}
    \lIF{price=$p^{+/-}_{\mathrm{s}, j} (k,m)$} 
    {$n_{\mathrm{s},j}(k,m) \gets n_{\mathrm{s},j}(k,m) + 1$
    \blue{// the number of useful samples}}
    \label{line:alg-rand-useful-server}
\ENDFOR
\STATE Run the system with the above set of prices for one time slot.
\STATE $t \leftarrow t + 1$
\STATE Terminate the algorithm when $t > T$
\UNTIL{for all queues $i$, $n_{\mathrm{c},i}(k,m)\ge N$, i.e., the price $p^{+/-}_{\mathrm{c}, i} (k,m)$ is run for at least $N$ times and for all queues $j$, $n_{\mathrm{s},j}(k,m)\ge N$, i.e., the price $p^{+/-}_{\mathrm{s}, j} (k,m)$ is run for at least $N$ times}
\label{line:alg-rand-N-samples}
\STATE \textbf{Return} $t$, $(A_{\mathrm{c},i}(t^{+/-}_{\mathrm{c}, i}(k,m,n)))_{n\in [N], i\in{\cal I}}$, $(A_{\mathrm{s},j}(t^{+/-}_{\mathrm{s}, j}(k,m,n)))_{n\in[N], j\in {\cal J}}$
\label{line:alg-rand-return}
\end{algorithmic}
\end{algorithm}

\clearpage
\newpage

\section{Details of the Simulation and Additional Numerical Results}
\label{app:simu}

\subsection{Single-Link System}

We present details of the simulation and additional simulation results in a single-link system in the following.

\subsubsection{Setting}
Consider a system with $I=1$ customer queue and $J=1$ server queue.
The demand function of the customer queue is $F(\lambda) = 2(1-\lambda)$, where $\lambda\in[0,1]$ is the arrival rate of the customer queue. The supply function of the server queue is $G(\mu) = 2\mu$, where $\mu\in[0,1]$ is the arrival rate of the server queue. We compare the performances among three algorithms:
\begin{itemize}[leftmargin=*]
    \item \textbf{Two-price policy} \cite{varma2023dynamic}: This policy assumes that \emph{the demand and supply functions are known}. To balance queue length and regret, this policy uses the following two-price method. If the queue length is empty, it uses the prices corresponding to the optimal solution of the fluid optimization problem. If the queue is nonempty, it uses a slightly-perturbed version of the optimal prices to reduce the arrival rates.
    \item \textbf{Threshold policy} \cite{yang2024learning}: This policy \emph{does not know} the demand and supply functions and uses a learning-based pricing algorithm. To balance queue length and regret, this policy uses a threshold method, rejecting arrivals when the queue length reaches or exceeds a threshold.
    \item \textbf{Probabilistic two-price policy} (proposed): This policy \emph{does not know} the demand and supply functions. It uses a novel probabilistic method, which adjusts the prices to reduce the queue length with a prefixed probability when the queue is nonempty. 
\end{itemize}

\subsubsection{Comparison}
For the proposed algorithm, we set $\gamma=1/6$, and $q^{\mathrm{th}}=t^{\gamma}$, $\epsilon=t^{-2\gamma}$, $\eta=\delta=0.2\times t^{-\gamma}$, $\alpha = 0.2\times t^{\gamma/2}$, $\beta=1.0$, $e_{\mathrm{c}}= e_{\mathrm{s}}=6.0\times \max\{\delta, \eta, \epsilon\}$, $a_{\min} = 0.01$, which are of the same order as in Corollary~\ref{cor:7}.
The common parameters of the three algorithm are set to be equal for fair comparison.  

To compare the performance of these algorithms, we consider an objective function defined as
\begin{align}\label{obj-org}
    & \sum_{\tau=1}^{t}\biggl( \sum_i \lambda_i(\tau)\left(F_i(\lambda_i(\tau)) -w\expt[W_{\mathrm{c},i}(\tau)]\right)\nonumber\\
    & \qquad - \sum_j \mu_j(\tau)\left(G_j(\mu_j(\tau)) + w\expt[W_{\mathrm{s},j}(\tau)]\right)\biggr),
\end{align}
where $W_{\mathrm{c},i}(\tau)$ denotes the waiting time of the customer arriving at queue $i$ at time slot $\tau$ and $W_{\mathrm{s},j}(\tau)$ denotes the waiting time of the server arriving at queue $j$ at time slot $\tau$. The constant $w>0$ can be viewed as the compensation the platform pays for per unit time the customer/server wait.
This objective \eqref{obj-org} is equivalent to
\begin{align}\label{obj}
    \text{Profit}(t) - w \left( \sum_i \sum_k \expt[W_{\mathrm{c},i,k}(t)] + \sum_j \sum_k \expt[W_{\mathrm{s},j,k}(t)] \right),
\end{align}
where $W_{\mathrm{c},i,k}(t)$ denotes the total waiting time up to time $t$ for the $k^{\mathrm{th}}$ customer arriving at queue $i$ , $W_{\mathrm{s},j,k}(t)$ denotes the total waiting time up to time $t$ for the $k^{\mathrm{th}}$ server arriving at queue $j$, and $w$ is a constant.
A similar objective is also used in \cite{varma2023dynamic}.
Note that the sum of waiting times of all customers and servers is equal to the sum of the queue lengths over time, which counts every customer and server present at each time slot. Hence, the objective \eqref{obj} is equal to
\begin{align*}
    \text{Profit}(t) - w \sum_{\tau=1}^{t} \expt \left[\sum_i Q_{\mathrm{c},i}(\tau) + \sum_j Q_{\mathrm{s},j}(\tau)\right].
\end{align*}
Then, the regret of this objective is given by:
\begin{align*}
    & f(\boldsymbol{x}^*) t - \text{Profit}(t) + w \sum_{\tau=1}^{t} \expt \left[\sum_i Q_{\mathrm{c},i}(\tau) + \sum_j Q_{\mathrm{s},j}(\tau)\right]\nonumber\\
    = & \expt[R(t)] + w \sum_{\tau=1}^{t} \expt \left[\sum_i Q_{\mathrm{c},i}(\tau) + \sum_j Q_{\mathrm{s},j}(\tau)\right]\nonumber\\
    = &  \expt[R(t)] + w t \text{AvgQLen}(t),
\end{align*}
which is the sum of the profit regret and the holding cost.

In the simulation, we test two different $w$, $w=0.001$ and $w=0.01$. Note that the optimal profit at each time slot is $f(\boldsymbol{x}^*)=0.25$ in the fluid solution. Therefore, choosing a even larger $w$ is not reasonable because further increasing $w$ will cause a negative utility even for the case where the demand and supply functions are known. 

Figure~\ref{fig:comp-0.001} in Section~\ref{sec:simu} shows the performance of these algorithms and the percentage improvement of the proposed \emph{probabilistic two-price policy} over the \emph{threshold policy} \cite{yang2024learning} with $w=0.001$. 

The shaded areas in all figures in this paper represent $95\%$ confidence intervals, calculated from 10 independent runs using the Python function \textit{seaborn.relplot}.

\begin{figure}[ht]
    \centering
    \subfloat[\texorpdfstring{$\expt[R(t)] + w t \text{AvgQLen}(t)$}{E[R(t)] + w t AvgQLen(t)}.]{
    \includegraphics[width=0.44\linewidth]{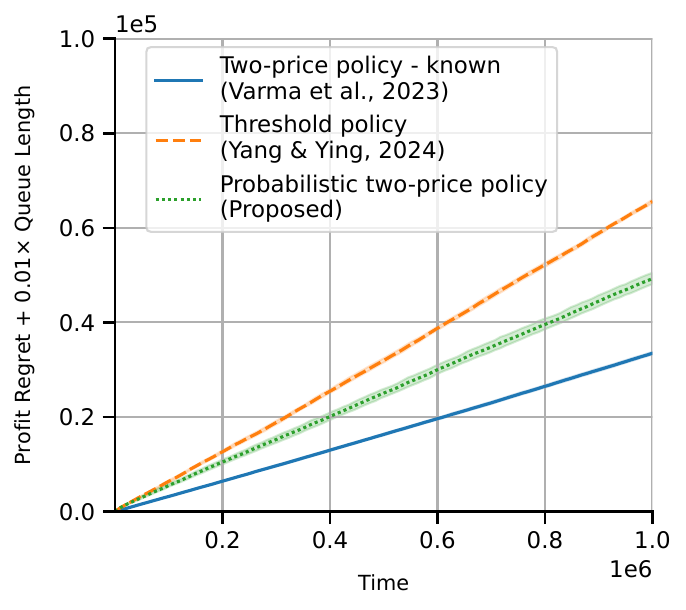}
    }
    \subfloat[Percentage improvement over the \emph{threshold policy} \cite{yang2024learning}.]{
    \includegraphics[width=0.53\linewidth]{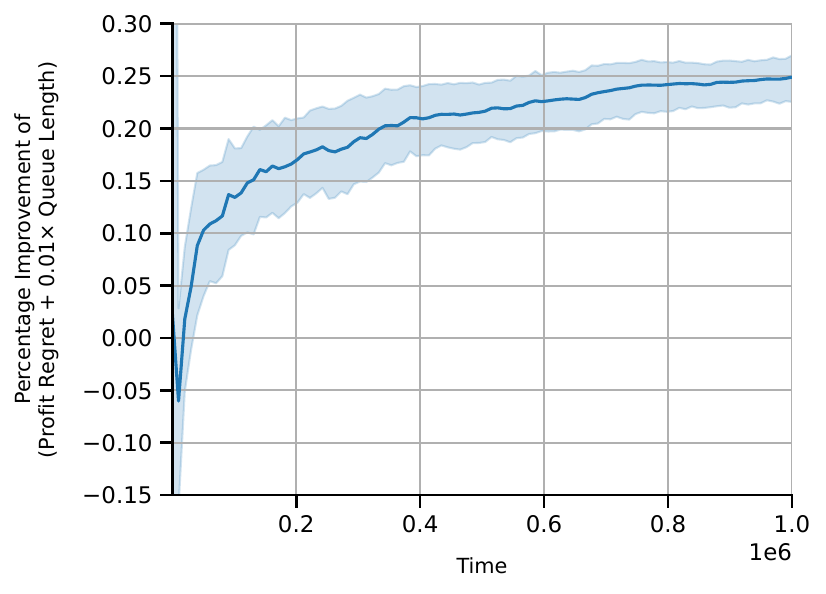}
    }
    \caption{Comparison among the \emph{two-price policy} (no learning, known demand and supply functions) \cite{varma2023dynamic}, the \emph{threshold policy} \cite{yang2024learning}, and the proposed \emph{probabilistic two-price policy} with $w=0.01$, in a single-link system.}
    \label{fig:comp-0.01}
\end{figure}

Figure~\ref{fig:comp-0.01} present the comparison results with a different weight, $w=0.01$, on the holding cost. We observe that the proposed \emph{probabilistic two-price policy} significantly improves upon the \emph{threshold policy} by up to $25\%$.

\begin{figure}[ht]
    \centering
    \subfloat[The profit regret.]{
    \includegraphics[width=0.46\linewidth]{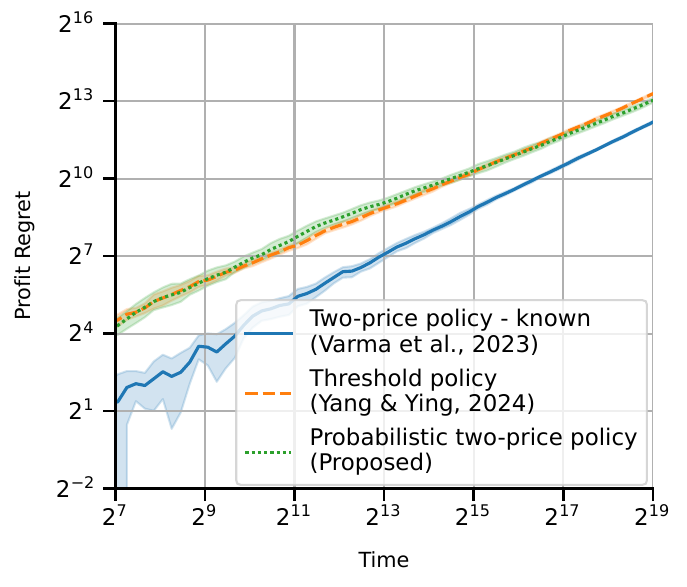}
    }
    \subfloat[The average queue length.]{
    \includegraphics[width=0.51\linewidth]{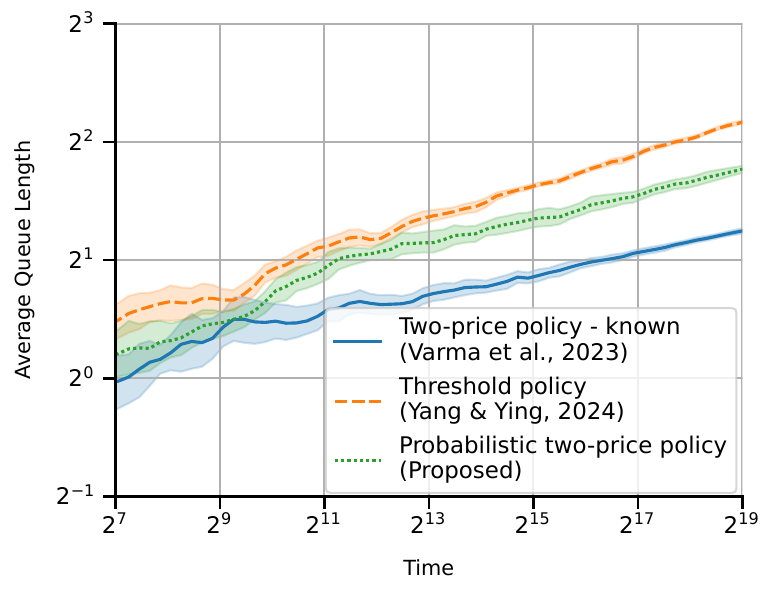}
    }
    \hfill
    \subfloat[The maximum queue length.]{
    \includegraphics[width=0.49\linewidth]{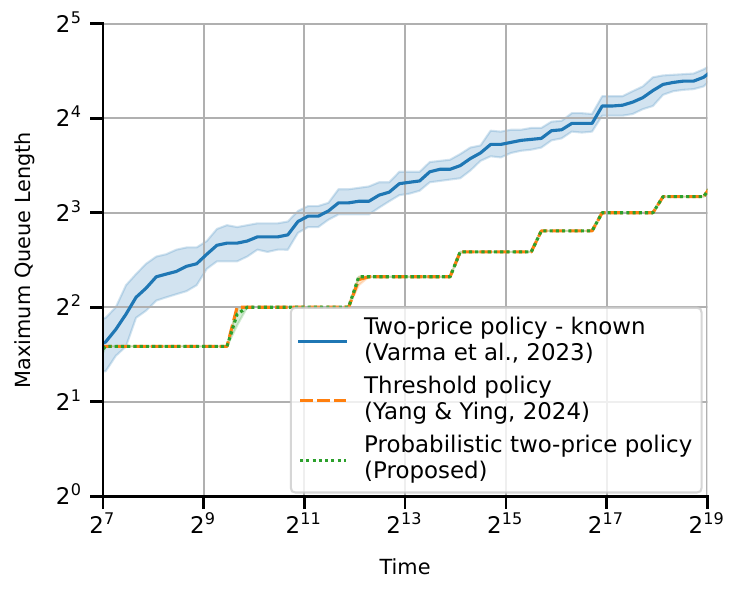}
    }
    \caption{Comparison of regret, average queue length, and maximum queue length among the \emph{two-price policy} (no learning, known demand and supply functions) \cite{varma2023dynamic}, the \emph{threshold policy} \cite{yang2024learning}, and the proposed \emph{probabilistic two-price policy}, in a single-link system.}
    \label{fig:simu-single-link}
\end{figure}

The regret, the average queue length, and the maximum queue length of the algorithms are shown in Figure~\ref{fig:simu-single-link}.
Note that the X-axis and Y-axis of Figure~\ref{fig:simu-single-link} are in log-scale. 
The regret performance of the proposed \textit{probabilistic two-price policy} is similar to the \textit{threshold policy}~\cite{yang2024learning}. However, we can see that the average queue length of the proposed \textit{probabilistic two-price policy} is significantly better than the \textit{threshold policy} because we use a novel probabilistic two-price method to reduce the average queue length without hurting the regret performance. 
The slope of the proposed \textit{probabilistic two-price policy} is significantly smaller than that of the \textit{threshold policy}~\cite{yang2024learning}, which is consistent with our theoretical results. 
As for the maximum queue length, the performance of the proposed \textit{probabilistic two-price policy} and that of the \textit{threshold policy} are the same, because both uses a threshold to reject the arrivals when the queue length exceeds the threshold. The \textit{two-price policy} has a larger maximum queue length because it does not use such a hard threshold.

\subsubsection{Tradeoff Between Regret and Queue Length}
\begin{figure}[ht]
    \centering
    \includegraphics[width=0.6\linewidth]{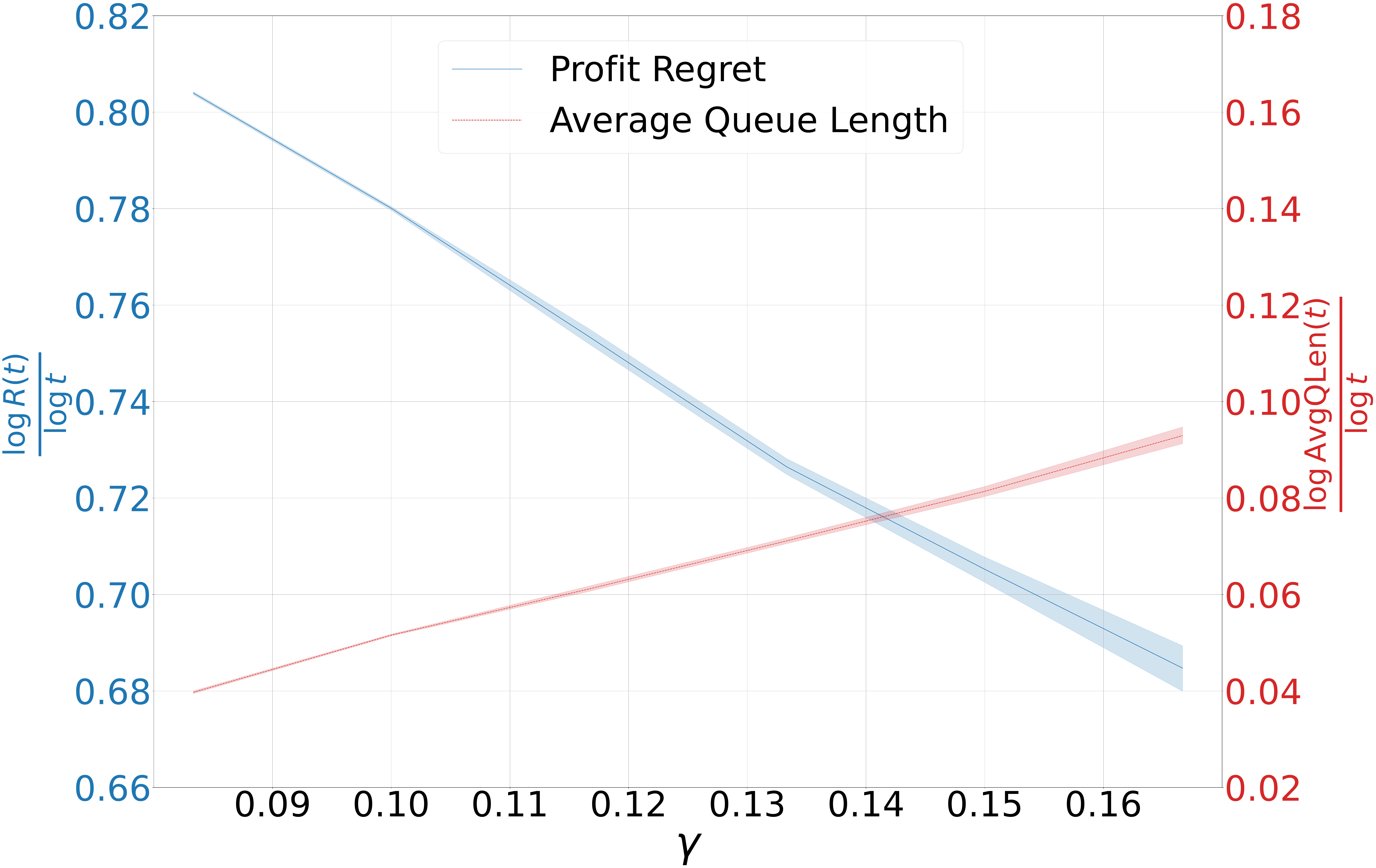}
    \caption{Tradeoff between regret and average queue length.}
    \label{fig:simu-tradeoff}
\end{figure}

Figure~\ref{fig:simu-tradeoff} shows the tradeoff between profit regret and average queue length for the proposed \textit{probabilistic two-price policy} as we change the parameter $\gamma$ from $1/12$ to $1/6$. Each point in the blue curve (regret) is generated by first computing the ratio $\frac{\log_2 R(t)}{\log_2 t}$, which represents the growth order of regret with respect to time, for each time slot $t\in[10^5, 10^6]$, and then averaging over these time slots. Similarly, each point in the red curve (average queue length) is obtained by computing $\frac{\log_2 \text{AvgQLen}(t)}{\log_2 t}$ and averaging over the same time slots.
We observe that as $\gamma$ increases, the growth order of regret decreases while the growth order of average queue length increases, confirming the tradeoff predicted by our theoretical results. By a simple least square fitting, we can obtain the slopes of the curves in Figure~\ref{fig:simu-tradeoff}. The fitted function of the blue curve (regret) is $y = -1.484 x + 0.927$ and the fitted function of the red curve (average queue length) is $y = 0.615x - 0.011$, which implies that the regret is $T^{0.927-1.484\gamma}$ and the average queue length is $T^{0.615\gamma - 0.011}$.
Although they are not exactly equal to $T^{1-\gamma}$ and $T^{0.5\gamma}$, $-1.484/0.615\approx -2.4$, which is similar to $-1/0.5=-2$. This result confirms the near-optimal tradeoff, $\tilde{O}(T^{1-\gamma})$ versus $\tilde{O}(T^{\gamma/2})$, in Corollary~\ref{cor:7}.

\subsubsection{Sensitivity to Parameter Tuning}

\begin{figure}[ht]
    \centering
    \subfloat[The profit regret.]{
    \includegraphics[width=0.47\linewidth]{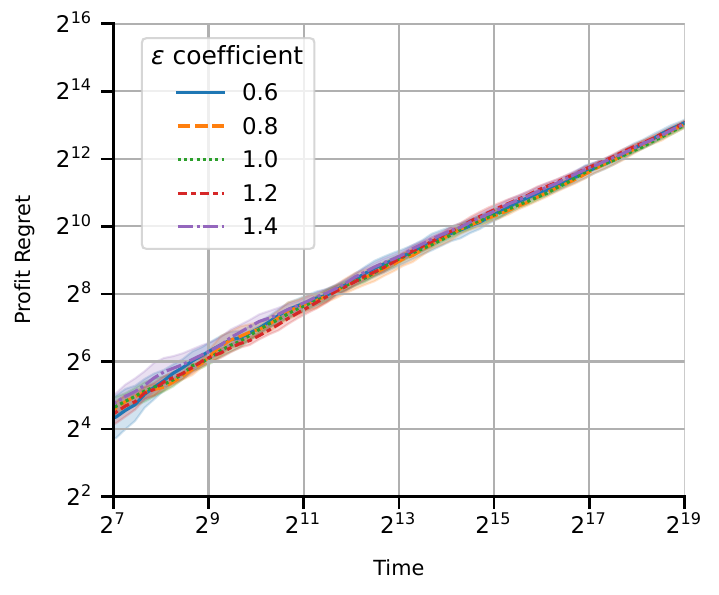}
    }
    \subfloat[The average queue length]{
    \includegraphics[width=0.51\linewidth]{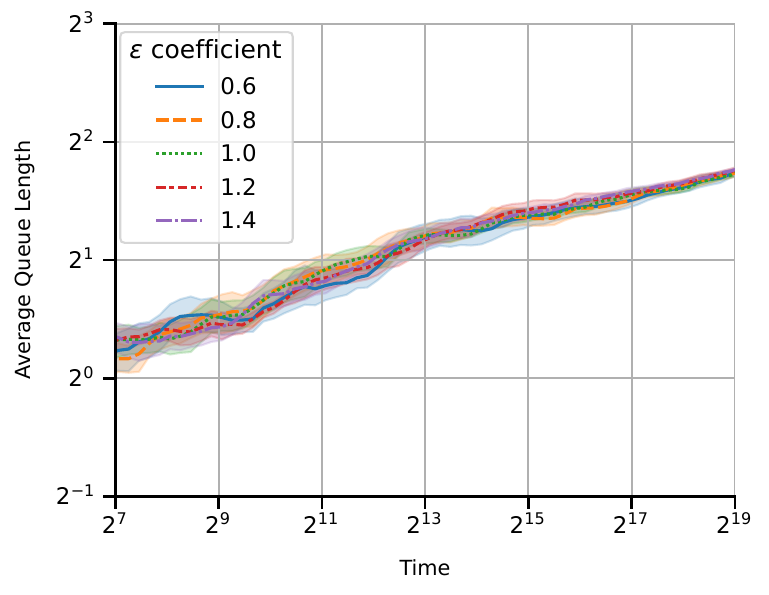}
    }
    \hfill
    \subfloat[The maximum queue length]{
    \includegraphics[width=0.49\linewidth]{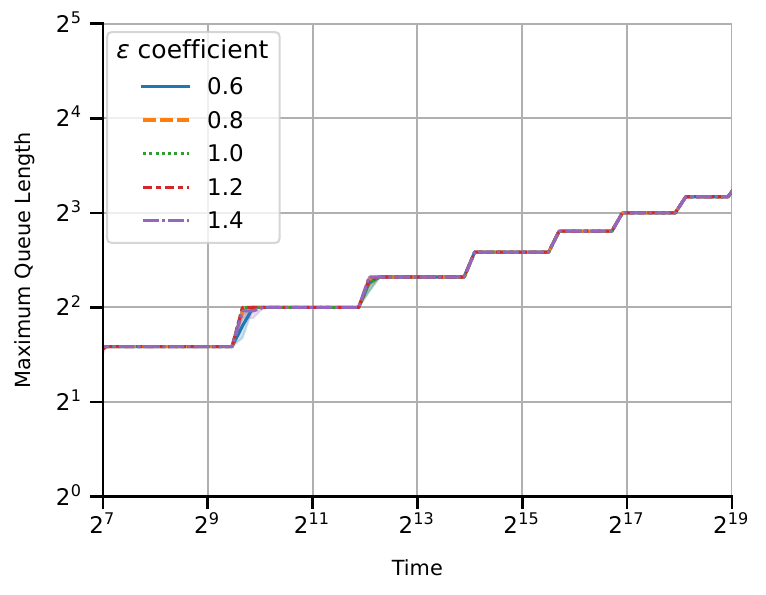}
    }
    \caption{Sensitivity testing for different $\epsilon$ for the proposed \emph{probabilistic two-price policy}.}
    \label{fig:simu-epsilon}
\end{figure}

\begin{figure}[ht]
    \centering
    \subfloat[The profit regret.]{
    \includegraphics[width=0.46\linewidth]{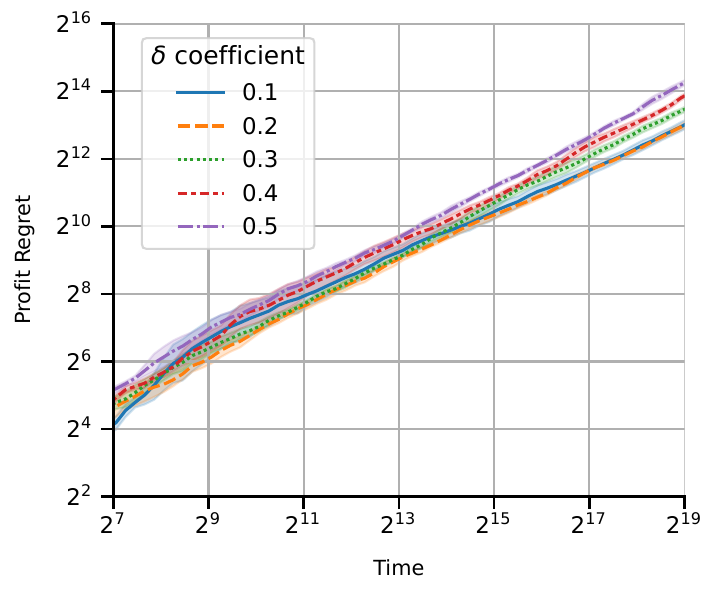}
    }
    \subfloat[The average queue length]{
    \includegraphics[width=0.51\linewidth]{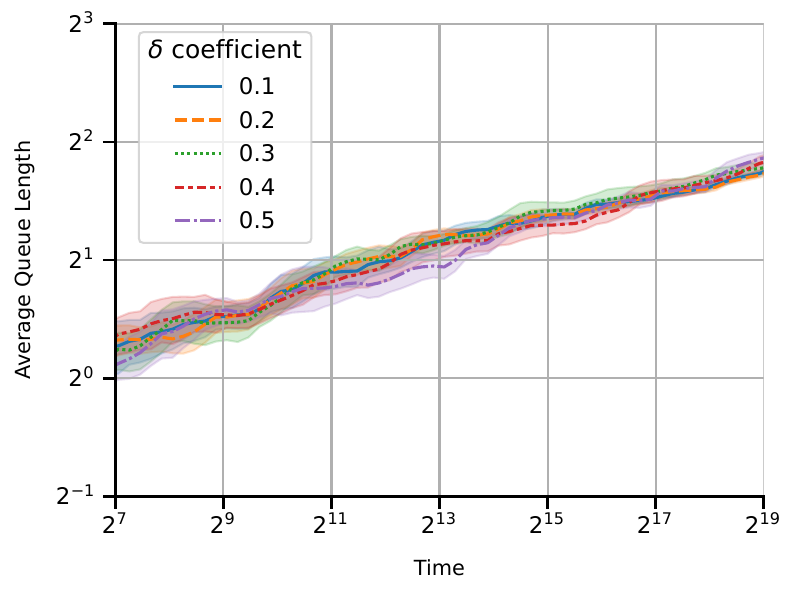}
    }
    \hfill
    \subfloat[The maximum queue length]{
    \includegraphics[width=0.49\linewidth]{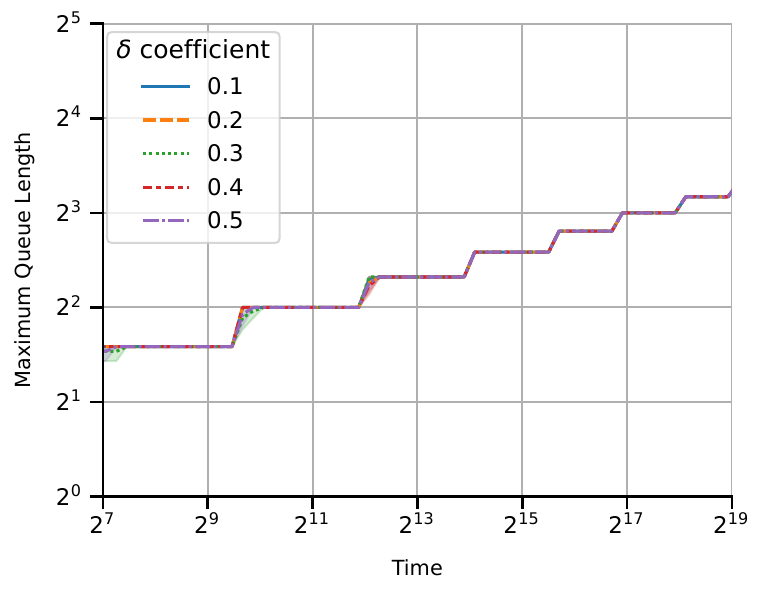}
    }
    \caption{Sensitivity testing for different $\delta$ for the proposed \emph{probabilistic two-price policy}.}
    \label{fig:simu-delta}
\end{figure}

\begin{figure}[ht]
    \centering
    \subfloat[The profit regret.]{
    \includegraphics[width=0.46\linewidth]{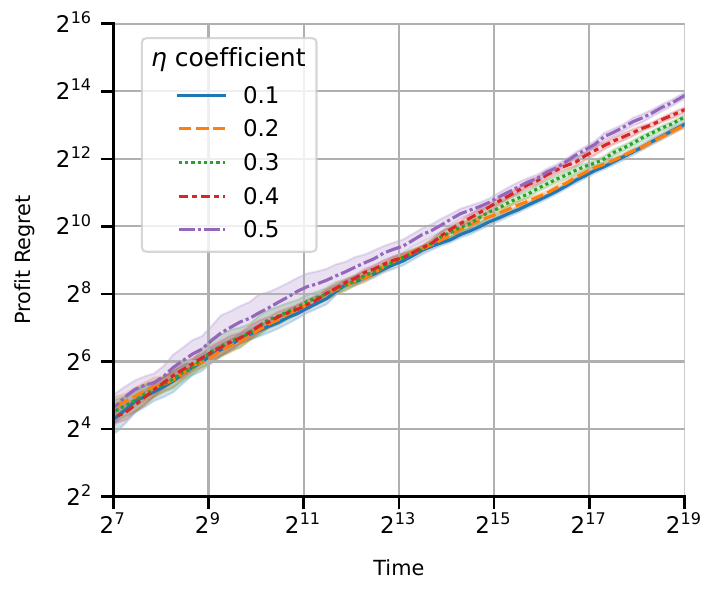}
    }
    \subfloat[The average queue length]{
    \includegraphics[width=0.51\linewidth]{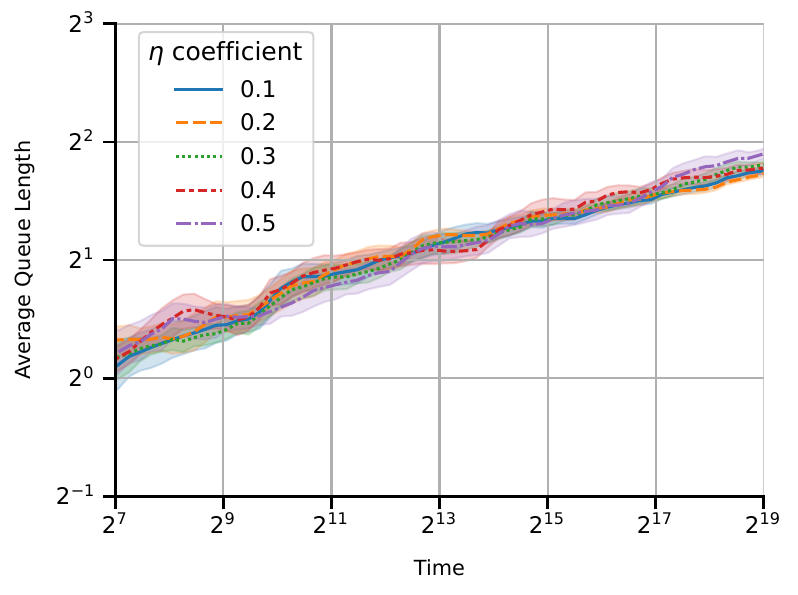}
    }
    \hfill
    \subfloat[The maximum queue length]{
    \includegraphics[width=0.49\linewidth]{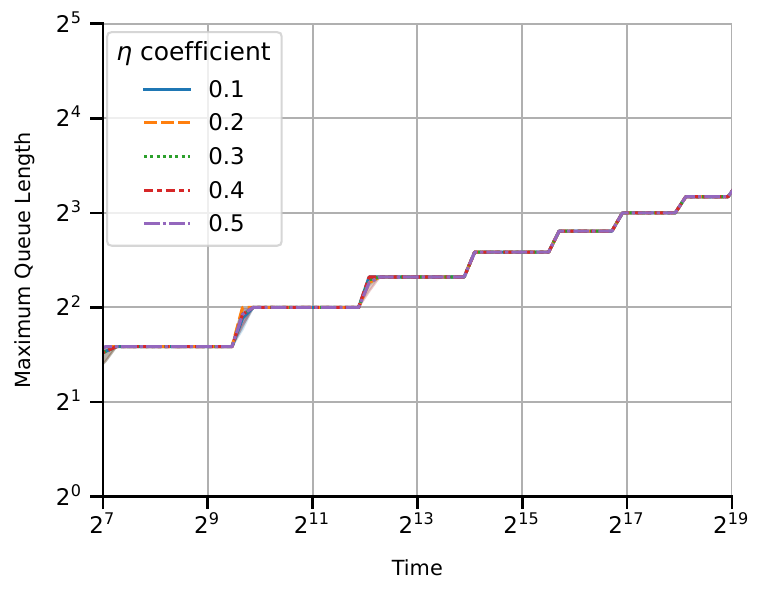}
    }
    \caption{Sensitivity testing for different $\eta$ for the proposed \emph{probabilistic two-price policy}.}
    \label{fig:simu-eta}
\end{figure}

We test the sensitivity of the performance of the proposed \textit{probabilistic two-price policy} to parameter tuning. We test the multiplicative constant of the parameters $\epsilon$, $\delta$, and $\eta$.
Figure~\ref{fig:simu-epsilon} shows the performances for different multiplicative constants in the parameter $\epsilon$, i.e., the constant $C$ for $\epsilon = C t^{-2\gamma}$. We can see that the performances are almost the same.
Figure~\ref{fig:simu-delta} shows the performances for different multiplicative constants in the parameter $\delta$, i.e., the constant $C$ for $\delta = C t^{-\gamma}$.
Figure~\ref{fig:simu-eta} shows the performances for different multiplicative constants in the parameter $\eta$, i.e., the constant $C$ for $\eta = C t^{-\gamma}$.

\subsection{Multi-Link System}

We present simulation results in a multi-link system in the following. 
\subsubsection{Setting}
Consider a system with three types of customers and three types of servers ($I=3, J=3$). The compatibility graph ${\cal E}$ is 
$
{\cal E} = \{(1,1), (1,2), (1,3),(2,1), (2,2), (3,2), (3,3)\}.
$
The demand functions are
$F_i(\lambda_i) = 2(1-\lambda_i)$ for all $i=1,2,3$.
The supply functions are
$G_j (\mu_j) = 2 \mu_j$ for all  $j=1,2,3$.
We compare the same three algorithms as in the single-link simulation, i.e., the \textit{two-price policy} \cite{varma2023dynamic}, the \textit{threshold policy} \cite{yang2024learning}, and the proposed \textit{probabilistic two-price policy}.

\subsubsection{Comparison}
For the proposed algorithm, we set $\gamma=1/6$, and $q^{\mathrm{th}}=t^{\gamma}$, $\epsilon=t^{-2\gamma}$, $\delta=0.2\times t^{-\gamma}$, $\eta=0.1\times t^{-\gamma}$, $\alpha = 0.2\times t^{\gamma/2}$, $\beta=1.0$, $e_{\mathrm{c}}= e_{\mathrm{s}}=8.0\times \max\{\delta, \eta, \epsilon\}$, $a_{\min} = 0.01$, which are of the same order as in Corollary~\ref{cor:7}.
The common parameters of the three algorithm are set to be equal for fair comparison.

Consider the same objective as that in the single-link simulation.

\begin{figure}[ht]
    \centering
    \subfloat[\texorpdfstring{$\expt[R(t)] + w t \text{AvgQLen}(t)$}{E[R(t)] + w t AvgQLen(t)}.]{
    \includegraphics[width=0.47\linewidth]{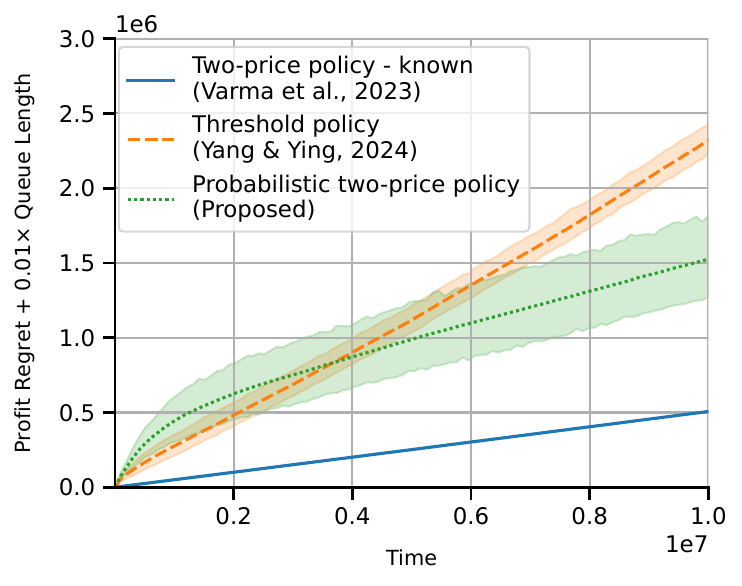}
    }
    \subfloat[Percentage improvement over the \emph{threshold policy} \cite{yang2024learning}.]{
    \includegraphics[width=0.505\linewidth]{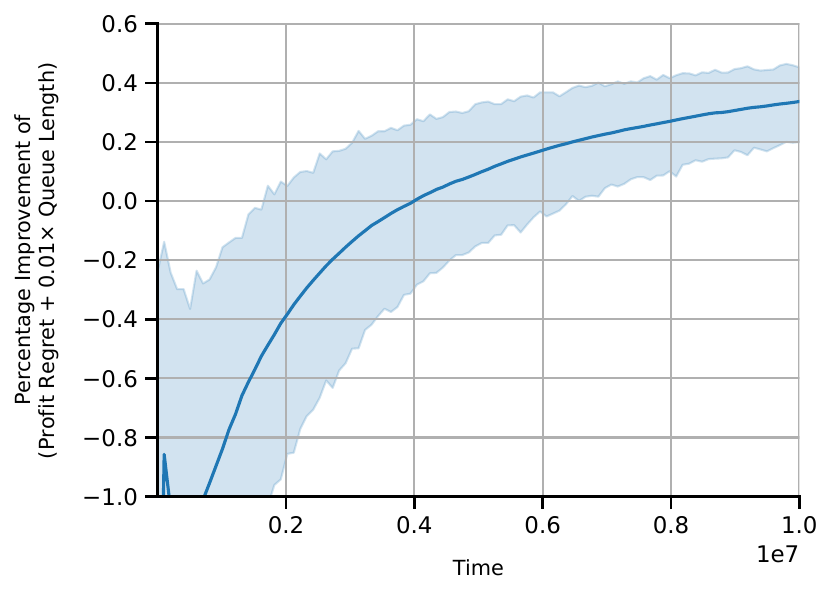}
    }
    \caption{Comparison among the \emph{two-price policy} (no learning, known demand and supply functions) \cite{varma2023dynamic}, the \emph{threshold policy} \cite{yang2024learning}, and the proposed \emph{probabilistic two-price policy} with $w=0.01$, in a multi-link system.}
    \label{fig:comp-0.01-multi-link}
\end{figure}

\begin{figure}[ht]
    \centering
    \subfloat[\texorpdfstring{$\expt[R(t)] + w t \text{AvgQLen}(t)$}{E[R(t)] + w t AvgQLen(t)}.]{
    \includegraphics[width=0.47\linewidth]{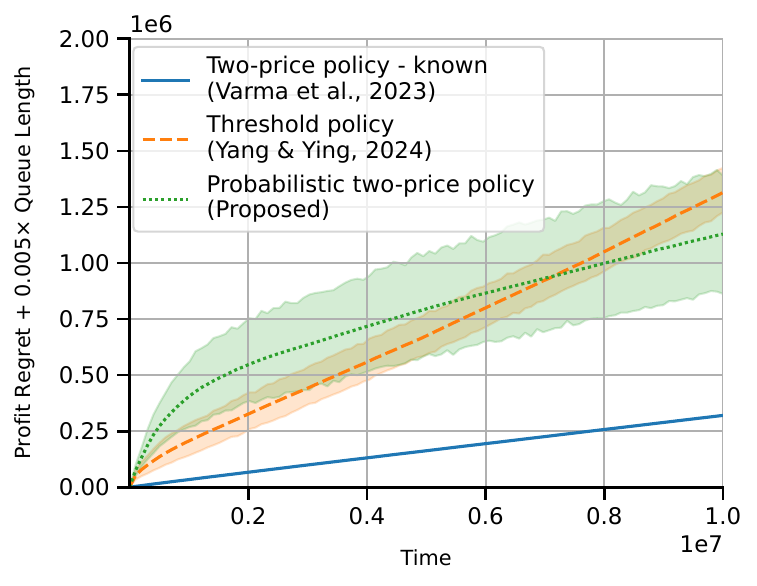}
    }
    \subfloat[Percentage improvement over the \emph{threshold policy} \cite{yang2024learning}.]{
    \includegraphics[width=0.495\linewidth]{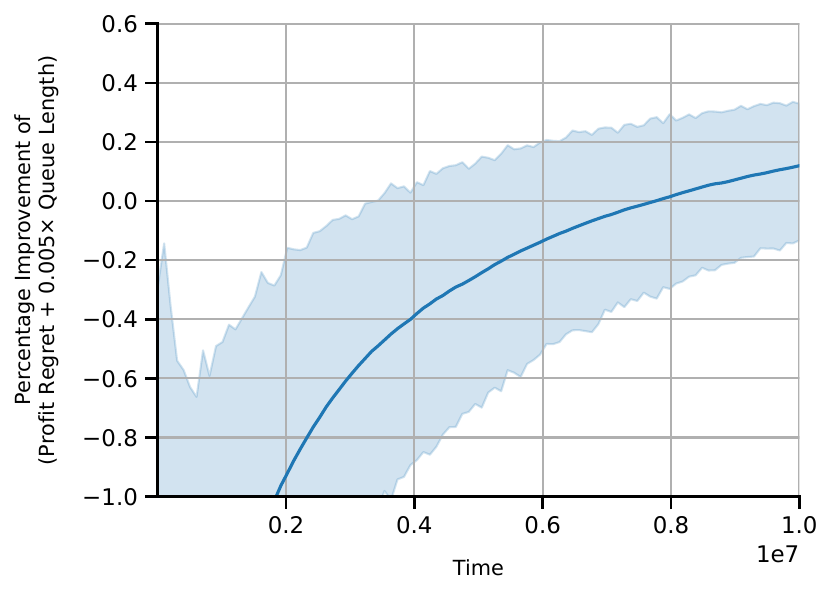}
    }
    \caption{Comparison among the \emph{two-price policy} (no learning, known demand and supply functions) \cite{varma2023dynamic}, the \emph{threshold policy} \cite{yang2024learning}, and the proposed \emph{probabilistic two-price policy} with $w=0.005$, in a multi-link system.}
    \label{fig:comp-0.005-multi-link}
\end{figure}

\begin{figure}[ht]
    \centering
    \subfloat[\texorpdfstring{$\expt[R(t)] + w t \text{AvgQLen}(t)$}{E[R(t)] + w t AvgQLen(t)}.]{
    \includegraphics[width=0.47\linewidth]{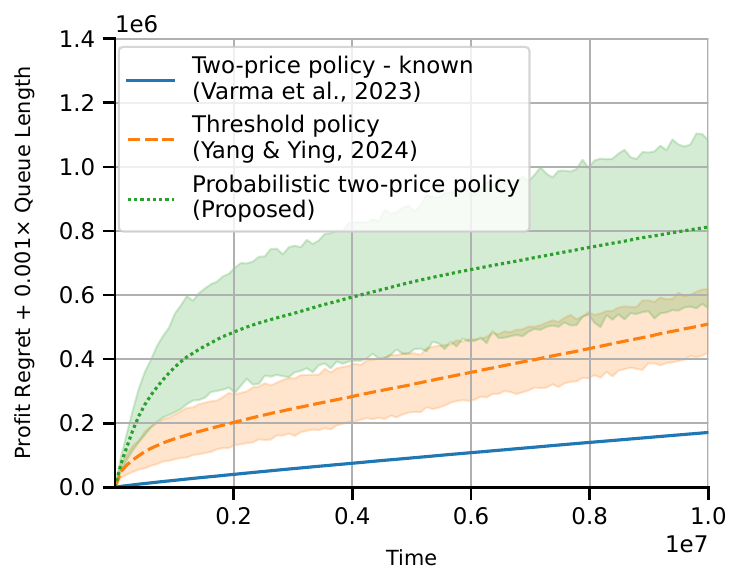}
    }
    \subfloat[Percentage improvement over the \emph{threshold policy} \cite{yang2024learning}.]{
    \includegraphics[width=0.505\linewidth]{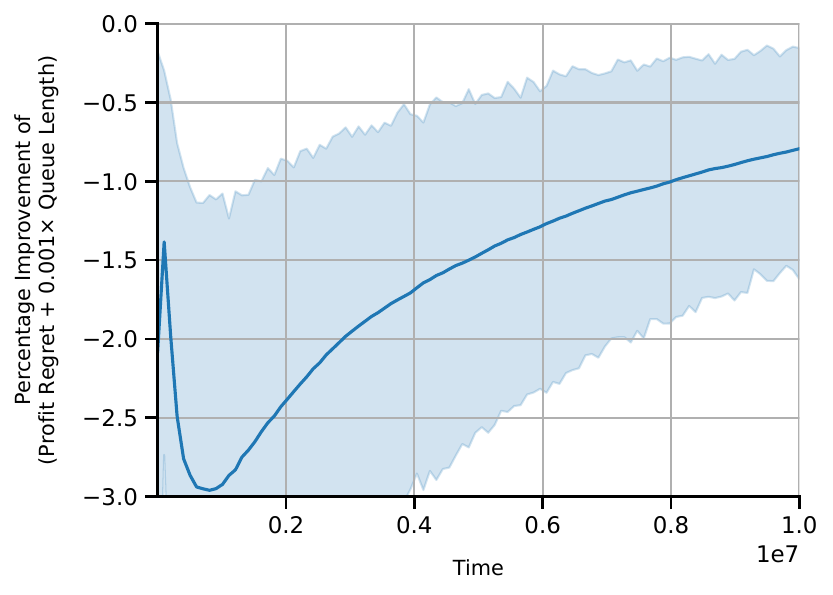}
    }
    \caption{Comparison among the \emph{two-price policy} (no learning, known demand and supply functions) \cite{varma2023dynamic}, the \emph{threshold policy} \cite{yang2024learning}, and the proposed \emph{probabilistic two-price policy} with $w=0.001$, in a multi-link system.}
    \label{fig:comp-0.001-multi-link}
\end{figure}

Figure~\ref{fig:comp-0.01-multi-link} present the comparison results with $w=0.01$, on the holding cost. We observe that the proposed \emph{probabilistic two-price policy} significantly improves upon the \emph{threshold policy} \cite{yang2024learning} by up to $44\%$ for large $t$.
Figure~\ref{fig:comp-0.005-multi-link} present the comparison results with a different weight, $w=0.005$. We observe that the proposed \emph{probabilistic two-price policy} significantly improves upon the \emph{threshold policy} by up to $30\%$ for large $t$.
We observe that in the multi-link system, initially for small $t$, the proposed \textit{probabilistic two-price policy} may not perform as good as the \textit{threshold policy}. As $t$ increases, our proposed policy quickly outperforms the \textit{threshold policy}.
This is because, as $t$ increases, the improvement in queue length performance under our proposed algorithm becomes more significant.
Figure~\ref{fig:comp-0.001-multi-link} shows the performance of these algorithms with a much smaller weight, $w=0.001$. In this case, although our proposed policy does not perform as good as the \textit{threshold policy}, it improves as $t$ increases. The comparison depends on the weight $w$ on the holding cost and the time horizon.

\begin{figure}[ht]
    \centering
    \subfloat[The profit regret.]{
    \includegraphics[width=0.50\linewidth]{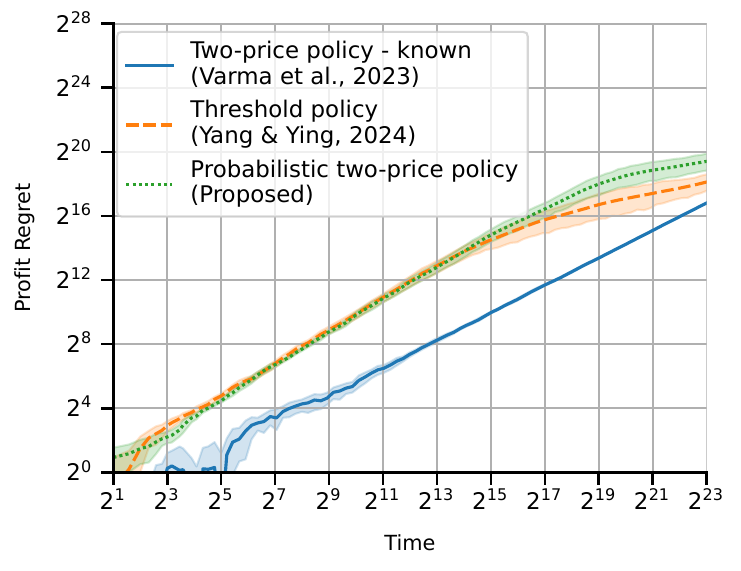}
    }
    \subfloat[The average queue length.]{
    \includegraphics[width=0.465\linewidth]{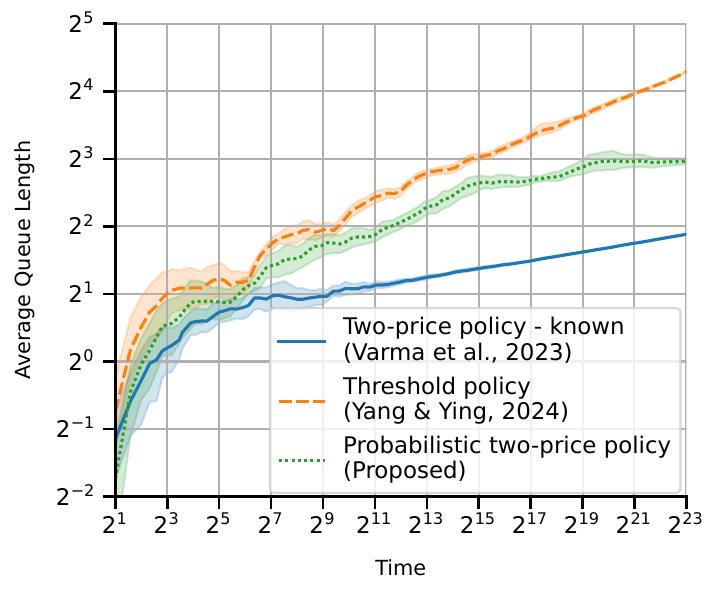}
    }
    \hfill
    \subfloat[The maximum queue length.]{
    \includegraphics[width=0.49\linewidth]{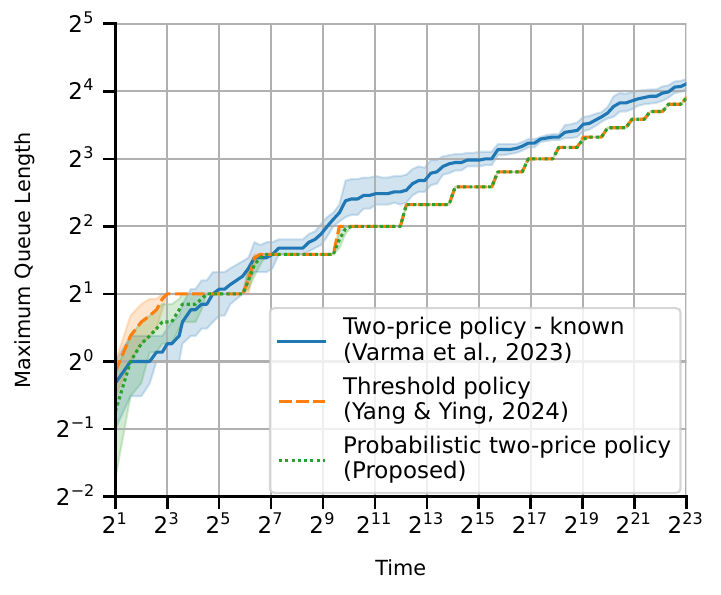}
    }
    \caption{Comparison of regret, average queue length, and maximum queue length among the \emph{two-price policy} (no learning, known demand and supply functions) \cite{varma2023dynamic}, the \emph{threshold policy} \cite{yang2024learning}, and the proposed \emph{probabilistic two-price policy}, in a multi-link system.}
    \label{fig:simu-multi-link}
\end{figure}

The regret, the average queue length, and the maximum queue length of the algorithms are shown in Figure~\ref{fig:simu-multi-link}. The observations are similar to the single-link simulation.
The regret performance of the proposed \textit{probabilistic two-price policy} is similar to the \textit{threshold policy}~\cite{yang2024learning}. However, the average queue length of the proposed \textit{probabilistic two-price policy} is significantly better than the \textit{threshold policy}.
As for the maximum queue length, the performance of the proposed \textit{probabilistic two-price policy} and that of the \textit{threshold policy} are the same. The \textit{two-price policy} has a larger maximum queue length.

\subsection{Compute Resources}
All the simulations are not compute-intensive, and all results can be reproduced on a standard personal computer. On a personal computer with Intel Core i5-9400 CPU @2.90GHz, 10 runs of the simulation of the single-link system ($T=10^6$) for all three algorithms take approximately 20 minutes, and 10 runs of the simulation of the multi-link system ($T=10^7$) for all three algorithms take approximately 6 hours. 

\clearpage
\newpage

\section{Societal Impacts}
\label{app:impact}

This paper presents work whose goal is to advance the fields of online learning in queueing systems. There are potential positive societal impacts, including improving efficiency and promoting profitability in applications like ride-hailing, meal-delivery, and crowdsourcing. We are not aware of any negative societal impacts of this work.

\clearpage
\newpage

\section{Proof of Lemma~\ref{lemma:lower-bound}}
\label{app:proof-lemma-lower-bound}

The fluid optimization problem \eqref{equ:fluid-opti-obj}-\eqref{equ:fluid-opti-constr4} for the single-link system reduces to 
\begin{align}\label{equ:one-link-fluid-opt}
    \max_{x} xF(x) - xG(x)  \quad \mathrm{s.t.} \quad x\in [0,1].
\end{align}
Let $f(x)\coloneqq xF(x) - xG(x)$. From the assumptions in Lemma~\ref{lemma:lower-bound}, we know that $\lambda^*=\mu^*=x^*\in(0,1)$ is the unique solution. Fix a $\gamma \in [0, 1/2]$.
Consider a class of pricing policies such that 
$\sqrt{\text{AvgQ$^2$Len}(T)} \le T^{\gamma/2}$ and $\expt [Q_{\mathrm{c}}(T+1)^2+Q_{\mathrm{s}}(T+1)^2] = o(T)$. Fix any policy in this class.

From the regret definition \eqref{equ:regret-def}, we have
\begin{align}\label{equ:regret-single-queue}
    \expt[R(T)] = \sum_{t=1}^{T} \left( \expt \left[
    x^* F(x^*) - \lambda(t) F(\lambda(t))
    \right]
    + \expt \left[
    \mu(t) G(\mu(t)) - x^* G(x^*) 
    \right] \right).
\end{align}
Let $r(x)\coloneqq xF(x)$. Since $-r(x)$ is strongly convex, we have \cite{boyd2004convex}:
\begin{align}\label{equ:strong-convex-r}
    r(x^*) - r(\lambda(t)) \ge r'(x^*) (x^* - \lambda(t)) + \nu_{r} (\lambda(t)-x^*)^2,
\end{align}
where $\nu_r>0$ is a constant.  Let $c(x)\coloneqq xG(x)$. Since $c(x)$ is strongly convex, we have \cite{boyd2004convex}:
\begin{align}\label{equ:strong-convex-c}
    c(\mu(t)) - c(x^*)  \ge c'(x^*) (\mu(t) - x^*) + \nu_{c} (\mu(t)-x^*)^2,
\end{align}
where $\nu_c>0$ is a constant.
From \eqref{equ:regret-single-queue}, \eqref{equ:strong-convex-r}, and \eqref{equ:strong-convex-c}, we have
\begin{align}\label{equ:lower-bound-1}
    \expt[R(T)] \ge & \sum_{t=1}^{T} \expt \left[ r'(x^*) (x^* - \lambda(t)) + \nu_{r} (\lambda(t)-x^*)^2 \right] \nonumber\\
    & + \sum_{t=1}^{T} \expt \left[ c'(x^*) (\mu(t) - x^*) + \nu_{c} (\mu(t)-x^*)^2 \right].
\end{align}
Since $x^*\in(0,1)$ is the optimal solution and $f$ is concave, we have $0 = f'(x^*)=r'(x^*)-c'(x^*)$, which implies that $r'(x^*)=c'(x^*)$. Hence, the bound \eqref{equ:lower-bound-1} can be reduced to
\begin{align}\label{equ:lower-bound-2}
    \expt[R(T)] \ge & \sum_{t=1}^{T}  \expt \left[ c'(x^*) ( \mu(t) - \lambda(t)) \right] 
    +  \sum_{t=1}^{T} \expt \left[ \nu_{c} (\mu(t)-x^*)^2   + \nu_{r} (\lambda(t)-x^*)^2 \right],
\end{align}
where $c'(x^*)\ge 0$ by Assumption~\ref{assum:1}.
From the queue dynamics, we have
\[
Q_{\mathrm{c}}(t+1) = Q_{\mathrm{c}}(t) + A_{\mathrm{c}}(t) - X(t)
\]
Taking expectation on both sides, we have
\[
\expt[Q_{\mathrm{c}}(t+1)] = \expt[Q_{\mathrm{c}}(t)] + \expt[\lambda(t)] - \expt[X(t)],
\]
where $X(t)$ is the number of matches on the link at time $t$.
Similarly, we have
\[
\expt[Q_{\mathrm{s}}(t+1)] = \expt[Q_{\mathrm{s}}(t)] + \expt[\mu(t)] - \expt[X(t)].
\]
Hence, we have
\[
\expt[Q_{\mathrm{c}}(t+1)] - \expt[Q_{\mathrm{s}}(t+1)] 
= \expt[Q_{\mathrm{c}}(t)] - \expt[Q_{\mathrm{s}}(t)]
+ \expt[\lambda(t) - \mu(t)].
\]
Define $Q(t)\coloneqq Q_{\mathrm{c}}(t) - Q_{\mathrm{s}}(t)$. Then, we have
\[
\expt[Q(t+1)] = \expt[Q(t)] + \expt[\lambda(t) - \mu(t)].
\]
Summing both sides from $t=1$ to $t=T$ and noting that $Q(1)=Q_{\mathrm{c}}(1) - Q_{\mathrm{s}}(1)=0$, we obtain
\begin{align}\label{equ:sum-arr-rate-eq-queue-len}
    \expt[Q(T+1)] = \sum_{t=1}^{T} \expt[\lambda(t) - \mu(t)].
\end{align}
From \eqref{equ:lower-bound-2} and \eqref{equ:sum-arr-rate-eq-queue-len}, we obtain
\begin{align}\label{equ:lower-bound-3}
    \expt[R(T)] \ge &  - c'(x^*) \expt[Q(T+1)] 
    +  \sum_{t=1}^{T} \expt \left[ \nu_{c} (\mu(t)-x^*)^2   + \nu_{r} (\lambda(t)-x^*)^2 \right]\nonumber\\
    \ge & - c'(x^*) |\expt[Q(T+1)]|
    +  \sum_{t=1}^{T} \expt \left[ \nu_{c} (\mu(t)-x^*)^2   + \nu_{r} (\lambda(t)-x^*)^2 \right],
\end{align}
By Jensen's inequality, we have
\begin{align}\label{equ:proof-lower-bound-q-T}
    |\expt[Q(T+1)]| \le \expt[|Q(T+1)|] \le \sqrt{\expt[|Q(T+1)|^2]} = \sqrt{\expt[Q_{\mathrm{c}}(T+1)^2 + Q_{\mathrm{s}}(T+1)^2]},
\end{align}
where the last equality is due to the fact that either $Q_{\mathrm{c}}(T+1)=0$ or $Q_{\mathrm{s}}(T+1)=0$ by Lemma~1 in~\cite{yang2024learning}. By the definition of the policy class and \eqref{equ:proof-lower-bound-q-T}, we have
\[
|\expt[Q(T+1)]| = o(\sqrt{T}).
\]
Hence, \eqref{equ:lower-bound-3} can be further bounded by
\begin{align}\label{equ:lower-bound-4}
    \expt[R(T)] \ge & - o(\sqrt{T})
    +  \sum_{t=1}^{T} \expt \left[ \nu_{c} (\mu(t)-x^*)^2   + \nu_{r} (\lambda(t)-x^*)^2 \right]\nonumber\\
    \ge & 
    - o(\sqrt{T})
    + \frac{\min\{\nu_{c}, \nu_{r}\}}{2} \sum_{t=1}^{T} \expt \left[ \left(|\mu(t)-x^*|   + |\lambda(t)-x^*| \right)^2 \right]
    \nonumber\\
    = & - o(\sqrt{T})
    + \frac{\min\{\nu_{c}, \nu_{r}\}}{4} \sum_{t=1}^{T} \expt \left[ \left(|\mu(t)-x^*|   + |\lambda(t)-x^*| \right)^2 \right]\nonumber\\
    & + \frac{\min\{\nu_{c}, \nu_{r}\}}{4} \sum_{t=1}^{T} \expt \left[ \left(|\mu(t)-x^*|   + |\lambda(t)-x^*| \right)^2 \right]
    \nonumber\\
    \ge & - o(\sqrt{T})
    + \frac{\min\{\nu_{c}, \nu_{r}\}}{4} \sum_{t=1}^{T} \expt \left[ \left(\mu(t)-\lambda(t) \right)^2 \right]\nonumber\\
    & + \frac{\min\{\nu_{c}, \nu_{r}\}}{4} \sum_{t=1}^{T} \expt \left[ \left(|\mu(t)-x^*|   + |\lambda(t)-x^*| \right)^2 \right]
    \nonumber\\
    \ge & - o(\sqrt{T})
    + \frac{\min\{\nu_{c}, \nu_{r}\}}{4} \sum_{t=1}^{T} \expt \left[ \left(\mu(t)-\lambda(t) \right)^2 \right]\nonumber\\
    & + \frac{\min\{\nu_{c}, \nu_{r}\}}{4} \sum_{t=1}^{T}  \left(\expt[|\mu(t)-x^*|+|\lambda(t)-x^*|] \right)^2\nonumber\\
    \ge & - o(\sqrt{T})
    + \frac{\min\{\nu_{c}, \nu_{r}\}}{4} \sum_{t=1}^{T} \expt \left[ \left(\mu(t)-\lambda(t) \right)^2 \right]\nonumber\\
    & + \frac{\min\{\nu_{c}, \nu_{r}\}}{4T} \left(\sum_{t=1}^{T}  \expt[|\mu(t)-x^*|+|\lambda(t)-x^*|] \right)^2
\end{align}
where the second inequality uses the fact that $a^2 + b^2 \ge  (a+b)^2/2$, the third inequality follows from the triangle inequality, the fourth inequality follows from Jensen's inequality, and the last inequality follows from Cauchy-Schwarz inequality.
Next, we will bound the term $\sum_{t=1}^{T} \expt [ (\mu(t)-\lambda(t) )^2 ]$ in \eqref{equ:lower-bound-4}  using the Lyapunov drift method. Consider the drift $\expt[Q^2(t+1)-Q^2(t)]$. By the queue dynamics, we can obtain
\begin{align}\label{equ:drift-square-q}
    \expt[Q^2(t+1)-Q^2(t)] = & \expt[(Q(t) + A_{\mathrm{c}}(t) - A_{\mathrm{s}}(t))^2-Q^2(t)]\nonumber\\
    = & \expt[(A_{\mathrm{c}}(t) - A_{\mathrm{s}}(t))^2] + 2\expt[ Q(t) \left(A_{\mathrm{c}}(t) - A_{\mathrm{s}}(t)\right)]\nonumber\\
    = & \expt[(A_{\mathrm{c}}(t) - A_{\mathrm{s}}(t))^2] + 2\expt[  Q(t) \expt[A_{\mathrm{c}}(t) - A_{\mathrm{s}}(t) | Q(t), \lambda(t), \mu(t)] ]\nonumber\\
    = & \expt[(A_{\mathrm{c}}(t) - A_{\mathrm{s}}(t))^2] + 2\expt[ Q(t)  \left(\lambda(t) - \mu(t)\right) ]
\end{align}
where the third line uses the law of iterated expectation. Summing both sides from $t=1$ to $t=T$, we obtain
\begin{align}\label{equ:lower-bound-4-drift}
    \expt[Q^2(T+1)] = & \sum_{t=1}^{T} \expt[(A_{\mathrm{c}}(t) - A_{\mathrm{s}}(t))^2] 
    + 2 \expt\left[\sum_{t=1}^T Q(t)  \left(\lambda(t) - \mu(t)\right) \right]\nonumber\\
    \ge & \sum_{t=1}^{T} \expt[(A_{\mathrm{c}}(t) - A_{\mathrm{s}}(t))^2] 
    - 2 \expt\left[\sum_{t=1}^T |Q(t)|  |\lambda(t) - \mu(t)| \right].
\end{align}
For the variance term $\expt[(A_{\mathrm{c}}(t) - A_{\mathrm{s}}(t))^2]$ in \eqref{equ:lower-bound-4-drift}, we have
\begin{align*}
    \expt[(A_{\mathrm{c}}(t) - A_{\mathrm{s}}(t))^2] 
    = & \expt \left[ A_{\mathrm{c}}^2(t) \right] + \expt \left[ A_{\mathrm{s}}^2(t) \right] - 2 \expt\left[A_{\mathrm{c}}(t)   A_{\mathrm{s}}(t) \right]\nonumber\\
    = &  \expt \left[ \expt \left[ A_{\mathrm{c}}^2(t) | \lambda(t) \right] \right]
    + \expt \left[  \expt \left[A_{\mathrm{s}}^2(t) | \mu(t) \right] \right] 
    - 2 \expt\left[ \expt \left[ A_{\mathrm{c}}(t)   A_{\mathrm{s}}(t) |  \lambda(t), \mu(t) \right]\right]\nonumber\\
    = & \expt\left[ \lambda(t) + \mu(t) - 2 \lambda(t)\mu(t) \right],
\end{align*}
where the second line uses the law of iterated expectation and the last line uses the independence between $A_{\mathrm{c}}(t)$ and $A_{\mathrm{s}}(t)$ conditioned on $\lambda(t)$ and $\mu(t)$. Adding and subtracting $x^*$ and $2x^*\mu(t)$, we obtain
\begin{align*}
    \expt[(A_{\mathrm{c}}(t) - A_{\mathrm{s}}(t))^2]
    = & \expt\left[ \lambda(t) - x^* + x^* + \mu(t) - 2 x^* \mu(t) + 2 (x^* - \lambda(t))\mu(t) \right]\nonumber\\
    = & \expt\left[ x^* + \mu(t) - 2 x^* \mu(t) \right] + \expt\left[ \lambda(t) - x^* + 2 (x^* - \lambda(t))\mu(t) \right]\nonumber\\
    \ge & \expt\left[ x^* + \mu(t) - 2 x^* \mu(t) \right] - 3 \expt\left[ |\lambda(t) - x^*|\right].
\end{align*}
Adding and subtracting $x^*$ and $2(x^*)^2$, we further obtain
\begin{align}\label{equ:lower-bound-4-variance}
    \expt[(A_{\mathrm{c}}(t) - A_{\mathrm{s}}(t))^2]
    \ge &  x^* + x^* - 2(x^*)^2 +\expt\left[\mu(t)-x^* + 2x^* (x^* -\mu(t)) \right] - 3 \expt\left[ |\lambda(t) - x^*|\right]\nonumber\\
    \ge &  2x^* (1 - x^*)
    - 3\expt\left[ |\mu(t) - x^*|\right]
    - 3 \expt\left[ |\lambda(t) - x^*|\right].
\end{align}
Hence, from \eqref{equ:lower-bound-4-drift} and \eqref{equ:lower-bound-4-variance}, we have
\begin{align*}
    \expt[Q^2(T+1)] \ge & 2 T x^* (1 - x^*)
    - 3 \sum_{t=1}^{T} \expt\left[ |\mu(t) - x^*|
    +|\lambda(t) - x^*|\right]\nonumber\\
    & - 2 \expt\left[\sum_{t=1}^T |Q(t)|  |\lambda(t) - \mu(t)| \right].
\end{align*}
Rearranging terms, we obtain
\begin{align}\label{equ:lower-bound-4-drift-begin}
    \expt\left[\sum_{t=1}^T |Q(t)|  |\lambda(t) - \mu(t)| \right] \ge & T x^* (1 - x^*)
    - \frac{3}{2} \sum_{t=1}^{T} \expt\left[ |\mu(t) - x^*|
    +|\lambda(t) - x^*|\right]\nonumber\\
    & - \frac{\expt[Q^2(T+1)]}{2}
\end{align}
By the definition of the policy class, we know $\expt[Q^2(T+1)] = \expt[Q_{\mathrm{c}}(T+1)^2 + Q_{\mathrm{s}}(T+1)^2] = o(T)$. Hence, we have
\[
    \expt\left[\sum_{t=1}^T |Q(t)|  |\lambda(t) - \mu(t)| \right] \ge T x^* (1 - x^*)
    - \frac{3}{2} \sum_{t=1}^{T} \expt\left[ |\mu(t) - x^*|
    +|\lambda(t) - x^*|\right]
    - o(T).
\]
Since $x^*\in(0,1)$, we have $2x^* (1 - x^*)>0$. Hence, $T x^* (1 - x^*) = \Theta(T)$. Hence, for sufficiently large $T$, we have
\begin{align}\label{equ:lower-bound-4-drift-final-1}
    \expt\left[\sum_{t=1}^T |Q(t)|  |\lambda(t) - \mu(t)| \right] \ge \Theta(T)
    - \frac{3}{2} \sum_{t=1}^{T} \expt\left[ |\mu(t) - x^*|
    +|\lambda(t) - x^*|\right].
\end{align}
For the left-hand-side of \eqref{equ:lower-bound-4-drift-final-1}, by Cauchy-Schwarz inequality, we have
\begin{align}\label{equ:lower-bound-4-drift-final-2}
    \expt\left[\sum_{t=1}^T |Q(t)|  |\lambda(t) - \mu(t)| \right] 
    \le & \expt\left[
    \sqrt{\sum_{t=1}^T |Q(t)|^2} \sqrt{ \sum_{t=1}^T |\lambda(t) - \mu(t)|^2} \right]\nonumber\\
    \le & \sqrt{ \expt\left[
    \sum_{t=1}^T |Q(t)|^2\right] \expt\left[ \sum_{t=1}^T |\lambda(t) - \mu(t)|^2 \right]}.
\end{align}
By the definition of the policy class, we know that
\begin{align}\label{equ:lower-bound-4-drift-final-3}
    \expt\left[
    \sum_{t=1}^T |Q(t)|^2\right]=
    \sum_{t=1}^T \expt\left[Q_{\mathrm{c}}(t)^2 + Q_{\mathrm{s}}(t)^2\right] = T \text{AvgQ$^2$Len}(T) \le T^{1+\gamma},
\end{align}
where the first equality is by Lemma 1 in \cite{yang2024learning}.
From \eqref{equ:lower-bound-4-drift-final-2} and \eqref{equ:lower-bound-4-drift-final-3}, we have
\begin{align}\label{equ:lower-bound-4-drift-final-4}
    \expt\left[\sum_{t=1}^T |Q(t)|  |\lambda(t) - \mu(t)| \right] 
    \le & T^{\frac{1}{2}+\frac{\gamma}{2}} \sqrt{ \expt\left[ \sum_{t=1}^T |\lambda(t) - \mu(t)|^2 \right]}.
\end{align}
From \eqref{equ:lower-bound-4-drift-final-1} and \eqref{equ:lower-bound-4-drift-final-4}, we have
\begin{align}\label{equ:lower-bound-4-drift-final}
     \sqrt{ \expt\left[ \sum_{t=1}^T |\lambda(t) - \mu(t)|^2 \right]} \ge \Theta(T^{\frac{1}{2}-\frac{\gamma}{2}})
     - \frac{3}{2} T^{-\frac{1}{2}-\frac{\gamma}{2}} \sum_{t=1}^{T} \expt\left[ |\mu(t) - x^*|
    +|\lambda(t) - x^*|\right].
\end{align}
From \eqref{equ:lower-bound-4} and \eqref{equ:lower-bound-4-drift-final} and noting that $\sqrt{\expt\left[ \sum_{t=1}^T |\lambda(t) - \mu(t)|^2 \right]}\ge 0$ , we have
\begin{align}\label{equ:lower-bound-5}
    \expt[R(T)]
    \ge & 
    \frac{\min\{\nu_{c}, \nu_{r}\}}{4} \left(\max\left\{0, 
    \Theta(T^{\frac{1}{2}-\frac{\gamma}{2}})
     - \frac{3}{2} T^{-\frac{1}{2}-\frac{\gamma}{2}} \sum_{t=1}^{T}         
    \expt\left[ |\mu(t) - x^*|
    +|\lambda(t) - x^*|\right]\right\}
    \right)^2 \nonumber\\
    & + \frac{\min\{\nu_{c}, \nu_{r}\}}{4T} \left(\sum_{t=1}^{T}  \expt[|\mu(t)-x^*|+|\lambda(t)-x^*|] \right)^2 - 
    o(\sqrt{T})\nonumber\\
    \ge & \frac{\min\{\nu_{c}, \nu_{r}\}}{4} \left(\max\left\{0, 
    \Theta(T^{\frac{1}{2}-\frac{\gamma}{2}})
     - \frac{3}{2} T^{-\frac{1}{2}-\frac{\gamma}{2}} \sum_{t=1}^{T}         
    \expt\left[ |\mu(t) - x^*|
    +|\lambda(t) - x^*|\right]\right\}
    \right)^2 \nonumber\\
    & + \frac{\min\{\nu_{c}, \nu_{r}\}}{4T} \left(\sum_{t=1}^{T}  \expt[|\mu(t)-x^*|+|\lambda(t)-x^*|] \right)^2 - 
    o(\sqrt{T})
\end{align}
Note that the term $\sum_{t=1}^{T}  \expt[|\mu(t)-x^*|+|\lambda(t)-x^*|]$ in \eqref{equ:lower-bound-5} takes values in $[0,2T]$. Consider two cases:
\begin{itemize}[leftmargin=*]
    \item $\sum_{t=1}^{T}  \expt[|\mu(t)-x^*|+|\lambda(t)-x^*|]=\Theta(T)$:
    In this case, from \eqref{equ:lower-bound-5}, we have
    \begin{align*}
        \expt[R(T)] \ge & \frac{\min\{\nu_{c}, \nu_{r}\}}{4T} \left(\sum_{t=1}^{T}  \expt[|\mu(t)-x^*|+|\lambda(t)-x^*|] \right)^2 - 
        o(\sqrt{T}) \nonumber\\
        \ge & \Theta(T) - o(\sqrt{T}) = \Theta(T).
    \end{align*}
    \item $\sum_{t=1}^{T}  \expt[|\mu(t)-x^*|+|\lambda(t)-x^*|]=o(T)$:
    In this case, from \eqref{equ:lower-bound-5}, we have
    \begin{align*}
        \expt[R(T)] \ge & \frac{\min\{\nu_{c}, \nu_{r}\}}{4} \left(\max\left\{0, 
    \Theta(T^{\frac{1}{2}-\frac{\gamma}{2}})
     - o(T^{\frac{1}{2}-\frac{\gamma}{2}})\right\}
    \right)^2
    -  o(\sqrt{T}) \nonumber\\
    =  & \Theta( T^{1-\gamma} )-  o(\sqrt{T})  = \Theta(T^{1-\gamma}),
    \end{align*}
    where the last equality is by $\gamma\le 1/2$.
\end{itemize}
Hence, we have
\[
\expt[R(T)] = \Omega (T^{1-\gamma}).
\]
Lemma~\ref{lemma:lower-bound} is proved.

\clearpage
\newpage

\section{
\texorpdfstring{Proof of $\sqrt{\text{AvgQ$^2$Len}(T)} = \Theta(\text{AvgQLen}(T))$ for a Class of Policies}{Proof of the Condition in Lemma~\ref{lemma:lower-bound} for a Class of Policies}
}
\label{app:proof-two-price-satisfy-cond}

For the single-link system, from the assumptions in Lemma~\ref{lemma:lower-bound}, we know that $\lambda^*=\mu^*=x^*\in(0,1)$ is the unique solution.
Consider a class of pricing policies that satisfies the following conditions: for some $\alpha\in(0, x^*(1-x^*)/3)$ ($\alpha$ could be a function of $T$),
\begin{itemize}[leftmargin=*]
    \item \textbf{Small perturbation condition:} 
    $\lambda(t) \in [x^*-\alpha, x^* + \alpha]$ and $\mu(t) \in [x^*-\alpha, x^* + \alpha]$.
    \item \textbf{Negative drift condition:}
    If $Q_{\mathrm{c}}(t)\ge \Theta(1/\alpha)$, $\mu(t) - \lambda(t) \ge \Theta(\alpha)$. If $Q_{\mathrm{s}}(t)\ge \Theta(1/\alpha)$, $\lambda(t) - \mu(t) \ge \Theta(\alpha)$. 
    \item \textbf{Bounded second moment:} $\expt [Q_{\mathrm{c}}(T+1)^2+Q_{\mathrm{s}}(T+1)^2] = o(T)$ as in Lemma~\ref{lemma:lower-bound}. 
\end{itemize}
Then, we can show that $\sqrt{\text{AvgQ$^2$Len}(T)} = \Theta(\text{AvgQLen}(T))$ for sufficiently large $T$ for any policy in this class.

We will first show that $\text{AvgQLen}(T) \ge \Theta(1/\alpha)$.
Let $Q(t)\coloneqq Q_{\mathrm{c}}(t) - Q_{\mathrm{s}}(t)$.
Consider the Lyapunov function $Q^2(t)$. Consider the drift $Q^2(t+1)-Q^2(t)$. Following the same argument as that in the proof of Lemma~\ref{lemma:lower-bound} in Appendix~\ref{app:proof-lemma-lower-bound} (Equation \eqref{equ:lower-bound-4-drift-begin}), we have
\begin{align*}
    \expt\left[\sum_{t=1}^T |Q(t)|  |\lambda(t) - \mu(t)| \right] \ge & T x^* (1 - x^*)
    - \frac{3}{2} \sum_{t=1}^{T} \expt\left[ |\mu(t) - x^*|
    +|\lambda(t) - x^*|\right]\nonumber\\
    & - \frac{\expt[Q^2(T+1)]}{2}
\end{align*}
From the small perturbation condition, we have $|\lambda(t) - \mu(t)|\le 2\alpha$, $|\lambda(t) - x^*|\le \alpha$, and $|\mu(t) - x^*|\le \alpha$. Hence,
\begin{align*}
    \expt\left[\sum_{t=1}^T |Q(t)| \right] \ge \frac{T x^* (1 - x^*)}{2\alpha}
    - \frac{3T}{2} 
    - \frac{\expt[Q^2(T+1)]}{2}.
\end{align*}
By the bounded second moment condition and Lemma 1 in \cite{yang2024learning}, we have $\expt[Q^2(T+1)]=\expt [Q_{\mathrm{c}}(T+1)^2+Q_{\mathrm{s}}(T+1)^2] = o(T)$. Hence, we have
\begin{align*}
    \expt\left[\sum_{t=1}^T |Q(t)| \right] \ge \frac{T x^* (1 - x^*)}{2\alpha}
    - \frac{3T}{2} 
    - o(T).
\end{align*}
Divided both sides by $T$, we have
\begin{align*}
    \text{AvgQLen}(T) \ge \frac{x^* (1 - x^*)}{2\alpha} - \frac{3}{2} - o(1).
\end{align*}
Since $\alpha \le x^*(1-x^*)/3$, we have $\frac{x^* (1 - x^*)}{2\alpha} > \frac{3}{2}$. Hence, for sufficiently large $T$, we have
\begin{align}\label{equ:proof-cond-lower-bound-avg-q}
    \text{AvgQLen}(T) \ge \Theta(1/\alpha).
\end{align}

Next, we will show that $\text{AvgQ$^2$Len}(T) \le \Theta(1/\alpha^2)$. Consider the Lyapunov function $Q^2(t)$. Consider the drift $Q^2(t+1)-Q^2(t)$. Following the same argument as that in the proof of Lemma~\ref{lemma:lower-bound} in Appendix~\ref{app:proof-lemma-lower-bound} (Equation \eqref{equ:lower-bound-4-drift}), we have
\begin{align}\label{equ:second-moment-drift}
    \expt[Q^2(T+1)] = & \sum_{t=1}^{T} \expt[(A_{\mathrm{c}}(t) - A_{\mathrm{s}}(t))^2] 
    + 2 \expt\left[\sum_{t=1}^T Q(t)  \left(\lambda(t) - \mu(t)\right) \right]\nonumber\\
    \le & T + 2 \expt\left[\sum_{t=1}^T Q(t)  \left(\lambda(t) - \mu(t)\right) \right].
\end{align}
Consider the following three cases:
\begin{itemize}[leftmargin=*]
    \item $Q_{\mathrm{c}}(t)\ge \Theta(1/\alpha)$:
    By Lemma 1 in \cite{yang2024learning}, $Q_{\mathrm{s}}(t)=0$. Hence, $Q(t) = Q_{\mathrm{c}}(t) \ge \Theta(1/\alpha)$. By the negative drift condition, we have
    $\mu(t) - \lambda(t)\ge \Theta(\alpha)$. 
    \item $Q_{\mathrm{s}}(t)\ge \Theta(1/\alpha)$: Similarly, we have $Q(t) = - Q_{\mathrm{s}}(t) \le - \Theta(1/\alpha)$. By the negative drift condition, we have
    $\lambda(t) - \mu(t) \ge \Theta(\alpha)$. 
    \item $Q_{\mathrm{c}}(t)< \Theta(1/\alpha)$ and $Q_{\mathrm{s}}(t)< \Theta(1/\alpha)$: In this case, we have $|Q(t)| = \max\{Q_{\mathrm{c}}(t), Q_{\mathrm{s}}(t)\} < \Theta(1/\alpha)$
\end{itemize}
Combining the three cases and \eqref{equ:second-moment-drift}, we have
\begin{align*}
    \expt[Q^2(T+1)]
    \le & T - 2 \alpha \expt\left[\sum_{t=1}^T \mathbb{1}_{Q_{\mathrm{c}}(t)\ge \Theta(1/\alpha)} |Q(t)| \right]
    - 2 \alpha \expt\left[\sum_{t=1}^T \mathbb{1}_{Q_{\mathrm{s}}(t)\ge \Theta(1/\alpha)} |Q(t)| \right]\nonumber\\
    & + 2 \expt\left[\sum_{t=1}^T \mathbb{1}_{|Q(t)| < \Theta(1/\alpha)} |Q(t)| \left|\lambda(t) - \mu(t)\right|\right]\nonumber\\
    \le & T - 2 \alpha \expt\left[\sum_{t=1}^T \mathbb{1}_{Q_{\mathrm{c}}(t)\ge \Theta(1/\alpha)} |Q(t)| \right]
    - 2 \alpha \expt\left[\sum_{t=1}^T \mathbb{1}_{Q_{\mathrm{s}}(t)\ge \Theta(1/\alpha)} |Q(t)| \right]
    + 4T,
\end{align*}
where the last inequality is by the small perturbation condition. Rearranging terms, we have 
\begin{align}\label{equ:second-moment-drift-2}
    \expt\left[\sum_{t=1}^T \mathbb{1}_{Q_{\mathrm{c}}(t)\ge \Theta(1/\alpha)} |Q(t)| \right]
    +\expt\left[\sum_{t=1}^T \mathbb{1}_{Q_{\mathrm{s}}(t)\ge \Theta(1/\alpha)} |Q(t)| \right] 
    \le \frac{5T}{2\alpha}.
\end{align}
Hence, we have
\begin{align}\label{equ:second-moment-drift-final}
    \expt\left[ \sum_{t=1}^{T} |Q(t)| \right] = & \expt\left[\sum_{t=1}^T \mathbb{1}_{Q_{\mathrm{c}}(t)\ge \Theta(1/\alpha)} |Q(t)| \right]
    +\expt\left[\sum_{t=1}^T \mathbb{1}_{Q_{\mathrm{s}}(t)\ge \Theta(1/\alpha)} |Q(t)| \right]\nonumber\\
    & + \expt\left[\sum_{t=1}^T \mathbb{1}_{|Q(t)| < \Theta(1/\alpha)} |Q(t)| \right]\nonumber\\
    \le & \frac{5T}{2\alpha} + \frac{T}{\alpha} = \frac{7T}{2\alpha},
\end{align}
where the last inequality is by \eqref{equ:second-moment-drift-2}.
Consider the Lyapunov function $Q^3_{\mathrm{c}}(t)+Q^3_{\mathrm{s}}(t)$. Consider the drift $\expt[Q^3_{\mathrm{c}}(t+1)+Q^3_{\mathrm{s}}(t+1) - Q^3_{\mathrm{c}}(t) - Q^3_{\mathrm{s}}(t)]$. Let $X(t)$ denote the number of matches in time slot $t$. By the queue dynamics, we have
\begin{align}\label{equ:third-moment-drift}
    & \expt[Q^3_{\mathrm{c}}(t+1)+Q^3_{\mathrm{s}}(t+1) - Q^3_{\mathrm{c}}(t) - Q^3_{\mathrm{s}}(t)]\nonumber\\
    = & \expt[(Q_{\mathrm{c}}(t)+A_{\mathrm{c}}(t)-X(t))^3 - Q^3_{\mathrm{c}}(t)
    + (Q_{\mathrm{s}}(t)+A_{\mathrm{s}}(t)-X(t))^3 - Q^3_{\mathrm{s}}(t)]\nonumber\\
    = & \expt[(A_{\mathrm{c}}(t)-X(t))^3 
    + 3 Q^2_{\mathrm{c}}(t) (A_{\mathrm{c}}(t)-X(t))
    + 3 Q_{\mathrm{c}}(t) (A_{\mathrm{c}}(t)-X(t))^2\nonumber\\
    & + (A_{\mathrm{s}}(t)-X(t))^3 
    + 3Q^2_{\mathrm{s}}(t) (A_{\mathrm{s}}(t)-X(t))
    + 3Q_{\mathrm{s}}(t) (A_{\mathrm{s}}(t)-X(t))^2]\nonumber\\
    \le & 2 + 3 \expt[|Q(t)|]  + 3 \expt[Q^2_{\mathrm{c}}(t) (A_{\mathrm{c}}(t)-X(t)) + Q^2_{\mathrm{s}}(t) (A_{\mathrm{s}}(t)-X(t))],
\end{align}
where the inequality is by the fact that the number of arrivals and the number of matches in each time slot are at most 1. Note that either $Q^2_{\mathrm{c}}(t)=0$ or $Q^2_{\mathrm{s}}(t)=0$ by Lemma 1 in \cite{yang2024learning}. Hence, for the term $\expt[Q^2_{\mathrm{c}}(t) (A_{\mathrm{c}}(t)-X(t)) + Q^2_{\mathrm{s}}(t) (A_{\mathrm{s}}(t)-X(t))]$ in \eqref{equ:third-moment-drift}, we have
\begin{align}\label{equ:third-moment-drift-2}
    & \expt[Q^2_{\mathrm{c}}(t) (A_{\mathrm{c}}(t)-X(t)) + Q^2_{\mathrm{s}}(t) (A_{\mathrm{s}}(t)-X(t))\nonumber\\
    = & \expt\left[ Q^2(t) \left[ \mathbb{1}_{Q_{\mathrm{c}}(t)>0} (A_{\mathrm{c}}(t)-X(t))
    + \mathbb{1}_{Q_{\mathrm{s}}(t)>0} (A_{\mathrm{s}}(t)-X(t))\right] \right]\nonumber\\
    = & \expt\left[ Q^2(t) \left[ \mathbb{1}_{Q_{\mathrm{c}}(t)>0} \expt[A_{\mathrm{c}}(t)-X(t)|Q(t)]
    + \mathbb{1}_{Q_{\mathrm{s}}(t)>0} \expt[A_{\mathrm{s}}(t)-X(t)|Q(t)]\right] \right]\nonumber\\
    = &  \expt\left[ Q^2(t) \left[ \mathbb{1}_{Q_{\mathrm{c}}(t)>0} (\lambda(t) - \mu(t))
    + \mathbb{1}_{Q_{\mathrm{s}}(t)>0} 
    (\mu(t)-\lambda(t))
    \right] \right],
\end{align}
where the second equality is by the law of iterated expectation, and the last equality is by the fact that we match all pairs whenever possible. 
Considering three cases, from  \eqref{equ:third-moment-drift-2}, we have
\begin{align}\label{equ:third-moment-drift-3}
    & \expt[Q^2_{\mathrm{c}}(t) (A_{\mathrm{c}}(t)-X(t)) + Q^2_{\mathrm{s}}(t) (A_{\mathrm{s}}(t)-X(t))]\nonumber\\
    \le &  
    \expt\left[ Q^2(t) \mathbb{1}_{Q_{\mathrm{c}}(t)\ge \Theta(1/\alpha)} (\lambda(t) - \mu(t))\right]
    + 
    \expt\left[ Q^2(t) \mathbb{1}_{Q_{\mathrm{s}}(t)\ge \Theta(1/\alpha)} 
    (\mu(t)-\lambda(t))\right]\nonumber\\
    & +
    \expt\left[ Q^2(t) \left[ \mathbb{1}_{|Q(t)|<\Theta(1/\alpha)} |\lambda(t) - \mu(t)|\right]\right]\nonumber\\
    \le & -\alpha  \expt\left[ Q^2(t) \mathbb{1}_{Q_{\mathrm{c}}(t)\ge \Theta(1/\alpha)}\right]
    -\alpha  \expt\left[ Q^2(t) \mathbb{1}_{Q_{\mathrm{s}}(t)\ge \Theta(1/\alpha)}\right]
    + \frac{2}{\alpha},
\end{align}
where the last inequality is by the negative drift condition and the small perturbation condtion.
From \eqref{equ:third-moment-drift} and \eqref{equ:third-moment-drift-3}, we have
\begin{align*}
    & \expt[Q^3_{\mathrm{c}}(t+1)+Q^3_{\mathrm{s}}(t+1) - Q^3_{\mathrm{c}}(t) - Q^3_{\mathrm{s}}(t)]\nonumber\\
    \le & 2 + 3 \expt[|Q(t)|]   -3\alpha  \expt\left[ Q^2(t) \mathbb{1}_{Q_{\mathrm{c}}(t)\ge \Theta(1/\alpha)}\right]
    -3\alpha  \expt\left[ Q^2(t) \mathbb{1}_{Q_{\mathrm{s}}(t)\ge \Theta(1/\alpha)}\right]
    + \frac{6}{\alpha}.
\end{align*}
Summing over $t$ from $t=1$ to $t=T$ and rearranging terms, we obtain
\begin{align}\label{equ:third-moment-drift-4}
    \sum_{t=1}^{T} \expt\left[ Q^2(t) \mathbb{1}_{Q_{\mathrm{c}}(t)\ge \Theta(1/\alpha)}\right]
     +  \sum_{t=1}^{T} \expt\left[ Q^2(t) \mathbb{1}_{Q_{\mathrm{s}}(t)\ge \Theta(1/\alpha)}\right]
    \le \frac{2T}{3\alpha} + \frac{1}{\alpha} \sum_{t=1}^{T} \expt[|Q(t)|]   
    + \frac{2T}{\alpha^2}.
\end{align}
From \eqref{equ:second-moment-drift-final} and \eqref{equ:third-moment-drift-4}, we have
\begin{align}\label{equ:third-moment-drift-5}
    \sum_{t=1}^{T} \expt\left[ Q^2(t) \mathbb{1}_{Q_{\mathrm{c}}(t)\ge \Theta(1/\alpha)}\right]
     +  \sum_{t=1}^{T} \expt\left[ Q^2(t) \mathbb{1}_{Q_{\mathrm{s}}(t)\ge \Theta(1/\alpha)}\right]
    \le \frac{2T}{3\alpha} + \frac{7T}{2\alpha^2}
    + \frac{2T}{\alpha^2} = \frac{2T}{3\alpha} + \frac{11T}{2\alpha^2}.
\end{align}
Hence, we have
\begin{align}\label{equ:third-moment-drift-final}
    & \sum_{t=1}^{T} \expt\left[ Q^2(t) \right]\nonumber\\
    = &
    \sum_{t=1}^{T} \expt\left[ Q^2(t) \mathbb{1}_{Q_{\mathrm{c}}(t)\ge \Theta(1/\alpha)}\right]
     +  \sum_{t=1}^{T} \expt\left[ Q^2(t) \mathbb{1}_{Q_{\mathrm{s}}(t)\ge \Theta(1/\alpha)}\right]
     + \sum_{t=1}^{T} \expt\left[ Q^2(t) \mathbb{1}_{|Q(t)|< \Theta(1/\alpha)}\right]\nonumber\\
     \le &  \frac{2T}{3\alpha} + \frac{11T}{2\alpha^2} + \frac{T}{\alpha^2} = \Theta(T/\alpha^2),
\end{align}
where the last inequality is by \eqref{equ:third-moment-drift-5}. 

Hence, combining \eqref{equ:proof-cond-lower-bound-avg-q} and \eqref{equ:third-moment-drift-final}, and noting that $\text{AvgQ$^2$Len}(T) \coloneqq \frac{1}{T} \sum_{t=1}^T \expt[Q_{\mathrm{c}}^2(t)+Q_{\mathrm{s}}^2(t)] = \frac{1}{T} \sum_{t=1}^T \expt[ Q^2(t)]$, we finally obtain
\[
 \sqrt{\text{AvgQ$^2$Len}(T)} \le \Theta(1/\alpha) \le  \Theta(\text{AvgQLen}(T)).
\]
By Jensen's inequality, we have
\[
 \text{AvgQLen}(T) = \frac{1}{T}\sum_{t=1}^{T} \expt [|Q(t)|]
 \le  \sqrt{\frac{1}{T}\sum_{t=1}^{T} (\expt [|Q(t)|])^2 }
 \le \sqrt{\text{AvgQ$^2$Len}(T)}.
\]
Therefore, we complete the proof of $\sqrt{\text{AvgQ$^2$Len}(T)} = \Theta(\text{AvgQLen}(T))$. 

\clearpage
\newpage

\section{Proof of Theorem~\ref{theo:3}}
\label{app:theo:3}

In this section, we present the complete proof of Theorem~\ref{theo:3}.

\subsection{Adding the Event of Concentration}
\label{sec:good-event}

We will use the ``good'' event ${\cal C}$ that was defined in \cite{yang2024learning} and is shown as follows for completeness.
\begin{align*}
    & {\cal C} \coloneqq \nonumber\\
    \Biggl\{
    & \text{for all outer iteration $k=1,\ldots,\lceil T/(2MN) \rceil$, all bisection iteration $m=1,\ldots, M$, all $i,j$, and}\nonumber\\ 
    & \text{all arrival rates $\lambda_i(t^{+}_{\mathrm{c}, i}(k,m,n)), \lambda_i(t^{-}_{\mathrm{c}, i}(k,m,n)), \mu_j(t^{+}_{\mathrm{s}, j}(k,m,n)), \mu_j(t^{-}_{\mathrm{s}, j}(k,m,n)) \in[0,1]$}, \nonumber\\
    & \biggl| \frac{1}{N} \sum_{n=1}^{N} A_{\mathrm{c},i} (t^{+}_{\mathrm{c}, i}(k,m,n)) - \lambda_i(t^{+}_{\mathrm{c}, i}(k,m,n)) \biggr| < \epsilon,\nonumber\\
    & \biggl| \frac{1}{N} \sum_{n=1}^{N} A_{\mathrm{c},i} (t^{-}_{\mathrm{c}, i}(k,m,n)) - \lambda_i(t^{-}_{\mathrm{c}, i}(k,m,n)) \biggr| < \epsilon, \nonumber\\
    & \biggl| \frac{1}{N} \sum_{n=1}^{N} A_{\mathrm{s},j} (t^{+}_{\mathrm{s}, j}(k,m,n)) - \mu_j(t^{+}_{\mathrm{s}, j}(k,m,n)) \biggr| < \epsilon,\nonumber\\
    & \biggl| \frac{1}{N} \sum_{n=1}^{N} A_{\mathrm{s},j} (t^{-}_{\mathrm{s}, j}(k,m,n)) - \mu_j(t^{-}_{\mathrm{s}, j}(k,m,n)) \biggr| < \epsilon
    \Biggr\},
\end{align*}
where $t^{+}_{\mathrm{c}, i}(k,m,n)$ and $t^{-}_{\mathrm{c}, i}(k,m,n)$ are the time slots when the prices $p^{+}_{\mathrm{c}, i} (k,m)$ and $p^{-}_{\mathrm{c}, i} (k,m)$ are applied for the $n^{\mathrm{th}}$ time in customer-side queue $i$, respectively. Similarly, $t^{+}_{\mathrm{s}, j}(k,m,n)$ and $t^{-}_{\mathrm{s}, j}(k,m,n)$ are the time slots when the prices $p^{+}_{\mathrm{s}, j} (k,m)$ and $p^{-}_{\mathrm{s}, j} (k,m)$ are applied for the $n^{\mathrm{th}}$ time in server-side queue $j$, respectively.

The ``good'' event ${\cal C}$ means that all estimated arrival rates are close to (i.e., concentrated around) their corresponding true arrival rates.
We call the iterations in Algorithm~\ref{alg:pricing} outer iterations, while call the iterations in Algorithm~\ref{alg:bisection} bisection iterations.
Let $K$ denote the number of outer iterations. Note that $K$ is at most $T/(2MN)$ because there are $2M$ bisection iterations in each outer iteration and there are at least $N$ time slots in each bisection iteration.
If the actual number of outer iterations $K< \lceil T/(2MN) \rceil$, additional hypothetical outer iterations can be introduced to extend the same process, ensuring that the definition of ${\cal C}$ remains valid. 

The paper \cite{yang2024learning} showed that the event ${\cal C}$ happens with high probability, as shown in the following lemma:
\begin{lemma}[\cite{yang2024learning}]
    \label{lemma:concentration}
    The probability of the complement of event ${\cal C}$ is bounded as
    \[
    \prob \left( {\cal C}^{\mathrm{c}} \right) \le \Theta\left(T\epsilon^{\frac{\beta}{2}+1} + \epsilon^{\frac{\beta}{2}-1} \log (1/\epsilon) \right),
    \]
    where the notation $\Theta(\cdot)$ omits dependence on variables other than $T$.
\end{lemma}
Adding the event $\mathcal{C}$ to the regret definition~\eqref{equ:regret-def}, the paper \cite{yang2024learning} showed that
\begin{align}
    & \expt [R(T)] 
    = \sum_{t=1}^{T}  \Biggl[ \biggl(\sum_i \lambda^*_i F_i(\lambda^*_i) - \sum_j \mu^*_j G_j(\mu^*_j) \biggr) \nonumber\\
    & -   \expt \left[  \sum_i \lambda_i(t) F_i(\lambda_i(t)) - \sum_j \mu_j(t) G_j(\mu_j(t))  \right]  \Biggr]\nonumber\\
    = & \expt \Biggl[ \mathbb{1}_{{\cal C}} \sum_{t=1}^{T} \biggl[ \biggl(\sum_i \lambda^*_i F_i(\lambda^*_i) - \sum_j \mu^*_j G_j(\mu^*_j) \biggr) 
    -    \biggl(  \sum_i \lambda_i(t) F_i(\lambda_i(t)) - \sum_j \mu_j(t) G_j(\mu_j(t))  \biggr) \biggr] \Biggr]\label{equ:theo-3-regret-term-1}\\
    & + \Theta\left(T^2\epsilon^{\frac{\beta}{2}+1} + T\epsilon^{\frac{\beta}{2}-1} \log (1/\epsilon) \right).\label{equ:theo-3-regret-term-2-bound}
\end{align}

\subsection{Bounding the Regret Caused by Rejecting Arrivals}
In this subsection, we will divide the term \eqref{equ:theo-3-regret-term-1} into two cases by comparing the queue lengths with the threshold $q^{\mathrm{th}}$, and then bound the regret induced by rejecting arrivals when the queue length exceeds the threshold $q^{\mathrm{th}}$.

Define the event ${\cal H}_t$:
\[
{\cal H}_t \coloneqq \left\{ Q_{\mathrm{c},i}(t) < q^{\mathrm{th}} \mbox{ for all } i, \mbox{ and } Q_{\mathrm{s},j}(t) < q^{\mathrm{th}} \mbox{ for all } j\right\}.
\]
Following the same argument as in \cite{yang2024learning}, we obtain
\begin{align}
    \eqref{equ:theo-3-regret-term-1} & =  \expt \Biggl[ \mathbb{1}_{{\cal C}} \sum_{t=1}^{T} \mathbb{1}_{{\cal H}_t^{\mathrm{c}}} \biggl[ \biggl(\sum_i \lambda^*_i F_i(\lambda^*_i) - \sum_j \mu^*_j G_j(\mu^*_j) \biggr)\nonumber\\ 
    & \qquad-    \biggl(  \sum_i \lambda_i(t) F_i(\lambda_i(t)) - \sum_j \mu_j(t) G_j(\mu_j(t))  \biggr) \biggr]  \Biggr]\nonumber\\
    & + \expt \Biggl[ \mathbb{1}_{{\cal C}} \sum_{t=1}^{T} \mathbb{1}_{{\cal H}_t} \biggl[ \biggl(\sum_i \lambda^*_i F_i(\lambda^*_i) - \sum_j \mu^*_j G_j(\mu^*_j) \biggr)\nonumber\\ 
    & \qquad -    \biggl(  \sum_i \lambda_i(t) F_i(\lambda_i(t)) - \sum_j \mu_j(t) G_j(\mu_j(t))  \biggr) \biggr] \Biggr]\nonumber\\
    & \le \biggl[\sum_i \lambda^*_i F_i(\lambda^*_i) - \sum_j \mu^*_j G_j(\mu^*_j) + \sum_j G_j(1)\biggr]\nonumber\\
    & \qquad \left(\sum_{t=1}^{T} \sum_i \expt \biggl[  \mathbb{1}_{\cal C} \mathbb{1} \left\{ Q_{\mathrm{c},i}(t) \ge q^{\mathrm{th}}\right\} \biggr] 
    + \sum_{t=1}^{T} \sum_j \expt \biggl[  \mathbb{1}_{\cal C} \mathbb{1} \left\{ Q_{\mathrm{s},j}(t) \ge q^{\mathrm{th}}\right\}   \biggr]\right)\label{equ:theo-3-regret-term-3}\\
    & +  \expt \Biggl[ \mathbb{1}_{{\cal C}} \sum_{t=1}^{T} \mathbb{1}_{{\cal H}_t} \biggl[ \biggl(\sum_i \lambda^*_i F_i(\lambda^*_i) - \sum_j \mu^*_j G_j(\mu^*_j) \biggr) \nonumber\\
    & \qquad -    \biggl(  \sum_i \lambda_i(t) F_i(\lambda_i(t)) - \sum_j \mu_j(t) G_j(\mu_j(t))  \biggr) \biggr]  \Biggr],\label{equ:theo-3-regret-term-4}
\end{align}
For the term \eqref{equ:theo-3-regret-term-3}, similar to Lemma 6 in \cite{yang2024learning}, we have the following lemma.
\begin{lemma}\label{lemma:theo-3-bound-time-large-queue}
    Let Assumption~\ref{assum:1}, Assumption~\ref{assum:3}, and Assumption~\ref{assum:5} hold. Suppose $\epsilon < \delta$ and $T$ is sufficiently large. Under the proposed algorithm, we have
    \begin{align*}
    &  \sum_{t=1}^{T} \sum_i \expt \Bigl[ 
    \mathbb{1}_{{\cal C}}
    \mathbb{1} \Bigl\{ Q_{\mathrm{c},i}(t) \ge q^{\mathrm{th}} \Bigr\}
    \Bigr] + \sum_{t=1}^{T} \sum_j \expt \Bigl[ 
    \mathbb{1}_{{\cal C}}
    \mathbb{1} \Bigl\{ Q_{\mathrm{s},j}(t) \ge q^{\mathrm{th}} \Bigr\}
    \Bigr] \nonumber\\
    \le & \Theta \left(
    \frac{T}{q^{\mathrm{th}}}
    + T^2\epsilon^{\frac{\beta}{2}+1} + T\epsilon^{\frac{\beta}{2}-1} \log (1/\epsilon) 
    + T \left(\frac{\eta \epsilon}{\delta} + \eta + \delta + \epsilon\right)  \right),
    \end{align*}
    where in the notation $\Theta(\cdot)$ we ignore the variables that do not depend on $T$.
\end{lemma}
\noindent Lemma~\ref{lemma:theo-3-bound-time-large-queue} provides upper bounds for the expected number of time slots when the queue length is greater than or equal to $q^{\mathrm{th}}$ and the ``good'' event ${\cal C}$ holds. Proof of Lemma~\ref{lemma:theo-3-bound-time-large-queue} can be found in Appendix~\ref{app:proof-lemma-theo-3-bound-time-large-queue}. Compared to Lemma 6 in \cite{yang2024learning}, the difference of the proof is that we need to consider the random reduction of arrival rates in the proposed probabilistic two-price policy.

From \eqref{equ:theo-3-regret-term-3} and Lemma~\ref{lemma:theo-3-bound-time-large-queue}, the regret caused by rejecting arrivals can be bounded by
\begin{align}\label{equ:theo-3-regret-term-3-final}
    \eqref{equ:theo-3-regret-term-3}
    \le &  \Theta \left(
    \frac{T}{q^{\mathrm{th}}}
    + T^2\epsilon^{\frac{\beta}{2}+1} + T\epsilon^{\frac{\beta}{2}-1} \log (1/\epsilon) 
    + T \left(\frac{\eta \epsilon}{\delta} + \eta + \delta + \epsilon\right)  \right).
\end{align}

\subsection{Bounding the Regret Caused by Reducing Arrival Rates}
\label{sec:bound-regret-reducing-arr}
In this subsection, we will bound the term \eqref{equ:theo-3-regret-term-4} in the regret bound, where we randomly reduce the arrival rate if the queue is not empty.

Define a function
\[
h(\boldsymbol{\lambda}, \boldsymbol{\mu}) \coloneqq \sum_i \lambda_i F_i(\lambda_i) - \sum_j \mu_j G_j (\mu_j).
\]
\begin{lemma}\label{lemma:h-properties}
    Let Assumption~\ref{assum:1} and Assumption~\ref{assum:7} hold. The function $h$ has the following properties:
    \begin{itemize}
        \item[(1)] $h$ is Lipschitz in $l_1$ norm, i.e., there exists a constant $L_h > 0$ such that for any arrival rate vectors $\boldsymbol{\lambda}^{(1)}$, $\boldsymbol{\mu}^{(1)}$, $\boldsymbol{\lambda}^{(2)}$, $\boldsymbol{\mu}^{(2)}$,
        \begin{align}\label{equ:lipschitz-h}
            \left|h(\boldsymbol{\lambda}^{(1)}, \boldsymbol{\mu}^{(1)}) - h(\boldsymbol{\lambda}^{(2)}, \boldsymbol{\mu}^{(2)})\right|
            \le L_h \left\| 
            \begin{bmatrix}
            \boldsymbol{\lambda}^{(1)} - \boldsymbol{\lambda}^{(2)} \\
            \boldsymbol{\mu}^{(1)} - \boldsymbol{\mu}^{(2)}
            \end{bmatrix}
            \right\|_1.
        \end{align}

        \item[(2)] $h$ is smooth, i,e, there exists a constant $\beta_h > 0$ such that for any arrival rate vectors $\boldsymbol{\lambda}^{(1)}$, $\boldsymbol{\mu}^{(1)}$, $\boldsymbol{\lambda}^{(2)}$, $\boldsymbol{\mu}^{(2)}$,
        \begin{align}\label{equ:smooth-h}
            \left\| \nabla h (\boldsymbol{\lambda}^{(1)}, \boldsymbol{\mu}^{(1)}) - \nabla h (\boldsymbol{\lambda}^{(2)}, \boldsymbol{\mu}^{(2)}) \right\|_2
            \le \beta_h \left\|
            \begin{bmatrix}
            \boldsymbol{\lambda}^{(1)} - \boldsymbol{\lambda}^{(2)}\\
            \boldsymbol{\mu}^{(1)} - \boldsymbol{\mu}^{(2)}
            \end{bmatrix}
            \right\|_2
        \end{align}
    \end{itemize}
\end{lemma}
\noindent  Proof of Lemma~\ref{lemma:h-properties} can be found in Appendix~\ref{app:proof-lemma-h-properties}.
By the definition of $h$, \eqref{equ:theo-3-regret-term-4} can be rewritten as
\begin{align*}
    \eqref{equ:theo-3-regret-term-4} = \expt \Biggl[ \mathbb{1}_{\cal C} \sum_{t=1}^{T} \mathbb{1}_{{\cal H}_t} \bigl[ h(\boldsymbol{\lambda}^*, \boldsymbol{\mu}^*)
    -    h(\boldsymbol{\lambda}(t), \boldsymbol{\mu}(t)) \bigr]  \Biggr].
\end{align*}
Define the following independent and identically distributed Bernoulli random variables with parameter $1/2$: $B_{\mathrm{c},i}(t)$ and $B_{\mathrm{s},j}(t)$, $i=1,\ldots,I$, $j=1,\ldots,J$, $t=1,\ldots,T$.
Then define
\[
\alpha_{\mathrm{c},i}(t) \coloneqq \begin{cases}
    \alpha B_{\mathrm{c},i}(t), \mbox{ if } Q_{\mathrm{c}, i}(t) > 0;\\
    0, \mbox{ else}
\end{cases},\qquad
\alpha_{\mathrm{s},j}(t) \coloneqq \begin{cases}
    \alpha B_{\mathrm{s},j}(t), \mbox{ if } Q_{\mathrm{s}, j}(t) > 0;\\
    0, \mbox{ else.}
\end{cases}
\]
Define the following arrival rates:
\begin{align*}
    \boldsymbol{\lambda}_{\alpha}(t) \coloneqq ( \lambda_{\alpha, i}(t) )_{i\in {\cal I}}
    \coloneqq (F^{-1}_i ( \min\{ p^{+/-}_{\mathrm{c},i}(k(t),m(t)) + \alpha_{\mathrm{c},i}(t), p_{\mathrm{c},i,\max}\} ) )_{i\in {\cal I}},\nonumber\\
    \boldsymbol{\mu}_{\alpha}(t) \coloneqq ( \mu_{\alpha, j}(t) )_{j\in {\cal J}}
\coloneqq (G^{-1}_j ( \max\{ p^{+/-}_{\mathrm{s},j}(k(t),m(t)) - \alpha_{\mathrm{s},j}(t), p_{\mathrm{s},j,\min} \}) )_{j\in {\cal J}}.
\end{align*}
Then by the \textit{probabilistic two-price policy} where we randomly adjust the price to reduce the arrival rate when the queue length is between $0$ and $q^{\mathrm{th}}$, we have
\begin{align}\label{equ:theo-3-regret-term-4-1}
    \eqref{equ:theo-3-regret-term-4} = \expt \Biggl[ \mathbb{1}_{\cal C} \sum_{t=1}^{T} \mathbb{1}_{{\cal H}_t} \bigl[ h(\boldsymbol{\lambda}^*, \boldsymbol{\mu}^*)
    -    h(\boldsymbol{\lambda}_{\alpha}(t), \boldsymbol{\mu}_{\alpha}(t))  \bigr]  \Biggr].
\end{align}
By Assumption~\ref{assum:7}(1), we know that $F_i^{-1}$ is smooth. Consider any smooth and monotone extension of the domain of $F_i^{-1}$.
Then, $F_i^{-1}$ has a quadratic upper bound~\cite{bertsekas1999nonlinear} as follows:
\begin{align}\label{equ:taylor-1}
    & \lambda_{\alpha, i}(t)\nonumber\\
    = & \max\{ F^{-1}_i (p^{+/-}_{\mathrm{c},i}(k(t),m(t)) + \alpha_{\mathrm{c},i}(t)), 0\}\nonumber\\
    \le & \max\left\{ F_i^{-1}(p^{+/-}_{\mathrm{c},i}(k(t),m(t))) + \alpha_{\mathrm{c},i}(t) \diff*{F_i^{-1}(p)}{p}{p=p^{+/-}_{\mathrm{c},i}(k(t),m(t))}  + \frac{\beta_{F_i^{-1}}}{2} \alpha_{\mathrm{c},i}^2(t),
    0 \right\}\nonumber\\
    = & \max\left\{ \lambda'_i(t) + \alpha_{\mathrm{c},i}(t) \diff*{F_i^{-1}(p)}{p}{p=p^{+/-}_{\mathrm{c},i}(k(t),m(t))}  + \frac{\beta_{F_i^{-1}}}{2} \alpha_{\mathrm{c},i}^2(t),
    0 \right\},
\end{align}
where we recall $\lambda'_i(t)\coloneqq F_i^{-1}(p^{+/-}_{\mathrm{c},i}(k(t),m(t)))$.
Similarly, we have
\begin{align}\label{equ:taylor-2}
    & \mu_{\alpha, j}(t) \nonumber\\
    = & \max\{ G^{-1}_j (p^{+/-}_{\mathrm{s},j}(k(t),m(t)) - \alpha_{\mathrm{s},j}(t)) , 0\}\nonumber\\
    \le & \max\left\{ G_j^{-1}(p^{+/-}_{\mathrm{s},j}(k(t),m(t))) - \alpha_{\mathrm{s},j}(t) \diff*{G_j^{-1}(p)}{p}{p=p^{+/-}_{\mathrm{s},j}(k(t),m(t))}  + \frac{\beta_{G_j^{-1}}}{2} \alpha^2_{\mathrm{s},j}(t),
    0 \right\}\nonumber\\
    = & \max\left\{\mu'_j(t) - \alpha_{\mathrm{s},j}(t) \diff*{G_j^{-1}(p)}{p}{p=p^{+/-}_{\mathrm{s},j}(k(t),m(t))}  + \frac{\beta_{G_j^{-1}}}{2} \alpha^2_{\mathrm{s},j}(t),
    0 \right\},
\end{align}
where we recall $\mu'_j(t) \coloneqq G_j^{-1}(p^{+/-}_{\mathrm{s},j}(k(t),m(t)))$.
Define
\begin{align}\label{equ:def-alpha-tilde-c}
    \tilde{\boldsymbol{\alpha}}_{\mathrm{c}}(t) \coloneqq
    \left(\tilde{\alpha}_{\mathrm{c},i}(t)\right)_{i\in {\cal I}} \coloneqq
    \left(-\alpha_{\mathrm{c},i}(t) \diff*{F_i^{-1}(p)}{p}{p=p^{+/-}_{\mathrm{c},i}(k(t),m(t))} \right)_{i\in {\cal I}}
\end{align}
and
\begin{align}\label{equ:def-alpha-tilde-s}
    \tilde{\boldsymbol{\alpha}}_{\mathrm{s}}(t) \coloneqq
    \left(\tilde{\alpha}_{\mathrm{s},j}(t)\right)_{j\in {\cal J}}
    \coloneqq \left(\alpha_{\mathrm{s},j}(t) \diff*{G_j^{-1}(p)}{p}{p=p^{+/-}_{\mathrm{s},j}(k(t),m(t))} \right)_{j\in {\cal J}}.
\end{align}
Then \eqref{equ:taylor-1} can be rewritten as
\begin{align}\label{equ:taylor-1-final}
    \lambda_{\alpha,i}(t)
    \le & \max\{\lambda'_i(t) - \tilde{\alpha}_{\mathrm{c},i}(t)  + \frac{\beta_{F_i^{-1}}}{2} \alpha_{\mathrm{c},i}^2(t), 0\}\nonumber\\
    \le & \max\{\lambda'_i(t) - \tilde{\alpha}_{\mathrm{c},i}(t)  + \Theta(\alpha^2), 0\},
\end{align}
where the last inequality holds since $\alpha_{\mathrm{c},i}(t) \le \alpha$.
Similarly, \eqref{equ:taylor-2} can be rewritten as
\begin{align}\label{equ:taylor-2-final}
    \mu_{\alpha, j}(t)
    \le & \max\{\mu'_j(t) - \tilde{\alpha}_{\mathrm{s},j}(t)  + \frac{\beta_{G_j^{-1}}}{2} \alpha^2_{\mathrm{s},j}(t), 0 \} \nonumber\\
    \le & \max\{\mu'_j(t) - \tilde{\alpha}_{\mathrm{s},j}(t)  + \Theta(\alpha^2), 0 \},
\end{align}
where the last inequality holds since $\alpha_{\mathrm{s},j}(t) \le \alpha$.
Since $F_i^{-1}$ and $G_j^{-1}$ are smooth, $-F_i^{-1}$ and $-G_j^{-1}$ are also smooth. Then by the quadratic upper bound of a smooth function~\cite{bertsekas1999nonlinear}, similar to \eqref{equ:taylor-1-final} and \eqref{equ:taylor-2-final}, we can also show that 
\begin{align}\label{equ:taylor-3-final}
\lambda_{\alpha,i}(t)
\ge \max\{\lambda'_i(t) - \tilde{\alpha}_{\mathrm{c},i}(t)  - \Theta(\alpha^2), 0\}
\end{align}
and
\begin{align}\label{equ:taylor-4-final}
\mu_{\alpha, j}(t)
\ge \max\{\mu'_j(t) - \tilde{\alpha}_{\mathrm{s},j}(t)  - \Theta(\alpha^2), 0\}.
\end{align}
From \eqref{equ:taylor-1-final}-\eqref{equ:taylor-4-final}, we have
\begin{align}\label{equ:taylor-final}
    |\max\{\lambda'_i(t) - \tilde{\alpha}_{\mathrm{c},i}(t), 0\} 
    - \lambda_{\alpha,i}(t)| 
    \le &  \Theta(\alpha^2) \mbox{ for all } i\nonumber\\
    |\max\{\mu'_j(t) - \tilde{\alpha}_{\mathrm{s},j}(t), 0\}
    - \mu_{\alpha, j}(t)| \le &  \Theta(\alpha^2) \mbox{ for all } j.
\end{align}
Then by \eqref{equ:taylor-final} and the Lipschitz property \eqref{equ:lipschitz-h} of $h$ in Lemma~\ref{lemma:h-properties}, we have
\begin{align}\label{equ:h-diff-alpha2}
    h(\boldsymbol{\lambda}_{\alpha}(t), \boldsymbol{\mu}_{\alpha}(t))
    \ge h ( \max\{\boldsymbol{\lambda}'(t) - \tilde{\boldsymbol{\alpha}}_{\mathrm{c}}(t), 0\}, 
    \max \{ \boldsymbol{\mu}'(t) - \tilde{\boldsymbol{\alpha}}_{\mathrm{s}}(t), 0\} )  - \Theta (\alpha^2),
\end{align}
where $\max\{\cdot, \cdot\}$ takes entrywise maximum, $\boldsymbol{\lambda}'(t) \coloneqq (\lambda'_i(t))_{i \in {\cal I}}$, and $\boldsymbol{\mu}'(t) \coloneqq (\mu'_j(t))_{j \in {\cal J}}$.
Hence, from \eqref{equ:theo-3-regret-term-4-1} and \eqref{equ:h-diff-alpha2}, we have
\begin{align*}
    \eqref{equ:theo-3-regret-term-4} 
    \le & \expt \Biggl[ \mathbb{1}_{\cal C} \sum_{t=1}^{T} \mathbb{1}_{{\cal H}_t}   \bigl[ h(\boldsymbol{\lambda}^*, \boldsymbol{\mu}^*)
    -     h ( \max\{ \boldsymbol{\lambda}'(t) - \tilde{\boldsymbol{\alpha}}_{\mathrm{c}}(t), 0\}, 
    \max\{ \boldsymbol{\mu}'(t) - \tilde{\boldsymbol{\alpha}}_{\mathrm{s}}(t), 0\} )   \bigr]  \Biggr]\nonumber\\
    & + \Theta ( T \alpha^2).
\end{align*}
Let $\boldsymbol{\lambda}^{\mathrm{tg}}(t) \coloneqq (\sum_{j\in {\cal E}_{\mathrm{c},i}} x_{i,j}(k(t)))_{i\in {\cal I}}$ and $\boldsymbol{\mu}^{\mathrm{tg}}(t) \coloneqq (\sum_{i\in {\cal E}_{\mathrm{s},j}} x_{i,j}(k(t)))_{j\in {\cal J}}$, which are vectors containing target arrival rates. Then by adding and subtracting $h(\boldsymbol{\lambda}^{\mathrm{tg}}(t), \boldsymbol{\mu}^{\mathrm{tg}}(t) )$, we have
\begin{align*}
    \eqref{equ:theo-3-regret-term-4} \le & \expt \Biggl[ \mathbb{1}_{\cal C} \sum_{t=1}^{T} \mathbb{1}_{{\cal H}_t}   \biggl[ h(\boldsymbol{\lambda}^*, \boldsymbol{\mu}^*)
     -  h(\boldsymbol{\lambda}^{\mathrm{tg}}(t), \boldsymbol{\mu}^{\mathrm{tg}}(t)) \nonumber\\
     & \qquad \qquad  \qquad + h(\boldsymbol{\lambda}^{\mathrm{tg}}(t), \boldsymbol{\mu}^{\mathrm{tg}}(t)) 
     -   h ( \max\{ \boldsymbol{\lambda}'(t) - \tilde{\boldsymbol{\alpha}}_{\mathrm{c}}(t), 0\}, 
    \max\{ \boldsymbol{\mu}'(t) - \tilde{\boldsymbol{\alpha}}_{\mathrm{s}}(t), 0\} )   \biggr]  \Biggr]\nonumber\\
    & + \Theta ( T \alpha^2).
\end{align*}
Next, by the Lipschitz property \eqref{equ:lipschitz-h} of $h$ in Lemma~\ref{lemma:h-properties} and the relation between $\boldsymbol{\lambda}'(t), \boldsymbol{\mu}'(t)$ and $\boldsymbol{\lambda}^{\mathrm{tg}}(t), \boldsymbol{\mu}^{\mathrm{tg}}(t)$ in Lemma~\ref{lemma:improved-bisection-error}, we obtain
\begin{align}\label{equ:theo-3-regret-term-4-2}
    \eqref{equ:theo-3-regret-term-4} 
    \le & \expt \Biggl[ \mathbb{1}_{\cal C} \sum_{t=1}^{T} \mathbb{1}_{{\cal H}_t}   \biggl[ h(\boldsymbol{\lambda}^*, \boldsymbol{\mu}^*)
     -  h(\boldsymbol{\lambda}^{\mathrm{tg}}(t), \boldsymbol{\mu}^{\mathrm{tg}}(t)) \nonumber\\
     & \qquad \qquad \quad + h(\boldsymbol{\lambda}^{\mathrm{tg}}(t), \boldsymbol{\mu}^{\mathrm{tg}}(t)) 
     -   h ( \max\{\boldsymbol{\lambda}^{\mathrm{tg}}(t) - \tilde{\boldsymbol{\alpha}}_{\mathrm{c}}(t), 0\}, 
    \max\{\boldsymbol{\mu}^{\mathrm{tg}}(t) - \tilde{\boldsymbol{\alpha}}_{\mathrm{s}}(t), 0\} )   \biggr]  \Biggr]\nonumber\\
    & + \Theta \left( T \left(\frac{\eta \epsilon}{\delta} + \eta + \delta + \epsilon\right) \right) + \Theta ( T \alpha^2).
\end{align}
By $\boldsymbol{x}(k(t)) \in {\cal D}'$, the definition of ${\cal D}'$, and the definitions of $\boldsymbol{\lambda}^{\mathrm{tg}}(t)$ and $\boldsymbol{\mu}^{\mathrm{tg}}(t)$,
we have $\lambda^{\mathrm{tg}}_i(t)\ge a_{\min}$ and $\mu^{\mathrm{tg}}_j(t)\ge a_{\min}$ for all $i,j$. Also note that the derivatives of $F_i^{-1}$ and $G_j^{-1}$ are bounded since they are smooth by Assumption~\ref{assum:7}(1). Hence, recalling the definitions of $\tilde{\boldsymbol{\alpha}}_{\mathrm{c}}(t)$ and $\tilde{\boldsymbol{\alpha}}_{\mathrm{s}}(t)$ in \eqref{equ:def-alpha-tilde-c} and \eqref{equ:def-alpha-tilde-s}, for sufficiently large $T$ such that $\alpha$ is sufficiently small, $\boldsymbol{\lambda}^{\mathrm{tg}}(t) - \tilde{\boldsymbol{\alpha}}_{\mathrm{c}}(t)$ and $\boldsymbol{\mu}^{\mathrm{tg}}(t) - \tilde{\boldsymbol{\alpha}}_{\mathrm{s}}(t)$ are both entrywise nonnegative. Therefore, from \eqref{equ:theo-3-regret-term-4-2}, we have
\begin{align}\label{equ:theo-3-regret-term-4-2-final}
    \eqref{equ:theo-3-regret-term-4}
    \le & \expt \Biggl[ \mathbb{1}_{\cal C} \sum_{t=1}^{T} \mathbb{1}_{{\cal H}_t} \biggl[ h(\boldsymbol{\lambda}^*, \boldsymbol{\mu}^*)
     -  h(\boldsymbol{\lambda}^{\mathrm{tg}}(t), \boldsymbol{\mu}^{\mathrm{tg}}(t)) \nonumber\\
     & \qquad \qquad \qquad + h(\boldsymbol{\lambda}^{\mathrm{tg}}(t), \boldsymbol{\mu}^{\mathrm{tg}}(t)) 
     -   h ( \boldsymbol{\lambda}^{\mathrm{tg}}(t) - \tilde{\boldsymbol{\alpha}}_{\mathrm{c}}(t), 
    \boldsymbol{\mu}^{\mathrm{tg}}(t) - \tilde{\boldsymbol{\alpha}}_{\mathrm{s}}(t) )   \biggr]  \Biggr]\nonumber\\
    & + \Theta \left( T \left(\frac{\eta \epsilon}{\delta} + \eta + \delta + \epsilon\right) \right) + \Theta ( T \alpha^2),
\end{align}
Next, we will bound the term $h(\boldsymbol{\lambda}^{\mathrm{tg}}(t), \boldsymbol{\mu}^{\mathrm{tg}}(t)) 
- h(\boldsymbol{\lambda}^{\mathrm{tg}}(t)-\tilde{\boldsymbol{\alpha}}_{\mathrm{c}}(t),
\boldsymbol{\mu}^{\mathrm{tg}}(t) - \tilde{\boldsymbol{\alpha}}_{\mathrm{s}}(t) )$ in \eqref{equ:theo-3-regret-term-4-2-final}. By Lemma~\ref{lemma:h-properties}, we know that $h$ is $\beta_h$-smooth. Then $-h$ is also $\beta_h$-smooth. Thus, $-h$ has a quadratic upper bound~\cite{bertsekas1999nonlinear} as follows:
\begin{align}\label{equ:h-quadr-upper-bound}
& - h(\boldsymbol{\lambda}^{\mathrm{tg}}(t)-\tilde{\boldsymbol{\alpha}}_{\mathrm{c}}(t),
\boldsymbol{\mu}^{\mathrm{tg}}(t) - \tilde{\boldsymbol{\alpha}}_{\mathrm{s}}(t) )\nonumber\\
\le & -h(\boldsymbol{\lambda}^{\mathrm{tg}}(t),
\boldsymbol{\mu}^{\mathrm{tg}}(t) )
+ [\tilde{\boldsymbol{\alpha}}_{\mathrm{c}}(t)^\top, \tilde{\boldsymbol{\alpha}}_{\mathrm{s}}(t)^\top ]
\nabla h(\boldsymbol{\lambda}^{\mathrm{tg}}(t),
\boldsymbol{\mu}^{\mathrm{tg}}(t) )
+ \frac{\beta_h}{2} \left\|
\begin{bmatrix}
\tilde{\boldsymbol{\alpha}}_{\mathrm{c}}(t)\\
\tilde{\boldsymbol{\alpha}}_{\mathrm{s}}(t)
\end{bmatrix}
\right\|^2_2.
\end{align}
Note that each entry in $\tilde{\boldsymbol{\alpha}}_{\mathrm{c}}(t)$ and $\tilde{\boldsymbol{\alpha}}_{\mathrm{s}}(t)$ is bounded and is less than or equal to $\Theta(\alpha)$ by definition since the derivatives of $F_i^{-1}$ and $G_j^{-1}$ are bounded.
Hence, the upper bound \eqref{equ:h-quadr-upper-bound} can be written as
\begin{align*}
    & - h(\boldsymbol{\lambda}^{\mathrm{tg}}(t)-\tilde{\boldsymbol{\alpha}}_{\mathrm{c}}(t),
    \boldsymbol{\mu}^{\mathrm{tg}}(t) - \tilde{\boldsymbol{\alpha}}_{\mathrm{s}}(t) )\nonumber\\
    \le & -h(\boldsymbol{\lambda}^{\mathrm{tg}}(t),
    \boldsymbol{\mu}^{\mathrm{tg}}(t) )
    + [\tilde{\boldsymbol{\alpha}}_{\mathrm{c}}(t)^\top, \tilde{\boldsymbol{\alpha}}_{\mathrm{s}}(t)^\top ]
    \nabla h(\boldsymbol{\lambda}^{\mathrm{tg}}(t),
    \boldsymbol{\mu}^{\mathrm{tg}}(t) )
    + \Theta (\alpha^2).
\end{align*}
Hence, we have
\begin{align}\label{equ:regret-reduce-arrival-per-time}
& h(\boldsymbol{\lambda}^{\mathrm{tg}}(t), \boldsymbol{\mu}^{\mathrm{tg}}(t)) 
- h(\boldsymbol{\lambda}^{\mathrm{tg}}(t)-\tilde{\boldsymbol{\alpha}}_{\mathrm{c}}(t),
\boldsymbol{\mu}^{\mathrm{tg}}(t) - \tilde{\boldsymbol{\alpha}}_{\mathrm{s}}(t) )\nonumber\\
\le & [\tilde{\boldsymbol{\alpha}}_{\mathrm{c}}(t)^\top, \tilde{\boldsymbol{\alpha}}_{\mathrm{s}}(t)^\top ]
\nabla h(\boldsymbol{\lambda}^{\mathrm{tg}}(t),
\boldsymbol{\mu}^{\mathrm{tg}}(t) )
+ \Theta (\alpha^2).
\end{align}
From \eqref{equ:theo-3-regret-term-4-2-final} and \eqref{equ:regret-reduce-arrival-per-time}, we have
\begin{align}
    \eqref{equ:theo-3-regret-term-4} \le & \expt \Biggl[ \mathbb{1}_{\cal C} \sum_{t=1}^{T} \mathbb{1}_{{\cal H}_t}   \bigl[ h(\boldsymbol{\lambda}^*, \boldsymbol{\mu}^*)
     -  h(\boldsymbol{\lambda}^{\mathrm{tg}}(t), \boldsymbol{\mu}^{\mathrm{tg}}(t)) \bigr]  \Biggr]\label{equ:theo-3-regret-term-4-3}\\
     & + \expt \Biggl[ \mathbb{1}_{\cal C} \sum_{t=1}^{T} \mathbb{1}_{{\cal H}_t}   
     [\tilde{\boldsymbol{\alpha}}_{\mathrm{c}}(t)^\top, \tilde{\boldsymbol{\alpha}}_{\mathrm{s}}(t)^\top ]
    \nabla h(\boldsymbol{\lambda}^{\mathrm{tg}}(t),
    \boldsymbol{\mu}^{\mathrm{tg}}(t) )
    \Biggr]\label{equ:theo-3-regret-term-4-4}\\
    & + \Theta \left( T \left(\frac{\eta \epsilon}{\delta} + \eta + \delta + \epsilon\right) \right) + \Theta ( T \alpha^2).\label{equ:theo-3-regret-term-4-5}
\end{align}
We will bound the term \eqref{equ:theo-3-regret-term-4-3} later in the next subsection. Now we focus on bounding the term \eqref{equ:theo-3-regret-term-4-4}, which, in fact, is the \textit{regret caused by reducing arrival rates}. We add and subtract $\nabla h (\boldsymbol{\lambda}^*, \boldsymbol{\mu}^*)$ as follows:
\begin{align}
    \eqref{equ:theo-3-regret-term-4-4}
    = & \expt \Biggl[ \mathbb{1}_{\cal C} \sum_{t=1}^{T} \mathbb{1}_{{\cal H}_t}
     [\tilde{\boldsymbol{\alpha}}_{\mathrm{c}}(t)^\top, \tilde{\boldsymbol{\alpha}}_{\mathrm{s}}(t)^\top ]
    [\nabla h(\boldsymbol{\lambda}^{\mathrm{tg}}(t),
    \boldsymbol{\mu}^{\mathrm{tg}}(t) )
    - \nabla h (\boldsymbol{\lambda}^*, \boldsymbol{\mu}^*)]
    \Biggr]\label{equ:theo-3-regret-term-4-4-1}\\
    & + \expt \Biggl[ \mathbb{1}_{\cal C} \sum_{t=1}^{T} \mathbb{1}_{{\cal H}_t}
     [\tilde{\boldsymbol{\alpha}}_{\mathrm{c}}(t)^\top, \tilde{\boldsymbol{\alpha}}_{\mathrm{s}}(t)^\top ]
     \nabla h (\boldsymbol{\lambda}^*, \boldsymbol{\mu}^*)
    \Biggr]\label{equ:theo-3-regret-term-4-4-2},
\end{align}

We will first bound the term \eqref{equ:theo-3-regret-term-4-4-2}.
The idea of bounding \eqref{equ:theo-3-regret-term-4-4-2} is borrowed from the proof of Lemma 2 in~\cite{varma2023dynamic}.
Recall the optimization problem \eqref{equ:fluid-opti-obj}-\eqref{equ:fluid-opti-constr4}:
\begin{align*}
    \max_{\boldsymbol{\lambda}, \boldsymbol{\mu}, \boldsymbol{x}} & h(\boldsymbol{\lambda}, \boldsymbol{\mu}) \\
    \mathrm{s.t.} \qquad \lambda_i = &  \sum_{j: (i,j)\in {\cal E}} x_{i,j}, \quad \text{for all } i\in{\cal I} \\
    \mu_j = &  \sum_{i: (i,j)\in {\cal E}} x_{i,j}, \quad \text{for all } j\in {\cal J} \\
    x_{i,j} \ge & 0, \quad \text{for all } (i,j)\in {\cal E}\\
    \lambda_i, \mu_j \in & [0,1], \quad \text{for all } i\in{\cal I}, ~j\in {\cal J},
\end{align*}
Recall that the problem is concave. By the Karush-Kuhn-Tucker (KKT) conditions~\cite{boyd2004convex}, we have
\begin{align}\label{equ:kkt-derivative}
    & \begin{bmatrix}
        \nabla_{\boldsymbol{\lambda}, \boldsymbol{\mu}} h(\boldsymbol{\lambda}, \boldsymbol{\mu}) \bigg|_{\boldsymbol{\lambda}=\boldsymbol{\lambda}^*, \boldsymbol{\mu}=\boldsymbol{\mu}^*}\\
        \boldsymbol{0}_{|{\cal E}|}
    \end{bmatrix}
    + \sum_{i\in {\cal I}} \kappa_{\mathrm{c},i} \nabla_{\boldsymbol{\lambda}, \boldsymbol{\mu},\boldsymbol{x}} \biggl(\lambda_i - \sum_{j: (i,j)\in {\cal E}} x_{i,j}\biggr)\bigg|_{\boldsymbol{\lambda}=\boldsymbol{\lambda}^*, \boldsymbol{\mu}=\boldsymbol{\mu}^*,\boldsymbol{x}=\boldsymbol{x}^*}\nonumber\\
    & + \sum_{j\in {\cal J}} \kappa_{\mathrm{s},j} \nabla_{\boldsymbol{\lambda}, \boldsymbol{\mu},\boldsymbol{x}} \biggl(\mu_j - \sum_{i: (i,j)\in {\cal E}} x_{i,j}\biggr) \bigg|_{\boldsymbol{\lambda}=\boldsymbol{\lambda}^*, \boldsymbol{\mu}=\boldsymbol{\mu}^*,\boldsymbol{x}=\boldsymbol{x}^*}\nonumber\\
    & + \sum_{(i,j)\in {\cal E}} \xi_{i,j} \nabla_{\boldsymbol{\lambda}, \boldsymbol{\mu},\boldsymbol{x}} (x_{i,j}) |_{\boldsymbol{\lambda}=\boldsymbol{\lambda}^*, \boldsymbol{\mu}=\boldsymbol{\mu}^*,\boldsymbol{x}=\boldsymbol{x}^*}\nonumber\\
    & + \sum_{i\in {\cal I}} \gamma_{\mathrm{c},i} \nabla_{\boldsymbol{\lambda}, \boldsymbol{\mu},\boldsymbol{x}}(1-\lambda_i)|_{\boldsymbol{\lambda}=\boldsymbol{\lambda}^*, \boldsymbol{\mu}=\boldsymbol{\mu}^*,\boldsymbol{x}=\boldsymbol{x}^*}
    + \sum_{j \in {\cal J}} \gamma_{\mathrm{s},j} \nabla_{\boldsymbol{\lambda}, \boldsymbol{\mu},\boldsymbol{x}} (1-\mu_j) |_{\boldsymbol{\lambda}=\boldsymbol{\lambda}^*, \boldsymbol{\mu}=\boldsymbol{\mu}^*,\boldsymbol{x}=\boldsymbol{x}^*}\nonumber\\
    & =0,
\end{align}
where $\boldsymbol{0}_{n}$ is a column vector containing $n$ zeros, $\kappa_{\mathrm{c},i}$, $\kappa_{\mathrm{s},j}$, $\xi_{i,j}$, $\gamma_{\mathrm{c},i}$, $\gamma_{\mathrm{s},j}$ are dual variables, $\xi_{i,j}\ge 0 \text{ for all } (i,j)\in {\cal E}$, $\gamma_{\mathrm{c},i}\ge 0 \text{ for all } i\in{\cal I}$, $\gamma_{\mathrm{s},j}\ge 0 \text{ for all } j\in {\cal J}$, and
\begin{align}
    \xi_{i,j} x^*_{i,j} = & 0  \text{ for all } (i,j)\in {\cal E}\label{equ:slackness-1}\\
    \gamma_{\mathrm{c},i} (1-\lambda^*_i) = &  0 \text{ for all } i\in{\cal I}\label{equ:slackness-2}\\
    \gamma_{\mathrm{s},j} (1-\mu^*_j) = & 0 \text{ for all } j\in{\cal J}.\label{equ:slackness-3}
\end{align}
The equation \eqref{equ:kkt-derivative} can be further simplified as
\begin{align}\label{equ:kkt-derivative-simp}
    & \begin{bmatrix}
        \nabla_{\boldsymbol{\lambda}, \boldsymbol{\mu}} h(\boldsymbol{\lambda}, \boldsymbol{\mu}) \bigg|_{\boldsymbol{\lambda}=\boldsymbol{\lambda}^*, \boldsymbol{\mu}=\boldsymbol{\mu}^*}\\
        \boldsymbol{0}_{|{\cal E}|}
    \end{bmatrix}
    +
    \begin{bmatrix}
        \boldsymbol{\kappa}_{\mathrm{c}}\\
        \boldsymbol{\kappa}_{\mathrm{s}}\\
        -\boldsymbol{\kappa}
    \end{bmatrix}
    +
    \begin{bmatrix}
        \boldsymbol{0}_{I}\\
        \boldsymbol{0}_{J}\\
        \boldsymbol{\xi}
    \end{bmatrix}
    +
    \begin{bmatrix}
        -\boldsymbol{\gamma}_{\mathrm{c}}\\
        -\boldsymbol{\gamma}_{\mathrm{s}}\\
        \boldsymbol{0}_{|{\cal E}|}
    \end{bmatrix}
    =0,
\end{align}
where $\boldsymbol{\kappa}_{\mathrm{c}}$, $\boldsymbol{\kappa}_{\mathrm{s}}$, $\boldsymbol{\kappa}$, $\boldsymbol{\xi}$, $\boldsymbol{\gamma}_{\mathrm{c}}$, $\boldsymbol{\gamma}_{\mathrm{s}}$ are column vectors containing $(\kappa_{\mathrm{c},i})_{i\in {\cal I}}$, $(\kappa_{\mathrm{s},j})_{j\in {\cal J}}$, $(\kappa_{\mathrm{c},i} + \kappa_{\mathrm{s},j})_{(i,j)\in {\cal E}}$, $(\xi_{i,j})_{(i,j)\in {\cal E}}$,
$(\gamma_{\mathrm{c},i})_{i\in {\cal I}}$, $(\gamma_{\mathrm{s},j})_{j\in {\cal J}}$, respectively.
Let $X_{i,j}(t)$ denotes the number of matches on link $(i,j)$ at time $t$. Let $\boldsymbol{X}(t)$ be a column vector containing $(X_{i,j}(t))_{(i,j)\in {\cal E}}$. Recall that $\boldsymbol{x}(k)$ is a column vector in the $k^{\mathrm{th}}$ iteration of the pricing algorithm.
Consider the following vector
\begin{align*}
    \boldsymbol{d} \coloneqq 
    \begin{bmatrix}
        \tilde{\boldsymbol{\alpha}}_{\mathrm{c}}(t)\\
        \tilde{\boldsymbol{\alpha}}_{\mathrm{s}}(t)\\
        \boldsymbol{x}(k(t)) - \boldsymbol{X}(t)
    \end{bmatrix}
    \mathbb{1}_{\cal C} \mathbb{1}_{{\cal H}_t}.
\end{align*}
Taking inner product between $\boldsymbol{d}$ and \eqref{equ:kkt-derivative-simp}, we obtain
\begin{align}\label{equ:inner-product-kkt}
    & \mathbb{1}_{\cal C} \mathbb{1}_{{\cal H}_t}   
     [\tilde{\boldsymbol{\alpha}}_{\mathrm{c}}(t)^\top, \tilde{\boldsymbol{\alpha}}_{\mathrm{s}}(t)^\top ]
     \nabla h (\boldsymbol{\lambda}^*, \boldsymbol{\mu}^*)
     + \mathbb{1}_{\cal C} \mathbb{1}_{{\cal H}_t}  
     \left[
     \tilde{\boldsymbol{\alpha}}_{\mathrm{c}}(t)^\top \boldsymbol{\kappa}_{\mathrm{c}}
     + \tilde{\boldsymbol{\alpha}}_{\mathrm{s}}(t)^\top \boldsymbol{\kappa}_{\mathrm{s}}
     - (\boldsymbol{x}(k(t)) - \boldsymbol{X}(t))^\top \boldsymbol{\kappa}
     \right]\nonumber\\
     & + \mathbb{1}_{\cal C} \mathbb{1}_{{\cal H}_t}  
     \left[
     (\boldsymbol{x}(k(t)) - \boldsymbol{X}(t))^\top \boldsymbol{\xi}
     - \tilde{\boldsymbol{\alpha}}_{\mathrm{c}}(t)^\top \boldsymbol{\gamma}_{\mathrm{c}}
     - \tilde{\boldsymbol{\alpha}}_{\mathrm{s}}(t)^\top \boldsymbol{\gamma}_{\mathrm{s}}
     \right]\nonumber\\
     & = 0.
\end{align}
If we pick the optimal solution $(\boldsymbol{\lambda}^*, \boldsymbol{\mu}^*, \boldsymbol{x}^*)$ in Assumption~\ref{assum:6}, we know that $x^*_{i,j} > 0$ for all $(i,j)\in {\cal E}$, $1 - \lambda^*_i > 0$ for all $i\in{\cal I}$, and $1 - \mu^*_j > 0$ for all $j \in {\cal J}$.
Then from \eqref{equ:slackness-1}-\eqref{equ:slackness-3} (complementary slackness), we know that $\xi_{i,j}=0$ for all $(i,j)\in {\cal E}$, $\gamma_{\mathrm{c},i}=0$ for all $i\in{\cal I}$, and $\gamma_{\mathrm{s},j} = 0$ for all $j \in {\cal J}$. Hence, we have
\begin{align}\label{equ:slackness-eq-0}
    \mathbb{1}_{\cal C} \mathbb{1}_{{\cal H}_t}  
     \left[
     (\boldsymbol{x}(k(t)) - \boldsymbol{X}(t))^\top \boldsymbol{\xi}
     - \tilde{\boldsymbol{\alpha}}_{\mathrm{c}}(t)^\top \boldsymbol{\gamma}_{\mathrm{c}}
     - \tilde{\boldsymbol{\alpha}}_{\mathrm{s}}(t)^\top \boldsymbol{\gamma}_{\mathrm{s}}
     \right] = 0
\end{align}
From \eqref{equ:inner-product-kkt} and \eqref{equ:slackness-eq-0}, we have
\begin{align*}
    & \mathbb{1}_{\cal C} \mathbb{1}_{{\cal H}_t}   
     [\tilde{\boldsymbol{\alpha}}_{\mathrm{c}}(t)^\top, \tilde{\boldsymbol{\alpha}}_{\mathrm{s}}(t)^\top ]
     \nabla h (\boldsymbol{\lambda}^*, \boldsymbol{\mu}^*)
     + \mathbb{1}_{\cal C} \mathbb{1}_{{\cal H}_t}  
     \left[
     \tilde{\boldsymbol{\alpha}}_{\mathrm{c}}(t)^\top \boldsymbol{\kappa}_{\mathrm{c}}
     + \tilde{\boldsymbol{\alpha}}_{\mathrm{s}}(t)^\top \boldsymbol{\kappa}_{\mathrm{s}}
     - (\boldsymbol{x}(k(t)) - \boldsymbol{X}(t))^\top \boldsymbol{\kappa}
     \right]\nonumber\\
     = & 0.
\end{align*}
Taking summation over $t=1$ to $t=T$ and taking expectation, we obtain
\begin{align}\label{equ:kkt-final}
     \eqref{equ:theo-3-regret-term-4-4-2} + \sum_{t=1}^{T} \expt \Biggl[
     \mathbb{1}_{\cal C} \mathbb{1}_{{\cal H}_t}  
     \left[
     \tilde{\boldsymbol{\alpha}}_{\mathrm{c}}(t)^\top \boldsymbol{\kappa}_{\mathrm{c}}
     + \tilde{\boldsymbol{\alpha}}_{\mathrm{s}}(t)^\top \boldsymbol{\kappa}_{\mathrm{s}}
     - (\boldsymbol{x}(k(t)) - \boldsymbol{X}(t))^\top \boldsymbol{\kappa}
     \right]
     \Biggr]
     = 0.
\end{align}
In order to bound \eqref{equ:theo-3-regret-term-4-4-2}, we need to bound the second term on the left-hand side of \eqref{equ:kkt-final}. We have the following lemma.
\begin{lemma}\label{lemma:avg-rate-balanced}
\begin{align*}
    & \sum_{t=1}^{T} \expt \Biggl[
     \mathbb{1}_{\cal C} \mathbb{1}_{{\cal H}_t}  
     \left[
     \tilde{\boldsymbol{\alpha}}_{\mathrm{c}}(t)^\top 
     (-\boldsymbol{\kappa}_{\mathrm{c}})
     + \tilde{\boldsymbol{\alpha}}_{\mathrm{s}}(t)^\top 
     (-\boldsymbol{\kappa}_{\mathrm{s}})
     + (\boldsymbol{x}(k(t)) - \boldsymbol{X}(t))^\top \boldsymbol{\kappa}
     \right]
     \Biggr] \nonumber\\
     \le &  \Theta\left(T\alpha^2 + T^2\epsilon^{\frac{\beta}{2}+1} + T\epsilon^{\frac{\beta}{2}-1} \log (1/\epsilon) 
    + T \left(\frac{\eta\epsilon}{\delta} + \eta + \delta + \epsilon \right)
    + \frac{T}{q^{\mathrm{th}}} 
    + q^{\mathrm{th}} \right).
\end{align*}
\end{lemma}
\noindent Proof of Lemma~\ref{lemma:avg-rate-balanced} can be found in Appendix~\ref{app:proof-lemma-avg-rate-balanced}.
By Lemma~\ref{lemma:avg-rate-balanced} and \eqref{equ:kkt-final}, we have
\begin{align}\label{equ:theo-3-regret-term-4-4-2-final}
    \eqref{equ:theo-3-regret-term-4-4-2} \le \Theta\left(T\alpha^2 + T^2\epsilon^{\frac{\beta}{2}+1} + T\epsilon^{\frac{\beta}{2}-1} \log (1/\epsilon) 
    + T \left(\frac{\eta\epsilon}{\delta} + \eta + \delta + \epsilon \right)
    + \frac{T}{q^{\mathrm{th}}} 
    + q^{\mathrm{th}} \right).
\end{align}

For the term \eqref{equ:theo-3-regret-term-4-4-1}, by Cauchy-Schwarz inequality, we obtain
\begin{align}\label{equ:theo-3-regret-term-4-4-1-1}
    \eqref{equ:theo-3-regret-term-4-4-1} \le &
    \expt \Biggl[ \mathbb{1}_{\cal C} \sum_{t=1}^{T} \mathbb{1}_{{\cal H}_t}
     \left\| 
     \begin{bmatrix}
         \tilde{\boldsymbol{\alpha}}_{\mathrm{c}}(t)\\
         \tilde{\boldsymbol{\alpha}}_{\mathrm{s}}(t)
     \end{bmatrix}
     \right\|_2
    \|\nabla h(\boldsymbol{\lambda}^{\mathrm{tg}}(t),
    \boldsymbol{\mu}^{\mathrm{tg}}(t) )
    - \nabla h (\boldsymbol{\lambda}^*, \boldsymbol{\mu}^*)\|_2
    \Biggr]
    \nonumber\\
    \le & \Theta(\alpha) \expt \Biggl[ \mathbb{1}_{\cal C} \sum_{t=1}^{T} \mathbb{1}_{{\cal H}_t}
    \|\nabla h(\boldsymbol{\lambda}^{\mathrm{tg}}(t),
    \boldsymbol{\mu}^{\mathrm{tg}}(t) )
    - \nabla h (\boldsymbol{\lambda}^*, \boldsymbol{\mu}^*)\|_2
    \Biggr]
    \nonumber\\
    \le & \Theta(\alpha) \expt \Biggl[ \mathbb{1}_{\cal C} \sum_{t=1}^{T} \mathbb{1}_{{\cal H}_t}
    \beta_h
    \left\|
    \begin{bmatrix}
        \boldsymbol{\lambda}^{\mathrm{tg}}(t) - \boldsymbol{\lambda}^*\\
        \boldsymbol{\mu}^{\mathrm{tg}}(t) - \boldsymbol{\mu}^*
    \end{bmatrix}
    \right\|_2
    \Biggr],
\end{align}
where the last inequality is by the smoothness property of $h$ in Lemma~\ref{lemma:h-properties}. Note that 
\begin{align}\label{equ:theo-3-regret-term-4-4-1-2}
    \left\|
    \begin{bmatrix}
        \boldsymbol{\lambda}^{\mathrm{tg}}(t) - \boldsymbol{\lambda}^*\\
        \boldsymbol{\mu}^{\mathrm{tg}}(t) - \boldsymbol{\mu}^*
    \end{bmatrix}
    \right\|_2
    \le &
    \left\|
    \begin{bmatrix}
        \boldsymbol{\lambda}^{\mathrm{tg}}(t) - \boldsymbol{\lambda}^*\\
        \boldsymbol{\mu}^{\mathrm{tg}}(t) - \boldsymbol{\mu}^*
    \end{bmatrix}
    \right\|_1\nonumber\\
    = & \sum_i \left| \sum_{j\in {\cal E}_{\mathrm{c},i}} \left(
    x_{i,j}(k(t)) - x^*_{i,j}
    \right) \right|
    + \sum_j \left| \sum_{i\in {\cal E}_{\mathrm{s},j}} \left(
    x_{i,j}(k(t)) - x^*_{i,j}
    \right) \right|\nonumber\\
    \le & 2 \sum_{(i,j)\in {\cal E}} \left|
    x_{i,j}(k(t)) - x^*_{i,j}
    \right|
    = 2 \left\| \boldsymbol{x}(k(t)) - \boldsymbol{x}^* \right\|_1\nonumber\\
    \le & 2\sqrt{|{\cal E}|} \left\| \boldsymbol{x}(k(t)) - \boldsymbol{x}^* \right\|_2,
\end{align}
where the last inequality is by Cauchy-Schwarz inequality. From \eqref{equ:theo-3-regret-term-4-4-1-1} and \eqref{equ:theo-3-regret-term-4-4-1-2}, we have
\begin{align}\label{equ:theo-3-regret-term-4-4-1-final}
    \eqref{equ:theo-3-regret-term-4-4-1} 
    \le 
    \Theta(\alpha) \expt \Biggl[ \mathbb{1}_{\cal C} \sum_{t=1}^{T} \mathbb{1}_{{\cal H}_t}
    \left\| \boldsymbol{x}(k(t)) - \boldsymbol{x}^* \right\|_2
    \Biggr] 
    \le \Theta(\alpha) \expt \Biggl[ \mathbb{1}_{\cal C} \sum_{t=1}^{T}
    \left\| \boldsymbol{x}(k(t)) - \boldsymbol{x}^* \right\|_2
    \Biggr].
\end{align}
Combining \eqref{equ:theo-3-regret-term-4-4-1}, \eqref{equ:theo-3-regret-term-4-4-2}, \eqref{equ:theo-3-regret-term-4-4-2-final}, and \eqref{equ:theo-3-regret-term-4-4-1-final}, we have
\begin{align}\label{equ:theo-3-regret-term-4-4-final}
    \eqref{equ:theo-3-regret-term-4-4} 
    \le & \Theta(\alpha) \expt \Biggl[ \mathbb{1}_{\cal C} \sum_{t=1}^{T}
    \left\| \boldsymbol{x}(k(t)) - \boldsymbol{x}^* \right\|_2
    \Biggr]\nonumber\\
    & + \Theta\left(T\alpha^2 + T^2\epsilon^{\frac{\beta}{2}+1} + T\epsilon^{\frac{\beta}{2}-1} \log (1/\epsilon) 
    + T \left(\frac{\eta\epsilon}{\delta} + \eta + \delta + \epsilon \right)
    + \frac{T}{q^{\mathrm{th}}} 
    + q^{\mathrm{th}} \right).
\end{align}
Combining \eqref{equ:theo-3-regret-term-4-3}, \eqref{equ:theo-3-regret-term-4-4}, \eqref{equ:theo-3-regret-term-4-5}, and \eqref{equ:theo-3-regret-term-4-4-final}, we have
\begin{align}\label{equ:theo-3-regret-term-4-final}
    \eqref{equ:theo-3-regret-term-4} 
    \le & \expt \Biggl[ \mathbb{1}_{\cal C} \sum_{t=1}^{T} \mathbb{1}_{{\cal H}_t}   \bigl[ h(\boldsymbol{\lambda}^*, \boldsymbol{\mu}^*)
     -  h(\boldsymbol{\lambda}^{\mathrm{tg}}(t), \boldsymbol{\mu}^{\mathrm{tg}}(t)) \bigr]  \Biggr]\nonumber\\
     & + \Theta(\alpha) \expt \Biggl[ \mathbb{1}_{\cal C} \sum_{t=1}^{T}
    \left\| \boldsymbol{x}(k(t)) - \boldsymbol{x}^* \right\|_2
    \Biggr]\nonumber\\
    & + \Theta\left(T\alpha^2 + T^2\epsilon^{\frac{\beta}{2}+1} + T\epsilon^{\frac{\beta}{2}-1} \log (1/\epsilon) 
    + T \left(\frac{\eta\epsilon}{\delta} + \eta + \delta + \epsilon \right)
    + \frac{T}{q^{\mathrm{th}}} 
    + q^{\mathrm{th}} \right).
\end{align}
Combining \eqref{equ:theo-3-regret-term-3}, \eqref{equ:theo-3-regret-term-4}, \eqref{equ:theo-3-regret-term-3-final}, and \eqref{equ:theo-3-regret-term-4-final}, we have
\begin{align}\label{equ:theo-3-regret-term-1-1}
    \eqref{equ:theo-3-regret-term-1}
    \le &
    \expt \Biggl[ \mathbb{1}_{\cal C} \sum_{t=1}^{T} \mathbb{1}_{{\cal H}_t}   \bigl[ h(\boldsymbol{\lambda}^*, \boldsymbol{\mu}^*)
     -  h(\boldsymbol{\lambda}^{\mathrm{tg}}(t), \boldsymbol{\mu}^{\mathrm{tg}}(t)) \bigr]  \Biggr]\nonumber\\
     & + \Theta(\alpha) \expt \Biggl[ \mathbb{1}_{\cal C} \sum_{t=1}^{T}
    \left\| \boldsymbol{x}(k(t)) - \boldsymbol{x}^* \right\|_2
    \Biggr]\nonumber\\
    & + \Theta\left(T\alpha^2 + T^2\epsilon^{\frac{\beta}{2}+1} + T\epsilon^{\frac{\beta}{2}-1} \log (1/\epsilon) 
    + T \left(\frac{\eta\epsilon}{\delta} + \eta + \delta + \epsilon \right)
    + \frac{T}{q^{\mathrm{th}}} 
    + q^{\mathrm{th}} \right).
\end{align}
Since $h(\boldsymbol{\lambda}^*, \boldsymbol{\mu}^*)
-  h(\boldsymbol{\lambda}^{\mathrm{tg}}(t), \boldsymbol{\mu}^{\mathrm{tg}}(t)) \ge 0$, \eqref{equ:theo-3-regret-term-1-1} can be further bounded by
\begin{align}
    \eqref{equ:theo-3-regret-term-1}
    \le & \expt \Biggl[ \mathbb{1}_{\cal C} \sum_{t=1}^{T} 
    \bigl[ h(\boldsymbol{\lambda}^*, \boldsymbol{\mu}^*)
     -  h(\boldsymbol{\lambda}^{\mathrm{tg}}(t), \boldsymbol{\mu}^{\mathrm{tg}}(t)) \bigr]  \Biggr]
    + \Theta(\alpha) \expt \Biggl[ \mathbb{1}_{\cal C} \sum_{t=1}^{T}
    \left\| \boldsymbol{x}(k(t)) - \boldsymbol{x}^* \right\|_2
    \Biggr]\nonumber\\
    & + \Theta\left(T\alpha^2 + T^2\epsilon^{\frac{\beta}{2}+1} + T\epsilon^{\frac{\beta}{2}-1} \log (1/\epsilon) 
    + T \left(\frac{\eta\epsilon}{\delta} + \eta + \delta + \epsilon \right)
    + \frac{T}{q^{\mathrm{th}}} 
    + q^{\mathrm{th}} \right)\nonumber\\
    = & \expt \Biggl[ \mathbb{1}_{\cal C} \sum_{t=1}^{T} 
    \bigl[ f(\boldsymbol{x}^*)
     -  f(\boldsymbol{x}(k(t))) \bigr]  \Biggr]
    + \Theta(\alpha) \expt \Biggl[ \mathbb{1}_{\cal C} \sum_{t=1}^{T}
    \left\| \boldsymbol{x}(k(t)) - \boldsymbol{x}^* \right\|_2
    \Biggr]\label{equ:theo-3-regret-term-1-final-1}\\
    & + \Theta\left(T\alpha^2 + T^2\epsilon^{\frac{\beta}{2}+1} + T\epsilon^{\frac{\beta}{2}-1} \log (1/\epsilon) 
    + T \left(\frac{\eta\epsilon}{\delta} + \eta + \delta + \epsilon \right)
    + \frac{T}{q^{\mathrm{th}}} 
    + q^{\mathrm{th}} \right).\label{equ:theo-3-regret-term-1-final-2}
\end{align}

\subsection{Bounding the Regret during Optimization}
In this subsection, we will bound the two terms in \eqref{equ:theo-3-regret-term-1-final-1}, which are the regret during optimization.

One main difficulty of bounding these two terms in \eqref{equ:theo-3-regret-term-1-final-1} is that the number of outer iterations is random and the number of time slots in each iteration is also random because we count the sample only when there is no arrival rate reduction and we need to collect at least $N$ samples for each queue for each bisection iteration. The reason why we only count the samples without arrival rate reduction is that the samples with arrival rate reduction are biased for estimating the arrival rates corresponding to the prices used in the bisection.
In order to bound \eqref{equ:theo-3-regret-term-1-final-1}, we need to bound the number of time slots in each bisection iteration.
In this subsection, we will first bound the number of time slots in each bisection iteration and then bound the regret over the gradient steps.

We first decompose the each of the regret terms in \eqref{equ:theo-3-regret-term-1-final-1}.
Let 
\[
{\cal T}^+(k,m)\coloneqq \{t: k(t)=k, m(t)=m, \mbox{ the bisection search is for } \boldsymbol{\lambda}^+(k) \mbox{ and } \boldsymbol{\mu}^+(k) \}
\]
and
\[
{\cal T}^-(k,m)\coloneqq \{t: k(t)=k, m(t)=m, \mbox{ the bisection search is for } \boldsymbol{\lambda}^-(k) \mbox{ and } \boldsymbol{\mu}^-(k) \}.
\]
Then
\begin{align*}
    & \expt \Biggl[ \mathbb{1}_{\cal C} \sum_{t=1}^{T} 
    \bigl[ f(\boldsymbol{x}^*)
     -  f(\boldsymbol{x}(k(t))) \bigr]  \Biggr]\nonumber\\
     = & \expt \Biggl[\mathbb{1}_{\cal C}
     \sum_{k=1}^{k(T)} \sum_{m=1}^{M}
    |{\cal T}^+(k,m)|
     \bigl[ f(\boldsymbol{x}^*)
     -  f(\boldsymbol{x}(k) \bigr]
     \Biggr]
      + \expt \Biggl[\mathbb{1}_{\cal C}
     \sum_{k=1}^{k(T)} \sum_{m=1}^{M}
     |{\cal T}^-(k,m)|
     \bigl[ f(\boldsymbol{x}^*)
     -  f(\boldsymbol{x}(k) \bigr]
     \Biggr].
\end{align*}
Note that $k(T)\le \lceil T/2MN \rceil$. Then we have
\begin{align}\label{equ:opti-error-decomp-1}
    & \expt \Biggl[ \mathbb{1}_{\cal C} \sum_{t=1}^{T} 
    \bigl[ f(\boldsymbol{x}^*)
     -  f(\boldsymbol{x}(k(t))) \bigr]  \Biggr]\nonumber\\
     \le & 
     \sum_{k=1}^{\lceil T/2MN \rceil} \sum_{m=1}^{M}
     \expt \biggl[\mathbb{1}_{\cal C}|{\cal T}^+(k,m)|
     \bigl[ f(\boldsymbol{x}^*)
     -  f(\boldsymbol{x}(k) \bigr]
     \biggr]\nonumber\\
     & + 
     \sum_{k=1}^{\lceil T/2MN \rceil} \sum_{m=1}^{M}
     \expt \biggl[\mathbb{1}_{\cal C} |{\cal T}^-(k,m)|
     \bigl[ f(\boldsymbol{x}^*)
     -  f(\boldsymbol{x}(k) \bigr]
     \biggr]\nonumber\\
     \le & \sum_{k=1}^{\lceil T/2MN \rceil} \sum_{m=1}^{M}
     \expt \biggl[|{\cal T}^+(k,m)|
     \bigl[ f(\boldsymbol{x}^*)
     -  f(\boldsymbol{x}(k) \bigr]
     \bigg| {\cal C}
     \biggr]\nonumber\\
     & + 
     \sum_{k=1}^{\lceil T/2MN \rceil} \sum_{m=1}^{M}
     \expt \biggl[ |{\cal T}^-(k,m)|
     \bigl[ f(\boldsymbol{x}^*)
     -  f(\boldsymbol{x}(k) \bigr]
     \bigg| {\cal C}
     \biggr],
\end{align}
where the last inequality is by law of total expectation. Similarly, we have
\begin{align}\label{equ:opti-error-decomp-1-other}
    & \expt \Biggl[ \mathbb{1}_{\cal C} \sum_{t=1}^{T}
    \left\| \boldsymbol{x}(k(t)) - \boldsymbol{x}^* \right\|_2
    \Biggr]\nonumber\\
    \le & \sum_{k=1}^{\lceil T/2MN \rceil} \sum_{m=1}^{M} \expt \biggl[
    |{\cal T}^+(k,m)|
     \left\| \boldsymbol{x}(k) - \boldsymbol{x}^* \right\|_2
     \bigg| {\cal C}
     \biggr]
      + \sum_{k=1}^{\lceil T/2MN \rceil} \sum_{m=1}^{M}
      \expt \biggl[
     |{\cal T}^-(k,m)|
     \left\| \boldsymbol{x}(k) - \boldsymbol{x}^* \right\|_2
     \bigg| {\cal C}
     \biggr].
\end{align}

\subsubsection{Bounding the Number of Time Slots in Each Bisection Iteration}
\label{app:sec:bound-num-slots-bisection}
We will first bound the terms $|{\cal T}^+(k,m)|$ and $|{\cal T}^-(k,m)|$.
Note that under the \textit{probabilistic two-price policy}, in each bisection iteration, we need to collect at least $N$ samples for each queue and the arrival rates are reduced randomly with probability $1/2$ if the queue is nonempty. Let $t^+_{k,m}$ denote the first time slot of the $k^{\mathrm{th}}$ outer iteration and the $m^{\mathrm{th}}$ bisection iteration for $(\boldsymbol{\lambda}^+(k), \boldsymbol{\mu}^+(k))$. Define $t^-_{k,m}$ similarly for $(\boldsymbol{\lambda}^-(k), \boldsymbol{\mu}^-(k))$.
Hence, $|{\cal T}^+(k,m)|$ can be rewritten as
\begin{align*}
    |{\cal T}^+(k,m)|
    = & \min\Biggl\{t: \min\Biggl\{  
    \min_i \sum_{t'=0}^{t-1} \mathbb{1}\{Q_{\mathrm{c},i}(t^+_{k,m} + t') = 0 \mbox{ or } B_{\mathrm{c},i}(t_{k,m} + t') = 0 \}, \nonumber\\
    & \qquad \qquad \qquad \min_j \sum_{t'=0}^{t-1} \mathbb{1}\{Q_{\mathrm{s},j}(t_{k,m} + t') = 0 \mbox{ or } B_{\mathrm{s},j}(t^+_{k,m} + t') = 0\}
    \Biggr\} = N
    \Biggr\},
\end{align*}
where we recall that $B_{\mathrm{c},i}(t)$ and $B_{\mathrm{s},j}(t)$ are i.i.d. Bernoulli random variables with parameter $1/2$. Dropping $\hat{Q}_{\mathrm{c},i}(t^+_{k,m} +t') = 0$ and $\hat{Q}_{\mathrm{s},j}(t^+_{k,m} +t') = 0$ gives us an upper bound for $|{\cal T}^+(k,m)|$, i.e.,
\begin{align}\label{equ:num-slots-bisection-bound-1}
    & |{\cal T}^+(k,m)| \nonumber\\
    \le & \min\Biggl\{t: \min\Biggl\{  
    \min_i \sum_{t'=0}^{t-1} \mathbb{1}\{ B_{\mathrm{c},i}(t^+_{k,m} +t') = 0 \},
    \min_j \sum_{t'=0}^{t-1} \mathbb{1}\{ B_{\mathrm{s},j}(t^+_{k,m} +t') = 0\}
    \Biggr\} = N
    \Biggr\}\nonumber\\
    = & \min\Biggl\{t: \min\Biggl\{  
    \min_i \sum_{t'=0}^{t-1} (1 - B_{\mathrm{c},i}(t^+_{k,m} +t') ),
    \min_j \sum_{t'=0}^{t-1} (1 - B_{\mathrm{s},j}(t^+_{k,m} +t') )
    \Biggr\} = N
    \Biggr\}.
\end{align}
Denote the right-hand side of \eqref{equ:num-slots-bisection-bound-1} by $\tau^+_{k,m}$. Then
\begin{align*}
    |{\cal T}^+(k,m)| \le \tau^+_{k,m}.
\end{align*}
Then for the first term in \eqref{equ:opti-error-decomp-1}, we have
\begin{align}\label{equ:opti-error-decomp-2}
    & \sum_{k=1}^{\lceil T/2MN \rceil} \sum_{m=1}^{M}
     \expt \biggl[|{\cal T}^+(k,m)|
     \bigl[ f(\boldsymbol{x}^*)
     -  f(\boldsymbol{x}(k) \bigr]
     \bigg| {\cal C}
     \biggr]\nonumber\\
     \le & \sum_{k=1}^{\lceil T/2MN \rceil} \sum_{m=1}^{M}
     \expt \biggl[ \tau^+_{k,m}
     \bigl[ f(\boldsymbol{x}^*)
     -  f(\boldsymbol{x}(k) \bigr]
     \bigg| {\cal C}
     \biggr].
\end{align}
Note that $\boldsymbol{x}(k)$ is independent of $\tau^+_{k,m}$. Also note that $\tau^+_{k,m}$ is independent of the event ${\cal C}$. Hence, given the event ${\cal C}$, $\boldsymbol{x}(k)$ is independent of $\tau^+_{k,m}$. Then from \eqref{equ:opti-error-decomp-2} we have
\begin{align}\label{equ:opti-error-decomp-3}
    & \sum_{k=1}^{\lceil T/2MN \rceil} \sum_{m=1}^{M}
     \expt \biggl[|{\cal T}^+(k,m)|
     \bigl[ f(\boldsymbol{x}^*)
     -  f(\boldsymbol{x}(k) \bigr]
     \bigg| {\cal C}
     \biggr]\nonumber\\
     \le & \sum_{k=1}^{\lceil T/2MN \rceil} \sum_{m=1}^{M}
     \expt \biggl[ \tau^+_{k,m}\biggr]
     \expt \biggl[\bigl[ f(\boldsymbol{x}^*)
     -  f(\boldsymbol{x}(k) \bigr]
     \bigg| {\cal C}
     \biggr].
\end{align}
Define $\tau^+_{k,m,\mathrm{c},i}$ by
\[
\tau^+_{k,m,\mathrm{c},i} \coloneqq \min\Biggl\{t: \sum_{t'=0}^{t-1} (1 - B_{\mathrm{c},i}(t^+_{k,m} +t') ) = N
    \Biggr\}
\]
for all $i$
and define $\tau^+_{k,m,\mathrm{s},j}$ by
\[
\tau^+_{k,m,\mathrm{s},j} \coloneqq \min\Biggl\{t: 
    \sum_{t'=0}^{t-1} (1 - B_{\mathrm{s},j}(t^+_{k,m} +t') ) = N
    \Biggr\}
\]
for all $j$. Then by the definition of $\tau^+_{k,m}$, we know
\begin{align}\label{equ:num-slots-bisection-bound-2}
    \tau^+_{k,m} = \max\{ \max_i \tau^+_{k,m,\mathrm{c},i}, \max_j \tau^+_{k,m,\mathrm{s},j}\} \le \sum_i \tau^+_{k,m,\mathrm{c},i} + \sum_j \tau^+_{k,m,\mathrm{s},j}.
\end{align}
By the definition of $\tau^+_{k,m,\mathrm{c},i}$, we have
\[
\sum_{t'=0}^{\tau^+_{k,m,\mathrm{c},i}-1} (1 - B_{\mathrm{c},i}(t^+_{k,m} + t') ) = N.
\]
Taking expectation on both sides, we have
\[
\expt \Biggl[\sum_{t'=0}^{\tau^+_{k,m,\mathrm{c},i}-1} (1 - B_{\mathrm{c},i}(t^+_{k,m} + t') )\Biggr] = N.
\]
Note that $(B_{\mathrm{c},i}(t^+_{k,m} + t'))_{t'=0}^{\infty}$ are i.i.d. Also note that $\tau^+_{k,m,\mathrm{c},i}$ is a stopping time. By Wald's lemma, we have
\begin{align*}
    N = \expt \Biggl[\sum_{t'=0}^{\tau^+_{k,m,\mathrm{c},i}-1} (1 - B_{\mathrm{c},i}(t^+_{k,m} + t') )\Biggr] = \expt [\tau^+_{k,m,\mathrm{c},i}] \expt[B_{\mathrm{c},i}(1)].
\end{align*}
Hence, we have
\begin{align}\label{equ:num-slots-bisection-bound-3}
    \expt [\tau^+_{k,m,\mathrm{c},i}] = \frac{N}{\expt[B_{\mathrm{c},i}(1)]} = 2N,
\end{align}
which holds for all $i$.
Similarly, we have
\begin{align}\label{equ:num-slots-bisection-bound-4}
    \expt [\tau^+_{k,m,\mathrm{s},j}] = 2N
\end{align}
for all $j$.
Then from \eqref{equ:num-slots-bisection-bound-2}, \eqref{equ:num-slots-bisection-bound-3}, and \eqref{equ:num-slots-bisection-bound-4}, we have
\begin{align}\label{equ:num-slots-bisection-bound-final}
    \expt [\tau^+_{k,m}] \le  \expt \biggl[ \sum_i \tau^+_{k,m,\mathrm{c},i} + \sum_j \tau^+_{k,m,\mathrm{s},j}\biggr]
    \le   2(I+J) N.
\end{align}
From \eqref{equ:opti-error-decomp-3} and \eqref{equ:num-slots-bisection-bound-final}, we have
\begin{align}\label{equ:opti-error-decomp-4}
    & \sum_{k=1}^{\lceil T/2MN \rceil} \sum_{m=1}^{M}
     \expt \biggl[|{\cal T}^+(k,m)|
     \bigl[ f(\boldsymbol{x}^*)
     -  f(\boldsymbol{x}(k) \bigr]
     \bigg| {\cal C}
     \biggr]\nonumber\\
     \le & 2(I+J) M N \sum_{k=1}^{\lceil T/2MN \rceil} 
     \expt \bigl[ f(\boldsymbol{x}^*)
     -  f(\boldsymbol{x}(k)
     \big| {\cal C}
     \bigr]
\end{align}
Similarly, we can obtain
\begin{align}\label{equ:opti-error-decomp-5}
    & \sum_{k=1}^{\lceil T/2MN \rceil} \sum_{m=1}^{M}
     \expt \biggl[ |{\cal T}^-(k,m)|
     \bigl[ f(\boldsymbol{x}^*)
     -  f(\boldsymbol{x}(k) \bigr]
     \bigg| {\cal C}
     \biggr] \nonumber\\
     \le & 2(I+J) M N
     \sum_{k=1}^{\lceil T/2MN \rceil} 
     \expt
     \bigl[ f(\boldsymbol{x}^*)
     -  f(\boldsymbol{x}(k) 
     \big| {\cal C}
     \bigr],
\end{align}
\begin{align}\label{equ:opti-error-decomp-6}
    & \sum_{k=1}^{\lceil T/2MN \rceil} \sum_{m=1}^{M} \expt \biggl[
    |{\cal T}^+(k,m)|
     \left\| \boldsymbol{x}(k) - \boldsymbol{x}^* \right\|_2
     \bigg| {\cal C}
     \biggr] \nonumber\\
     \le & 2(I+J) M N
     \sum_{k=1}^{\lceil T/2MN \rceil}
     \expt
     \bigl[ \left\| \boldsymbol{x}(k) - \boldsymbol{x}^* \right\|_2 
     \big| {\cal C}
     \bigr],
\end{align}
and
\begin{align}\label{equ:opti-error-decomp-7}
    & \sum_{k=1}^{\lceil T/2MN \rceil} \sum_{m=1}^{M} \expt \biggl[
    |{\cal T}^-(k,m)|
     \left\| \boldsymbol{x}(k) - \boldsymbol{x}^* \right\|_2
     \bigg| {\cal C}
     \biggr] \nonumber\\
     \le & 2(I+J) M N
     \sum_{k=1}^{\lceil T/2MN \rceil}
     \expt
     \bigl[ \left\| \boldsymbol{x}(k) - \boldsymbol{x}^* \right\|_2 
     \big| {\cal C}
     \bigr],
\end{align}
From \eqref{equ:theo-3-regret-term-1-final-1}, \eqref{equ:opti-error-decomp-1}, \eqref{equ:opti-error-decomp-1-other}, \eqref{equ:opti-error-decomp-4}-\eqref{equ:opti-error-decomp-7}, we have
\begin{align}\label{equ:opti-error-decomp-final}
     \eqref{equ:theo-3-regret-term-1-final-1}
     \le & 4(I+J) M N \sum_{k=1}^{\lceil T/2MN \rceil}
     \expt \bigl[ f(\boldsymbol{x}^*)
     -  f(\boldsymbol{x}(k)
     \big| {\cal C}
     \bigr] \nonumber\\
     & + 4(I+J) M N \Theta(\alpha)
     \sum_{k=1}^{\lceil T/2MN \rceil}
     \expt
     \bigl[ \left\| \boldsymbol{x}(k) - \boldsymbol{x}^* \right\|_2 
     \big| {\cal C}
     \bigr].
\end{align}

\subsubsection{Bounding the Regret Over the Gradient Steps}

For the term $\sum_{k=1}^{\lceil T/2MN \rceil} \expt [ \| \boldsymbol{x}(k) - \boldsymbol{x}^* \|_2 | {\cal C} ]$, by Cauchy-Schwarz inequality, we have
\begin{align*}
    &\sum_{k=1}^{\lceil T/2MN \rceil}
    \expt \Bigl[ \left\| \boldsymbol{x}(k) - \boldsymbol{x}^* \right\|_2
    \big| {\cal C}
    \Bigr] \nonumber\\
    \le & \sqrt{\lceil T/2MN \rceil} \sqrt{\sum_{k=1}^{\lceil T/2MN \rceil}
    \left(\expt \Bigl[ \left\| \boldsymbol{x}(k) - \boldsymbol{x}^* \right\|_2
    \big| {\cal C}
    \Bigr]\right)^2
    }\nonumber\\
    \le & \sqrt{\lceil T/2MN \rceil} \sqrt{\sum_{k=1}^{\lceil T/2MN \rceil}
    \expt \Bigl[ \left\| \boldsymbol{x}(k) - \boldsymbol{x}^* \right\|_2^2
    \big| {\cal C}
    \Bigr]},
\end{align*}
where the last inequality is by Jensen's inequality. By Assumption~\ref{assum:6}, we know that $\| \boldsymbol{x} - \boldsymbol{x}^* \|_2^2 \le \frac{2}{\nu} [f(\boldsymbol{x}^*) - f(\boldsymbol{x})]$ for all $x\in {\cal D}'$. Hence, we can further obtain
\begin{align}\label{equ:opti-error-decomp-final-2}
    \sum_{k=1}^{\lceil T/2MN \rceil}
    \expt \Bigl[ \left\| \boldsymbol{x}(k) - \boldsymbol{x}^* \right\|_2
    \big| {\cal C}
    \Bigr]
    \le \sqrt{\frac{2}{\nu} \lceil T/2MN \rceil} \sqrt{\sum_{k=1}^{\lceil T/2MN \rceil}
    \expt \Bigl[ f(\boldsymbol{x}^*) - f(\boldsymbol{x}(k))
    \big| {\cal C}
    \Bigr]}.
\end{align}
From \eqref{equ:opti-error-decomp-final} and \eqref{equ:opti-error-decomp-final-2}, we have
\begin{align}
    \eqref{equ:theo-3-regret-term-1-final-1}
    \le & \Theta\left(MN\right) \expt \Biggl[
     \sum_{k=1}^{\lceil T/2MN \rceil}
     \bigl[ f(\boldsymbol{x}^*)
     -  f(\boldsymbol{x}(k) \bigr]
     \bigg| {\cal C}
     \Biggr]\label{equ:opti-error-1}\\
     & + \Theta\left(\alpha \sqrt{TMN} \right) \sqrt{\sum_{k=1}^{\lceil T/2MN \rceil}
    \expt \Bigl[ f(\boldsymbol{x}^*) - f(\boldsymbol{x}(k))
    \big| {\cal C}
    \Bigr]}.\label{equ:opti-error-2}
\end{align}
We will next use a lemma from \cite{yang2024learning}:
\begin{lemma}[\cite{yang2024learning}]
\label{lemma:optimization}
    Let Assumption~\ref{assum:1}--\ref{assum:3} hold. Then for any fixed integer $K$, we have
    \begin{align*}
        \sum_{k=1}^{K} \expt \Bigl[f(\boldsymbol{x}^*) - f(\boldsymbol{x}(k))
        \Bigl|\Bigr. {\cal C} \Bigr] 
        \le  \Theta \left( \frac{1}{\eta} +  K\eta + \frac{K\eta \epsilon^2}{\delta^2} + \frac{K\epsilon}{\delta} + K\delta\right),
    \end{align*}
    where in the notation $\Theta(\cdot)$, we omit the variables that do not depend on $T$ or $K$.
\end{lemma}
Lemma~\ref{lemma:optimization} bounds the sum of suboptimality gaps over outer iterations. By Lemma~\ref{lemma:optimization}, \eqref{equ:opti-error-1} can be bounded by 
\begin{align}\label{equ:opti-error-1-final}
    \eqref{equ:opti-error-1} 
    \le & \Theta\left( \frac{MN}{\eta} 
    + T \left(\eta + \frac{\eta \epsilon^2}{\delta^2} + \frac{\epsilon}{\delta} + \delta\right) \right)\nonumber\\
    \le & \Theta\left( \frac{\log^2 (1/\epsilon) }{\eta \epsilon^2 } 
    + T \left( \frac{\epsilon}{\delta} + \eta + \delta \right) \right),
\end{align}
where the last inequality is by substituting the definition of $M$ and $N$ and $\epsilon < \delta$. Similarly, \eqref{equ:opti-error-2} can be bounded by
\begin{align}\label{equ:opti-error-2-final}
    \eqref{equ:opti-error-2} 
    \le & \Theta\left(
    \alpha \sqrt{TMN}
    \sqrt{\frac{1}{\eta} + \frac{T\eta}{MN} + \frac{T\epsilon}{MN \delta} + \frac{T\delta}{MN}}
    \right)\nonumber\\
    \le & \Theta\left(
    \frac{\alpha \sqrt{T} \log (1/\epsilon)}{\epsilon \sqrt{\eta}}
    + T \alpha \left( \sqrt{\eta} + \sqrt{\frac{\epsilon}{\delta}} + \sqrt{\delta}\right)
    \right).
\end{align}
Combining \eqref{equ:opti-error-1}, \eqref{equ:opti-error-2}, \eqref{equ:opti-error-1-final}, and \eqref{equ:opti-error-2-final}, we obtain
\begin{align}\label{equ:theo-3-regret-term-1-final-1-bound}
    \eqref{equ:theo-3-regret-term-1-final-1}
    \le & \Theta\left( \frac{\log^2 (1/\epsilon) }{\eta \epsilon^2 } 
    + T \left( \frac{\epsilon}{\delta} + \eta + \delta \right) 
    + \frac{\alpha \sqrt{T} \log (1/\epsilon)}{\epsilon \sqrt{\eta}}
    + T \alpha \left( \sqrt{\eta} + \sqrt{\frac{\epsilon}{\delta}} + \sqrt{\delta}\right)
    \right)
\end{align}
Then from \eqref{equ:theo-3-regret-term-1-final-1}, \eqref{equ:theo-3-regret-term-1-final-2}, and \eqref{equ:theo-3-regret-term-1-final-1-bound}, we have
\begin{align}\label{equ:theo-3-regret-term-1-bound}
    \eqref{equ:theo-3-regret-term-1} 
    \le & 
    \Theta\Biggl( \frac{\log^2 (1/\epsilon) }{\eta \epsilon^2 } 
    + T \left( \frac{\epsilon}{\delta} + \eta + \delta \right) 
    +
    \frac{\alpha \sqrt{T} \log (1/\epsilon)}{\epsilon \sqrt{\eta}}
    + T \alpha \left( \sqrt{\eta} + \sqrt{\frac{\epsilon}{\delta}} + \sqrt{\delta}\right) \Biggr)\nonumber\\
    & + \Theta\left(T\alpha^2 + T^2\epsilon^{\frac{\beta}{2}+1} + T\epsilon^{\frac{\beta}{2}-1} \log (1/\epsilon) 
    + \frac{T}{q^{\mathrm{th}}} 
    + q^{\mathrm{th}} \right).
\end{align}
From \eqref{equ:theo-3-regret-term-1}, \eqref{equ:theo-3-regret-term-2-bound}, and \eqref{equ:theo-3-regret-term-1-bound}, we have
\begin{align*}
    \expt [R(T)] 
    = & O \Biggl( \frac{\log^2 (1/\epsilon) }{\eta \epsilon^2 } 
    + T \left( \frac{\epsilon}{\delta} + \eta + \delta \right) 
    +
    \frac{\alpha \sqrt{T} \log (1/\epsilon)}{\epsilon \sqrt{\eta}}
    + T \alpha \left( \sqrt{\eta} + \sqrt{\frac{\epsilon}{\delta}} + \sqrt{\delta}\right)
    \nonumber\\
    & \qquad
    + T\alpha^2 + T^2\epsilon^{\frac{\beta}{2}+1} + T\epsilon^{\frac{\beta}{2}-1} \log (1/\epsilon) 
    + \frac{T}{q^{\mathrm{th}}} 
    + q^{\mathrm{th}} \Biggr).
\end{align*}
The regret bound~\eqref{equ:regret-bound-3} in Theorem~\ref{theo:3} is proved.

\subsection{Bounding Average Queue Length}
\label{app:sec:bound-avg-queue}

In this subsection, we will prove the average queue length bound \eqref{equ:avg-queue-length-bound-3}. Consider Lyapunov function $\sum_i Q_{\mathrm{c}, i}^2 (t)$.
Using the same argument as that in the proof of Lemma~\ref{lemma:theo-3-bound-time-large-queue} in Appendix~\ref{app:proof-lemma-theo-3-bound-time-large-queue}, we have
\begin{align}\label{equ:avg-queue-drift-1}
    & \expt \left[\sum_i Q_{\mathrm{c},i}^2(t+1) - \sum_i Q_{\mathrm{c},i}^2(t) | \boldsymbol{Q}_{\mathrm{c}}(t) \right] \nonumber\\
    \le & 
     \sum_i |{\cal E}_{\mathrm{c},i}|^2 + 2 \sum_i Q_{\mathrm{c},i}(t) \biggl( \expt [ {A}_{\mathrm{c},i}(t) | \boldsymbol{Q}_{\mathrm{c}}(t)] -  \expt \biggl[\sum_{j\in {\cal E}_{\mathrm{c},i}} {X}_{i,j}(t) \biggl|\biggr. \boldsymbol{Q}_{\mathrm{c}}(t) \biggr] \biggr),
\end{align}
where
\begin{align}\label{equ:avg-queue-arrival-term-1}
    & \expt [  {A}_{\mathrm{c},i}(t) |  {\boldsymbol{Q}}_{\mathrm{c}}(t) ] \nonumber\\
    = &  \mathbb{1} \left\{  {Q}_{\mathrm{c},i}(t) = 0 \right\}
     \expt \left[ {\lambda}'_i(t) |  {\boldsymbol{Q}}_{\mathrm{c}}(t) \right]
    + \mathbb{1} \left\{ 0 < {Q}_{\mathrm{c},i}(t) < q^{\mathrm{th}} \right\}
     \expt \left[ {\lambda}_{\alpha,i}(t) |  {\boldsymbol{Q}}_{\mathrm{c}}(t) \right]
\end{align}
and
\begin{align}\label{equ:avg-queue-service-term}
    \sum_i  {Q}_{\mathrm{c},i}(t) \expt \biggl[\sum_{j\in {\cal E}_{\mathrm{c},i}}  {X}_{i,j}(t) \biggl|\biggr.  {\boldsymbol{Q}}_{\mathrm{c}}(t)  \biggr]
    = & \sum_j
    \expt \left[  {\mu}'_j(t)  {Q}_{\mathrm{c},i^*_j(t)}(t) \Bigl|\bigr.  {\boldsymbol{Q}}_{\mathrm{c}}(t)  \right].
\end{align}
From \eqref{equ:avg-queue-arrival-term-1}, we have
\begin{align*}
    & \expt [  {A}_{\mathrm{c},i}(t) |  {\boldsymbol{Q}}_{\mathrm{c}}(t) ] \nonumber\\
    \le & \mathbb{1} \left\{  {Q}_{\mathrm{c},i}(t) = 0 \right\}
     \expt \left[ {\lambda}'_i(t) |  {\boldsymbol{Q}}_{\mathrm{c}}(t) \right]
    + \mathbb{1} \left\{ {Q}_{\mathrm{c},i}(t) > 0 \right\}
     \expt \left[ {\lambda}_{\alpha,i}(t) |  {\boldsymbol{Q}}_{\mathrm{c}}(t) \right].
\end{align*}
Recall from \eqref{equ:taylor-final} that
\[
|\max\{\lambda'_i(t) - \tilde{\alpha}_{\mathrm{c},i}(t), 0\} 
- \lambda_{\alpha,i}(t)| 
\le  \Theta(\alpha^2) \mbox{ for all } i.
\]
Hence, we have
\begin{align}\label{equ:avg-queue-arrival-term-2}
    & \expt [  {A}_{\mathrm{c},i}(t) |  {\boldsymbol{Q}}_{\mathrm{c}}(t) ] \nonumber\\
    \le & \mathbb{1} \left\{  {Q}_{\mathrm{c},i}(t) = 0 \right\}
     \expt \left[ {\lambda}'_i(t) |  {\boldsymbol{Q}}_{\mathrm{c}}(t) \right]
    + \mathbb{1} \left\{ {Q}_{\mathrm{c},i}(t) > 0 \right\}
     \expt \left[ \max\{\lambda'_i(t) - \tilde{\alpha}_{\mathrm{c},i}(t), 0\} |  {\boldsymbol{Q}}_{\mathrm{c}}(t) \right] + \Theta(\alpha^2).
\end{align}
Hence, from \eqref{equ:avg-queue-drift-1}, \eqref{equ:avg-queue-service-term}, and \eqref{equ:avg-queue-arrival-term-2}, we have
\begin{align}\label{equ:avg-queue-drift-2}
    & \expt \left[\sum_i  {Q}_{\mathrm{c},i}^2(t+1) - \sum_i  {Q}_{\mathrm{c},i}^2(t) |  {\boldsymbol{Q}}_{\mathrm{c}}(t)     \right] \nonumber\\
    \le & 
     2 \sum_i  {Q}_{\mathrm{c},i}(t) 
     \Biggl(
     \mathbb{1} \left\{  {Q}_{\mathrm{c},i}(t) = 0 \right\}
     \expt \left[ {\lambda}'_i(t) |  {\boldsymbol{Q}}_{\mathrm{c}}(t) \right]\nonumber\\
     & \qquad \qquad \quad + \mathbb{1} \left\{ {Q}_{\mathrm{c},i}(t) > 0 \right\}
     \expt \left[ \max\{\lambda'_i(t) - \tilde{\alpha}_{\mathrm{c},i}(t), 0\} |  {\boldsymbol{Q}}_{\mathrm{c}}(t) \right] \Biggr) \nonumber\\
    & - 2 \sum_j
     \expt \left[ {\mu}'_j(t)    {Q}_{\mathrm{c},i^*_j(t)}(t) \Bigl|\bigr.  {\boldsymbol{Q}}_{\mathrm{c}}(t)     \right]
    + \Theta (\alpha^2) \sum_i  {Q}_{\mathrm{c},i}(t)
    + \Theta(1)\nonumber\\
    = & 2 \sum_i  {Q}_{\mathrm{c},i}(t) 
     \mathbb{1} \left\{ {Q}_{\mathrm{c},i}(t) > 0 \right\}
     \expt \left[ \max\{\lambda'_i(t) - \tilde{\alpha}_{\mathrm{c},i}(t), 0\} |  {\boldsymbol{Q}}_{\mathrm{c}}(t) \right] \nonumber\\
    & - 2 \sum_j
     \expt \left[ {\mu}'_j(t)    {Q}_{\mathrm{c},i^*_j(t)}(t) \Bigl|\bigr.  {\boldsymbol{Q}}_{\mathrm{c}}(t)     \right]
    + \Theta (\alpha^2) \sum_i  {Q}_{\mathrm{c},i}(t)
    + \Theta(1).
\end{align}
Next, we add the indicator of the ``good'' event ${\cal C}$ so that we can relate $\lambda'_i(t)$ and $\mu'_j(t)$ with $x_{i,j}(k(t))$ using Lemma~\ref{lemma:improved-bisection-error}. Adding $\mathbb{1}_{\cal C}$ into \eqref{equ:avg-queue-drift-2}, we obtain
\begin{align*}
    & \expt \left[\sum_i  {Q}_{\mathrm{c},i}^2(t+1) - \sum_i  {Q}_{\mathrm{c},i}^2(t) |  {\boldsymbol{Q}}_{\mathrm{c}}(t)     \right] \nonumber\\
    \le & 
    2 \sum_i  {Q}_{\mathrm{c},i}(t) 
    \mathbb{1} \left\{ {Q}_{\mathrm{c},i}(t) > 0 \right\}
     \expt \left[ \mathbb{1}_{\cal C } \max\{\lambda'_i(t) - \tilde{\alpha}_{\mathrm{c},i}(t), 0\} |  {\boldsymbol{Q}}_{\mathrm{c}}(t) \right] \nonumber\\
    & - 2 \sum_j
     \expt \left[\mathbb{1}_{\cal C } {\mu}'_j(t) {Q}_{\mathrm{c},i^*_j(t)}(t) \Bigl|\bigr.  {\boldsymbol{Q}}_{\mathrm{c}}(t)     \right]
     + \Theta(q^{\mathrm{th}})\expt [ \mathbb{1}_{\cal C }^{\mathrm{c}} |  {\boldsymbol{Q}}_{\mathrm{c}}(t) ]
     + \Theta (\alpha^2) \sum_i  {Q}_{\mathrm{c},i}(t)
     + \Theta(1).
\end{align*}
By Lemma~\ref{lemma:improved-bisection-error}, we obtain
\begin{align}\label{equ:avg-queue-drift-3}
    & \expt \left[\sum_i  {Q}_{\mathrm{c},i}^2(t+1) - \sum_i  {Q}_{\mathrm{c},i}^2(t) \bigg|  {\boldsymbol{Q}}_{\mathrm{c}}(t)     \right] \nonumber\\
    \le & 
    2 \sum_i  {Q}_{\mathrm{c},i}(t) 
    \mathbb{1} \left\{ {Q}_{\mathrm{c},i}(t) > 0 \right\}
     \expt \biggl[ \mathbb{1}_{\cal C } \max\biggl\{ \sum_{j\in {\cal E}_{\mathrm{c},i}} x_{i,j}(k(t)) - \tilde{\alpha}_{\mathrm{c},i}(t), 0\biggr\} |  {\boldsymbol{Q}}_{\mathrm{c}}(t) \biggr] \nonumber\\
    & - 2 \sum_j
     \expt \biggl[\mathbb{1}_{\cal C } \sum_{i\in {\cal E}_{\mathrm{s},j}} x_{i,j}(k(t)) {Q}_{\mathrm{c},i^*_j(t)}(t) \Bigl|\bigr.  {\boldsymbol{Q}}_{\mathrm{c}}(t)     \biggr]\nonumber\\
    & + \Theta(q^{\mathrm{th}})\expt [ \mathbb{1}_{\cal C }^{\mathrm{c}} |  {\boldsymbol{Q}}_{\mathrm{c}}(t) ]
    + \Theta\left(\alpha^2 + \frac{\eta \epsilon}{\delta} + \eta + \delta + \epsilon\right) \sum_i Q_{\mathrm{c},i}(t)
    + \Theta\left(1\right).
\end{align}
Recall that we have shown in Appendix~\ref{sec:bound-regret-reducing-arr} that for sufficiently large $T$ such that $\alpha$ is sufficiently small,
\[
\sum_{j\in {\cal E}_{\mathrm{c},i}} x_{i,j}(k(t)) - \tilde{\alpha}_{\mathrm{c},i}(t) \ge 0
\]
for all $i$. Hence, from \eqref{equ:avg-queue-drift-3}, we have
\begin{align}\label{equ:avg-queue-drift-3-0}
    & \expt \left[\sum_i  {Q}_{\mathrm{c},i}^2(t+1) - \sum_i  {Q}_{\mathrm{c},i}^2(t) \bigg|  {\boldsymbol{Q}}_{\mathrm{c}}(t)     \right] \nonumber\\
    \le & 
    2 \sum_i  {Q}_{\mathrm{c},i}(t) 
    \Biggl(
     \expt \biggl[ \mathbb{1}_{\cal C } \sum_{j\in {\cal E}_{\mathrm{c},i}} x_{i,j}(k(t))  |  {\boldsymbol{Q}}_{\mathrm{c}}(t) \biggr] 
     - \expt \biggl[ \mathbb{1}_{\cal C } \tilde{\alpha}_{\mathrm{c},i}(t) |  {\boldsymbol{Q}}_{\mathrm{c}}(t) \biggr] \Biggr)\nonumber\\
    & - 2 \sum_j
     \expt \biggl[\mathbb{1}_{\cal C } \sum_{i\in {\cal E}_{\mathrm{s},j}} x_{i,j}(k(t)) {Q}_{\mathrm{c},i^*_j(t)}(t) \Bigl|\bigr.  {\boldsymbol{Q}}_{\mathrm{c}}(t)     \biggr]\nonumber\\
    & + \Theta(q^{\mathrm{th}})\expt [ \mathbb{1}_{\cal C }^{\mathrm{c}} |  {\boldsymbol{Q}}_{\mathrm{c}}(t) ]
    + \Theta\left(\alpha^2 + \frac{\eta \epsilon}{\delta} + \eta + \delta + \epsilon\right) \sum_i Q_{\mathrm{c},i}(t)
    + \Theta\left(1\right).
\end{align}
Recall the definition of $\tilde{\alpha}_{\mathrm{c},i}(t)$ in Appendix~\ref{sec:bound-regret-reducing-arr}. Then, from \eqref{equ:avg-queue-drift-3-0}, we have
\begin{align}\label{equ:avg-queue-drift-3-1}
    & \expt \left[\sum_i  {Q}_{\mathrm{c},i}^2(t+1) - \sum_i  {Q}_{\mathrm{c},i}^2(t) \bigg|  {\boldsymbol{Q}}_{\mathrm{c}}(t)     \right] \nonumber\\
    \le & 
    2 \sum_i  {Q}_{\mathrm{c},i}(t) 
    \Biggl(
     \expt \biggl[ \mathbb{1}_{\cal C } \sum_{j\in {\cal E}_{\mathrm{c},i}} x_{i,j}(k(t))  |  {\boldsymbol{Q}}_{\mathrm{c}}(t) \biggr]\nonumber\\ 
     & \qquad \qquad \quad - \alpha \expt \biggl[ \mathbb{1}_{\cal C } B_{\mathrm{c},i}(t) \biggl|\diff*{F_i^{-1}(p)}{p}{p=p^{+/-}_{\mathrm{c},i}(k(t),m(t))}\biggr| \biggl|\biggr.  {\boldsymbol{Q}}_{\mathrm{c}}(t) \biggr] \Biggr)\nonumber\\
    & - 2 \sum_j
     \expt \biggl[\mathbb{1}_{\cal C } \sum_{i\in {\cal E}_{\mathrm{s},j}} x_{i,j}(k(t)) {Q}_{\mathrm{c},i^*_j(t)}(t) \biggl|\biggr.  {\boldsymbol{Q}}_{\mathrm{c}}(t)     \biggr]\nonumber\\
    & + \Theta(q^{\mathrm{th}})\expt [ \mathbb{1}_{\cal C }^{\mathrm{c}} |  {\boldsymbol{Q}}_{\mathrm{c}}(t) ]
    + \Theta\left(\alpha^2 + \frac{\eta \epsilon}{\delta} + \eta + \delta + \epsilon\right) \sum_i Q_{\mathrm{c},i}(t)
    + \Theta\left(1\right).
\end{align}
By Assumption~\ref{assum:7}, we know that $\left|\diff*{F_i^{-1}(p)}{p}{p=p^{+/-}_{\mathrm{c},i}(k(t),m(t))}\right|$ is both upper and lower bounded. Hence, we have
\begin{align}\label{equ:avg-queue-drift-3-1-1}
    & \expt \left[ \mathbb{1}_{\cal C } B_{\mathrm{c},i}(t) \biggl| \diff*{F_i^{-1}(p)}{p}{p=p^{+/-}_{\mathrm{c},i}(k(t),m(t))} \biggr| 
     \bigg|  {\boldsymbol{Q}}_{\mathrm{c}}(t) \right] \nonumber\\
     \ge & 
     \expt \left[ B_{\mathrm{c},i}(t)  \biggl| \diff*{F_i^{-1}(p)}{p}{p=p^{+/-}_{\mathrm{c},i}(k(t),m(t))} \biggr| 
     \bigg|  {\boldsymbol{Q}}_{\mathrm{c}}(t) \right]
      - \Theta(1) \expt [ \mathbb{1}_{\cal C }^{\mathrm{c}} |  {\boldsymbol{Q}}_{\mathrm{c}}(t) ]\nonumber\\
      \ge & C_{\mathrm{L}} \expt \left[ B_{\mathrm{c},i}(t) \bigg|  {\boldsymbol{Q}}_{\mathrm{c}}(t) \right]
      - \Theta(1) \expt [ \mathbb{1}_{\cal C }^{\mathrm{c}} |  {\boldsymbol{Q}}_{\mathrm{c}}(t) ].
\end{align}
From \eqref{equ:avg-queue-drift-3-1} and \eqref{equ:avg-queue-drift-3-1-1}, we have
\begin{align}\label{equ:avg-queue-drift-3-2}
    & \expt \left[\sum_i  {Q}_{\mathrm{c},i}^2(t+1) - \sum_i  {Q}_{\mathrm{c},i}^2(t) \bigg|  {\boldsymbol{Q}}_{\mathrm{c}}(t)     \right] \nonumber\\
    \le & 
    2 \sum_i  {Q}_{\mathrm{c},i}(t) 
    \Biggl(
     \expt \biggl[ \mathbb{1}_{\cal C } \sum_{j\in {\cal E}_{\mathrm{c},i}} x_{i,j}(k(t))  |  {\boldsymbol{Q}}_{\mathrm{c}}(t) \biggr] 
     - \alpha C_{\mathrm{L}} \expt \left[ B_{\mathrm{c},i}(t) \bigg|  {\boldsymbol{Q}}_{\mathrm{c}}(t) \right] \Biggr)\nonumber\\
    & - 2 \sum_j
     \expt \biggl[\mathbb{1}_{\cal C } \sum_{i\in {\cal E}_{\mathrm{s},j}} x_{i,j}(k(t)) {Q}_{\mathrm{c},i^*_j(t)}(t) \biggl|\biggr.  {\boldsymbol{Q}}_{\mathrm{c}}(t)     \biggr]\nonumber\\
    & + \Theta(q^{\mathrm{th}})\expt [ \mathbb{1}_{\cal C }^{\mathrm{c}} |  {\boldsymbol{Q}}_{\mathrm{c}}(t) ]
    + \Theta\left(\alpha^2 + \frac{\eta \epsilon}{\delta} + \eta + \delta + \epsilon\right) \sum_i Q_{\mathrm{c},i}(t)
    + \Theta\left(1\right).
\end{align}
Rearranging terms, we obtain
\begin{align}
    & \expt \left[\sum_i  {Q}_{\mathrm{c},i}^2(t+1) - \sum_i  {Q}_{\mathrm{c},i}^2(t) |  {\boldsymbol{Q}}_{\mathrm{c}}(t)     \right] \nonumber\\
    \le & 2 \expt \left[\mathbb{1}_{\cal C }
    \sum_{(i,j)\in {\cal E}} x_{i,j}(k(t))
    \left(
    {Q}_{\mathrm{c},i}(t) - {Q}_{\mathrm{c},i^*_j(t)}(t)
    \right)
    \bigg| {\boldsymbol{Q}}_{\mathrm{c}}(t) \right] \label{equ:avg-queue-drift-4}\\
    &
    - 2 \alpha C_{\mathrm{L}} \sum_i  {Q}_{\mathrm{c},i}(t) 
     \expt \left[  B_{\mathrm{c},i}(t)
     \bigg|  {\boldsymbol{Q}}_{\mathrm{c}}(t) \right]
    \label{equ:avg-queue-drift-5}\\
    & + \Theta(q^{\mathrm{th}})\expt [ \mathbb{1}_{\cal C }^{\mathrm{c}} |  {\boldsymbol{Q}}_{\mathrm{c}}(t) ]
    + \Theta\left(\alpha^2 + \frac{\eta \epsilon}{\delta} + \eta + \delta + \epsilon\right) \sum_i Q_{\mathrm{c},i}(t)
    + \Theta\left(1\right).\label{equ:avg-queue-drift-6}
\end{align}
Note that under the matching algorithm, during a single time slot, there are at most $J$ departures from any customer-side queue \cite{yang2024learning}. Hence, for the term \eqref{equ:avg-queue-drift-4}, we have
\begin{align}\label{equ:avg-queue-drift-4-bound}
    {Q}_{\mathrm{c},i}(t) - {Q}_{\mathrm{c},i^*_j(t)}(t) \le J.
\end{align}
For the term \eqref{equ:avg-queue-drift-5}, recall that $B_{\mathrm{c},i}(t)$ is a Bernoulli random variable with probability $1/2$, which is independent of ${\boldsymbol{Q}}_{\mathrm{c}}(t)$. Hence, we have
\begin{align}\label{equ:avg-queue-drift-5-bound}
    \expt \left[  B_{\mathrm{c},i}(t)
     \bigg|  {\boldsymbol{Q}}_{\mathrm{c}}(t) \right]
     = \expt \left[  B_{\mathrm{c},i}(t) \right]
     = \frac{1}{2}.
\end{align}
Hence, from \eqref{equ:avg-queue-drift-4}-\eqref{equ:avg-queue-drift-5-bound}, we have
\begin{align*}
    & \expt \left[\sum_i  {Q}_{\mathrm{c},i}^2(t+1) - \sum_i  {Q}_{\mathrm{c},i}^2(t) |  {\boldsymbol{Q}}_{\mathrm{c}}(t)     \right] \nonumber\\
    \le
    & - \left( C_{\mathrm{L}} \alpha - \Theta\left(\alpha^2 + \frac{\eta \epsilon}{\delta} + \eta + \delta + \epsilon\right) \right) \sum_i  {Q}_{\mathrm{c},i}(t)
    + \Theta(q^{\mathrm{th}})\expt [ \mathbb{1}_{\cal C }^{\mathrm{c}} |  {\boldsymbol{Q}}_{\mathrm{c}}(t) ]
    + \Theta\left(1\right).
\end{align*}
Since $\alpha$ is orderwise greater than $\left(\frac{\eta \epsilon}{\delta} + \eta + \delta + \epsilon\right)$ and $\alpha^2$, for sufficiently large $T$, we have $ C_{\mathrm{L}} \alpha - \Theta\left(\alpha^2 + \frac{\eta \epsilon}{\delta} + \eta + \delta + \epsilon\right) \ge \frac{1}{2}C_{\mathrm{L}} \alpha$. Hence, we have
\begin{align*}
    & \expt \left[\sum_i  {Q}_{\mathrm{c},i}^2(t+1) - \sum_i  {Q}_{\mathrm{c},i}^2(t) |  {\boldsymbol{Q}}_{\mathrm{c}}(t)     \right] \nonumber\\
    \le
    & - \frac{1}{2} C_{\mathrm{L}} \alpha \sum_i  {Q}_{\mathrm{c},i}(t) 
    + \Theta(q^{\mathrm{th}})\expt [ \mathbb{1}_{\cal C }^{\mathrm{c}} |  {\boldsymbol{Q}}_{\mathrm{c}}(t) ]
    + \Theta\left( 1 \right).
\end{align*}
Taking expectation on both sides and using Lemma~\ref{lemma:concentration}, we have
\begin{align*}
    & \expt \left[\sum_i  {Q}_{\mathrm{c},i}^2(t+1) - \sum_i  {Q}_{\mathrm{c},i}^2(t)   \right] \nonumber\\
    \le
    & - \frac{1}{2} C_{\mathrm{L}} \alpha \expt \left[ \sum_i  {Q}_{\mathrm{c},i}(t) 
     \right]
    + \Theta\left(1 + q^{\mathrm{th}} \left(T\epsilon^{\frac{\beta}{2}+1} + \epsilon^{\frac{\beta}{2}-1} \log (1/\epsilon) \right)
     \right).
\end{align*}
Taking summation from $t=1$ to $t=T$ and rearranging terms, we obtain
\begin{align}\label{equ:avg-queue-drift-7}
    \sum_{t=1}^{T} \expt \left[ \sum_i  {Q}_{\mathrm{c},i}(t) 
     \right]
    \le   
    T \Theta\left( \frac{1}{\alpha} + \frac{q^{\mathrm{th}}}{\alpha} \left(T\epsilon^{\frac{\beta}{2}+1} + \epsilon^{\frac{\beta}{2}-1} \log (1/\epsilon) 
    \right)
    \right).
\end{align}
Dividing both sides by $T$, we have
\begin{align*}
    \frac{1}{T}\sum_{t=1}^{T} \expt \left[
    \sum_i  {Q}_{\mathrm{c},i}(t)
    \right] 
    \le \Theta\left( \frac{1}{\alpha} + \frac{q^{\mathrm{th}}}{\alpha} \left(T\epsilon^{\frac{\beta}{2}+1} + \epsilon^{\frac{\beta}{2}-1} \log (1/\epsilon) 
    \right)
    \right).
\end{align*}
The average queue length bound~\eqref{equ:avg-queue-length-bound-3} in Theorem~\ref{theo:3} is proved.

\clearpage
\newpage

\section{Proof of Lemma~\ref{lemma:theo-3-bound-time-large-queue}}
\label{app:proof-lemma-theo-3-bound-time-large-queue}

Similar to the proof of Lemma~6 in \cite{yang2024learning}, we use Lyapunov drift method to prove Lemma~\ref{lemma:theo-3-bound-time-large-queue}. The main difference is that under the \textit{probabilistic two-price policy}, the arrival rates are randomly reduced when the queue length is greater than zero and less than the threshold $q^{\mathrm{th}}$.

Recall the Lyapunov function $V_{\mathrm{c}}(t)\coloneqq \sum_i Q_{\mathrm{c},i}^2(t)$. Following the same argument as the proof of Lemma 6 in \cite{yang2024learning}, we obtain
\begin{align}\label{equ:theo-3-drift-1}
    & \expt [V_{\mathrm{c}}(t+1) - V_{\mathrm{c}}(t) | \boldsymbol{Q}_{\mathrm{c}}(t) ] \nonumber\\
    \le & 
     \sum_i |{\cal E}_{\mathrm{c},i}|^2 + 2 \sum_i Q_{\mathrm{c},i}(t) \biggl( \expt [ A_{\mathrm{c},i}(t) | \boldsymbol{Q}_{\mathrm{c}}(t) ] -  \expt \biggl[\sum_{j\in {\cal E}_{\mathrm{c},i}} X_{i,j}(t) \biggl|\biggr. \boldsymbol{Q}_{\mathrm{c}}(t) \biggr] \biggr).
\end{align}
Under the \textit{probabilistic two-price policy}, if the length of a customer queue $i$ is less than $q^{\mathrm{th}}$, the arrival rate of the queue will be at most $\lambda'_i(t)$, where $\lambda'_i(t)\coloneqq F^{-1}_i(p^{+/-}_{\mathrm{c},i}(k(t),m(t)))$. If the length of a queue is greater than or equal to $q^{\mathrm{th}}$, then the arrival rate of the queue will be zero. Hence,
for the term $\expt [ A_{\mathrm{c},i}(t) | \boldsymbol{Q}_{\mathrm{c}}(t) ]$ in \eqref{equ:theo-3-drift-1}, we have
\begin{align}\label{equ:theo-3-drift-1-1}
    \expt [ A_{\mathrm{c},i}(t) | \boldsymbol{Q}_{\mathrm{c}}(t) ] 
    = & \mathbb{1} \left\{ Q_{\mathrm{c},i}(t) < q^{\mathrm{th}} \right\} \expt [ A_{\mathrm{c},i}(t) | \boldsymbol{Q}_{\mathrm{c}}(t) ]\nonumber\\
    = & \mathbb{1} \left\{ Q_{\mathrm{c},i}(t) < q^{\mathrm{th}} \right\}
    \expt \bigl[  
        \expt [ A_{\mathrm{c},i}(t) 
        | \lambda'_i(t), \boldsymbol{Q}_{\mathrm{c}}(t) ]
    \bigl|\bigr. \boldsymbol{Q}_{\mathrm{c}}(t) \bigr]\nonumber\\
    \le & \mathbb{1} \left\{ Q_{\mathrm{c},i}(t) < q^{\mathrm{th}} \right\}
    \expt \bigl[  \lambda'_i(t)  \bigl|\bigr. \boldsymbol{Q}_{\mathrm{c}}(t) \bigr].
\end{align}
For the term $\sum_i Q_{\mathrm{c},i}(t) \expt [\sum_{j\in {\cal E}_{\mathrm{c},i}} X_{i,j}(t) | \boldsymbol{Q}_{\mathrm{c}}(t) ]$ in \eqref{equ:theo-3-drift-1}, following the same argument as the proof of Lemma 6 in \cite{yang2024learning}, we obtain
\begin{align}\label{equ:theo-3-drift-1-2}
    \sum_i Q_{\mathrm{c},i}(t) \expt \biggl[\sum_{j\in {\cal E}_{\mathrm{c},i}} X_{i,j}(t) \biggl|\biggr. \boldsymbol{Q}_{\mathrm{c}}(t) \biggr]
    = &  \sum_j
    \mathbb{1} \biggl\{ \sum_{i\in {\cal E}_{\mathrm{s},j}} Q_{\mathrm{c},i}(t)>0 \biggr\}
    \expt \bigl[ \mathbb{1} \left\{ A_{\mathrm{s},j}(t) = 1 \right\}  Q_{\mathrm{c},i^*_j(t)}(t) 
    \bigl|\bigr. \boldsymbol{Q}_{\mathrm{c}}(t) \bigr],
\end{align}
where
\begin{align}\label{equ:theo-3-drift-1-3}
    \expt \bigl[ \mathbb{1} \left\{ A_{\mathrm{s},j}(t) = 1 \right\}  Q_{\mathrm{c},i^*_j(t)}(t) 
    \bigl|\bigr. \boldsymbol{Q}_{\mathrm{c}}(t) \bigr]
    = & \expt \left[   \expt \bigl[ A_{\mathrm{s},j}(t)   
     \bigl|\bigr. \boldsymbol{Q}_{\mathrm{c}}(t), i^*_j(t), \mu'_j(t)  \bigr] Q_{\mathrm{c},i^*_j(t)}(t) \Bigl|\bigr. \boldsymbol{Q}_{\mathrm{c}}(t) \right].
\end{align}
Here we cite a lemma from \cite{yang2024learning} as follows:
\begin{lemma}[\cite{yang2024learning}]\label{lemma:empty-queues}
    Assuming that all queues are initially empty, under the matching algorithm we use and any pricing algorithm, at any time slot $t$, 
    if a queue $i$ (or $j$) is non-empty before arrivals, then all queues on the opposite side that are connected to this queue must be empty, i.e., $\sum_{j: (i,j)\in{\cal E}} Q_{\mathrm{s}, j}(t) = 0$ (or $\sum_{i: (i,j)\in{\cal E}} Q_{\mathrm{c}, i}(t) = 0$).
\end{lemma}
Note that we have $\sum_{i\in {\cal E}_{\mathrm{s},j}} Q_{\mathrm{c},i}(t)>0$ and thus $Q_{\mathrm{s},j}(t)=0$ by Lemma~\ref{lemma:empty-queues}. Note that under the \textit{probabilistic two-price policy}, if the queue length is zero, we do not reduce the arrival rate. Hence, the arrival rate of the server queue $j$ is $\mu'_j(t)$, where $\mu'_j(t)\coloneqq G^{-1}_j(p^{+/-}_{\mathrm{s},j}(k(t),m(t)))$. Hence, we have
\begin{align}\label{equ:theo-3-drift-1-4}
    \expt \bigl[ A_{\mathrm{s},j}(t)   
     \bigl|\bigr. \boldsymbol{Q}_{\mathrm{c}}(t), i^*_j(t), \mu'_j(t)  \bigr] = \mu'_j(t).
\end{align}
Combining \eqref{equ:theo-3-drift-1-2}, \eqref{equ:theo-3-drift-1-3}, and \eqref{equ:theo-3-drift-1-4}, we have
\begin{align}\label{equ:theo-3-drift-1-5}
    \sum_i Q_{\mathrm{c},i}(t) \expt \biggl[\sum_{j\in {\cal E}_{\mathrm{c},i}} X_{i,j}(t) \biggl|\biggr. \boldsymbol{Q}_{\mathrm{c}}(t) \biggr]
    = & \sum_j
    \mathbb{1} \biggl\{ \sum_{i\in {\cal E}_{\mathrm{s},j}} Q_{\mathrm{c},i}(t)>0 \biggr\} 
    \expt \left[ \mu'_j(t) Q_{\mathrm{c},i^*_j(t)}(t) \Bigl|\bigr. \boldsymbol{Q}_{\mathrm{c}}(t) \right]\nonumber\\
    = & \sum_j
    \expt \left[ \mu'_j(t) Q_{\mathrm{c},i^*_j(t)}(t) \Bigl|\bigr. \boldsymbol{Q}_{\mathrm{c}}(t) \right].
\end{align}
where the last equality holds since $Q_{\mathrm{c},i^*_j(t)}(t)=0$ if $\sum_{i\in {\cal E}_{\mathrm{s},j}} Q_{\mathrm{c},i}(t)=0$. From \eqref{equ:theo-3-drift-1}, \eqref{equ:theo-3-drift-1-1}, and \eqref{equ:theo-3-drift-1-5}, we have
\begin{align}\label{equ:theo-3-drift-2}
    & \expt [V_{\mathrm{c}}(t+1) - V_{\mathrm{c}}(t) | \boldsymbol{Q}_{\mathrm{c}}(t) ] \nonumber\\
    \le & 
     \sum_i |{\cal E}_{\mathrm{c},i}|^2 + 2 \sum_i Q_{\mathrm{c},i}(t) \mathbb{1} \left\{ Q_{\mathrm{c},i}(t) < q^{\mathrm{th}} \right\}
    \expt \bigl[  \lambda'_i(t)  \bigl|\bigr. \boldsymbol{Q}_{\mathrm{c}}(t) \bigr]  - 2 \sum_j
    \expt \left[   \mu'_j(t) Q_{\mathrm{c},i^*_j(t)}(t) \Bigl|\bigr. \boldsymbol{Q}_{\mathrm{c}}(t) \right].
\end{align}
Similarly to the proof of Lemma~6 in \cite{yang2024learning}, we next add the indicator of the ``good'' event ${\cal C}$ into \eqref{equ:theo-3-drift-2}, obtaining
\begin{align}\label{equ:theo-3-drift-3}
    & \expt [V_{\mathrm{c}}(t+1) - V_{\mathrm{c}}(t) | \boldsymbol{Q}_{\mathrm{c}}(t) ] \nonumber\\
    \le & \sum_i |{\cal E}_{\mathrm{c},i}|^2 + 2 \sum_i Q_{\mathrm{c},i}(t) \mathbb{1} \left\{ Q_{\mathrm{c},i}(t) < q^{\mathrm{th}} \right\}
     \expt \bigl[  \lambda'_i(t) \mathbb{1}_{{\cal C}} \bigl|\bigr. \boldsymbol{Q}_{\mathrm{c}}(t) \bigr]\nonumber\\
     & + 2 I q^{\mathrm{th}} \expt \bigl[ \mathbb{1}_{{\cal C}^{\mathrm{c}}} \bigl|\bigr. \boldsymbol{Q}_{\mathrm{c}}(t) \bigr]  - 2 \sum_j
    \expt \left[   \mu'_j(t) Q_{\mathrm{c},i^*_j(t)}(t) \mathbb{1}_{{\cal C}}
    \Bigl|\bigr. \boldsymbol{Q}_{\mathrm{c}}(t)  \right]
\end{align}

Next, we will relate $\lambda'_i(t)$ and $\mu'_j(t)$ to the arrival rates $\sum_{j\in {\cal E}_{\mathrm{c},i}} x_{i,j}(k(t))$ and $\sum_{i\in {\cal E}_{\mathrm{s},j}} x_{i,j}(k(t))$, where $k(t)$ denotes the outer iteration at time $t$. We will use the following lemma by \cite{yang2024learning}:
\begin{lemma}[\cite{yang2024learning}]
\label{lemma:improved-bisection-error}
    Let Assumption~\ref{assum:1}, Assumption~\ref{assum:3}, and Assumption~\ref{assum:5} hold. Suppose $T$ is sufficiently large. Then under the proposed algorithm and the event ${\cal C}$, for all $t\ge 1$, we have
    \begin{align*}
    \biggl|\lambda'_i(t) - \sum_{j\in {\cal E}_{\mathrm{c},i}} x_{i,j}(k(t))\biggr|
    \le & 2 L_{F^{-1}_i} e_{\mathrm{c},i} + 2\epsilon \left[1 + L_{F^{-1}_i} \left( p_{\mathrm{c},i,\max} - p_{\mathrm{c},i,\min}\right)\right]  + \delta \sqrt{|{\cal E}_{\mathrm{c},i}|}, \mbox{ for all } i;\nonumber\\
    \biggl|\mu'_j(t) - \sum_{i\in {\cal E}_{\mathrm{s},j}} x_{i,j}(k(t))\biggr|
    \le & 2 L_{G^{-1}_j} e_{\mathrm{s},j} + 2\epsilon \left[1 + L_{G^{-1}_j} \left( p_{\mathrm{s},j,\max} - p_{\mathrm{s},j,\min}\right)\right]  + \delta \sqrt{|{\cal E}_{\mathrm{s},j}|}, \mbox{ for all } j.
    \end{align*}
\end{lemma}
Lemma~\ref{lemma:improved-bisection-error} means that the arrival rates corresponding to the bisection prices are close to the target arrival rates corresponding to the current outer iteration. From \eqref{equ:theo-3-drift-3}, using Lemma~\ref{lemma:improved-bisection-error} and following the same proof as that in the proof of Lemma~6 in \cite{yang2024learning}, we obtain
\begin{align}\label{equ:theo-3-drift-final-1}
    \sum_{t=1}^{T} \sum_i \expt \Bigl[ 
    \mathbb{1}_{{\cal C}}
    \mathbb{1} \Bigl\{ Q_{\mathrm{c},i}(t) \ge q^{\mathrm{th}} \Bigr\}
    \Bigr]
    \le \Theta \left(
    \frac{T}{q^{\mathrm{th}}}
    + T^2\epsilon^{\frac{\beta}{2}+1} + T\epsilon^{\frac{\beta}{2}-1} \log (1/\epsilon)
    + T \left(\frac{\eta \epsilon}{\delta} + \eta + \delta + \epsilon\right)
    \right).
\end{align}
and for the server side,
\begin{align}\label{equ:theo-3-drift-final-2}
    \sum_{t=1}^{T} \sum_j \expt \Bigl[ 
    \mathbb{1}_{{\cal C}}
    \mathbb{1} \Bigl\{ Q_{\mathrm{s},j}(t) \ge q^{\mathrm{th}} \Bigr\}
    \Bigr]
    \le \Theta \left(
    \frac{T}{q^{\mathrm{th}}}
    + T^2\epsilon^{\frac{\beta}{2}+1} + T\epsilon^{\frac{\beta}{2}-1} \log (1/\epsilon)
    + T \left(\frac{\eta \epsilon}{\delta} + \eta + \delta + \epsilon\right)
    \right).
\end{align}
From \eqref{equ:theo-3-drift-final-1} and \eqref{equ:theo-3-drift-final-2}, Lemma~\ref{lemma:theo-3-bound-time-large-queue} is proved.

\clearpage
\newpage

\section{Proof of Lemma~\ref{lemma:h-properties}}
\label{app:proof-lemma-h-properties}

We first prove the Lipschitz property \eqref{equ:lipschitz-h}.

For any arrival rate vectors $\boldsymbol{\lambda}^{(1)}$, $\boldsymbol{\mu}^{(1)}$, $\boldsymbol{\lambda}^{(2)}$, $\boldsymbol{\mu}^{(2)}$, we have
\begin{align*}
    & \left|h(\boldsymbol{\lambda}^{(1)}, \boldsymbol{\mu}^{(1)}) - h(\boldsymbol{\lambda}^{(2)}, \boldsymbol{\mu}^{(2)})\right|\nonumber\\
    = & \left|\left(\sum_i \lambda_i^{(1)} F_i(\lambda_i^{(1)}) - \sum_j \mu_j^{(1)} G_j(\mu_j^{(1)}) \right) - \left(\sum_i \lambda_i^{(2)} F_i(\lambda_i^{(2)}) - \sum_j \mu_j^{(2)} G_j(\mu_j^{(2)}) \right)\right|\nonumber\\
    = & \left|\sum_i  \lambda_i^{(1)} F_i(\lambda_i^{(1)}) - \lambda_i^{(2)} F_i(\lambda_i^{(2)}) 
    + \sum_j  \mu_j^{(2)} G_j(\mu_j^{(2)}) -  \mu_j^{(1)} G_j(\mu_j^{(1)})\right|\nonumber\\
    \le & \sum_i \left| \lambda_i^{(1)} F_i(\lambda_i^{(1)}) - \lambda_i^{(1)} F_i(\lambda_i^{(2)}) + \lambda_i^{(1)} F_i(\lambda_i^{(2)}) - \lambda_i^{(2)} F_i(\lambda_i^{(2)}) \right|\nonumber\\
    & + \sum_j \left| \mu_j^{(2)} G_j(\mu_j^{(2)}) - \mu_j^{(2)} G_j(\mu_j^{(1)}) + \mu_j^{(2)} G_j(\mu_j^{(1)}) -  \mu_j^{(1)} G_j(\mu_j^{(1)})\right|\nonumber\\
    \le & \sum_i\left(  L_{F_i} |\lambda_i^{(1)} - \lambda_i^{(2)} | 
    + p_{\mathrm{c},i,\max} |\lambda_i^{(1)} - \lambda_i^{(2)} | \right)
    + \sum_j \left( L_{G_j} |\mu_j^{(1)} - \mu_j^{(2)} | + p_{\mathrm{s},j,\max} |\mu_j^{(1)} - \mu_j^{(2)} | \right)\nonumber\\
    \le & L_h \left\| 
    \begin{bmatrix}
    \boldsymbol{\lambda}^{(1)} - \boldsymbol{\lambda}^{(2)} \\
    \boldsymbol{\mu}^{(1)} - \boldsymbol{\mu}^{(2)}
    \end{bmatrix}
    \right\|_1,
\end{align*}
where in the first inequality we add and subtract the same items and use the triangle inequality, in the second inequality we use the Lipschitz properties in Assumption~\ref{assum:1}, and in the last inequality $L_h\coloneqq \max\{ \max_i (L_{F_i} + p_{\mathrm{c},i,\max}), \max_j (L_{G_j} +p_{\mathrm{s},j,\max} ) \}$.

Next, we prove the smoothness property \eqref{equ:smooth-h}.

For any arrival rate vectors $\boldsymbol{\lambda}^{(1)}$, $\boldsymbol{\mu}^{(1)}$, $\boldsymbol{\lambda}^{(2)}$, $\boldsymbol{\mu}^{(2)}$, we have
\begin{align*}
    & \left\| \nabla h (\boldsymbol{\lambda}^{(1)}, \boldsymbol{\mu}^{(1)}) - \nabla h (\boldsymbol{\lambda}^{(2)}, \boldsymbol{\mu}^{(2)}) \right\|_2^2 \nonumber\\
    = & \sum_i \left[ F_i(\lambda_i^{(1)}) + \lambda_i^{(1)} \diff*{F_i(\lambda)}{\lambda}{\lambda=\lambda_i^{(1)}} - F_i(\lambda_i^{(2)}) - \lambda_i^{(2)} \diff*{F_i(\lambda)}{\lambda}{\lambda=\lambda_i^{(2)}} \right]^2\nonumber\\
    & + \sum_j \left[ G_j(\mu_j^{(1)}) + \mu_j^{(1)} \diff*{G_j(\mu)}{\mu}{\mu=\mu_j^{(1)}} - G_j(\mu_j^{(2)}) - \mu_j^{(2)} \diff*{G_j(\mu)}{\mu}{\mu=\mu_j^{(2)}} \right]^2.
\end{align*}
Then by the lipschitz properties in Assumption~\ref{assum:1}, we have
\begin{align*}
    & \left\| \nabla h (\boldsymbol{\lambda}^{(1)}, \boldsymbol{\mu}^{(1)}) - \nabla h (\boldsymbol{\lambda}^{(2)}, \boldsymbol{\mu}^{(2)}) \right\|_2^2 \nonumber\\
    \le & \sum_i \left[ L_{F_i}|\lambda_i^{(1)} -\lambda_i^{(2)} | +  \lambda_i^{(1)} \diff*{F_i(\lambda)}{\lambda}{\lambda=\lambda_i^{(1)}} - \lambda_i^{(2)} \diff*{F_i(\lambda)}{\lambda}{\lambda=\lambda_i^{(2)}} \right]^2\nonumber\\
    & + \sum_j \left[ L_{G_j}|\mu_j^{(1)} - \mu_j^{(2)}| + \mu_j^{(1)} \diff*{G_j(\mu)}{\mu}{\mu=\mu_j^{(1)}} - \mu_j^{(2)} \diff*{G_j(\mu)}{\mu}{\mu=\mu_j^{(2)}} \right]^2.
\end{align*}
Adding and subtracting the same terms, we have
\begin{align*}
    & \left\| \nabla h (\boldsymbol{\lambda}^{(1)}, \boldsymbol{\mu}^{(1)}) - \nabla h (\boldsymbol{\lambda}^{(2)}, \boldsymbol{\mu}^{(2)}) \right\|_2^2 \nonumber\\
    = & \sum_i \Biggl[ L_{F_i}|\lambda_i^{(1)} -\lambda_i^{(2)} |
    +  \lambda_i^{(1)} \left(\diff*{F_i(\lambda)}{\lambda}{\lambda=\lambda_i^{(1)}} - \diff*{F_i(\lambda)}{\lambda}{\lambda=\lambda_i^{(2)}} \right) \nonumber\\
    &\qquad + \left(\lambda_i^{(2)} - \lambda_i^{(1)}\right) \left( - \diff*{F_i(\lambda)}{\lambda}{\lambda=\lambda_i^{(2)}}\right)
    \Biggr]^2\nonumber\\
    & + \sum_j \Biggl[ L_{G_j}|\mu_j^{(1)} - \mu_j^{(2)}|
    + \mu_j^{(1)} \left( \diff*{G_j(\mu)}{\mu}{\mu=\mu_j^{(1)}} 
    -  \diff*{G_j(\mu)}{\mu}{\mu=\mu_j^{(2)}} \right) \nonumber\\
    &\qquad  + \left(\mu_j^{(1)} - \mu_j^{(2)}\right) \diff*{G_j(\mu)}{\mu}{\mu=\mu_j^{(2)}}
    \Biggr]^2.
\end{align*}
Note that by Assumption~\ref{assum:7}(1), we know that $F_i$ and $G_j$ are smooth for all $i,j$. Hence, we have
\begin{align*}
    & \left\| \nabla h (\boldsymbol{\lambda}^{(1)}, \boldsymbol{\mu}^{(1)}) - \nabla h (\boldsymbol{\lambda}^{(2)}, \boldsymbol{\mu}^{(2)}) \right\|_2^2 \nonumber\\
    \le & \sum_i \Biggl[ L_{F_i}|\lambda_i^{(1)} -\lambda_i^{(2)} |
    +  \lambda_i^{(1)} \beta_{F_i} |\lambda_i^{(1)} -\lambda_i^{(2)} |  
    + \left(\lambda_i^{(2)} - \lambda_i^{(1)}\right) \left( - \diff*{F_i(\lambda)}{\lambda}{\lambda=\lambda_i^{(2)}}\right)
    \Biggr]^2\nonumber\\
    & + \sum_j \Biggl[ L_{G_j}|\mu_j^{(1)} - \mu_j^{(2)}|
    + \mu_j^{(1)} \beta_{G_j} |\mu_j^{(1)} - \mu_j^{(2)}|  
    + \left(\mu_j^{(1)} - \mu_j^{(2)}\right) \diff*{G_j(\mu)}{\mu}{\mu=\mu_j^{(2)}}
    \Biggr]^2.
\end{align*}
Note that $- \diff*{F_i(\lambda)}{\lambda}{\lambda=\lambda_i^{(2)}}$ and $\diff*{G_j(\mu)}{\mu}{\mu=\mu_j^{(2)}}$ are positive since $F_i$ is strictly decreasing and $G_j$ is strictly increasing by Assumption~\ref{assum:1}. They are also bounded by the smoothness properties in Assumption~\ref{assum:7}(1). Hence, we have
\begin{align*}
    & \left\| \nabla h (\boldsymbol{\lambda}^{(1)}, \boldsymbol{\mu}^{(1)}) - \nabla h (\boldsymbol{\lambda}^{(2)}, \boldsymbol{\mu}^{(2)}) \right\|_2^2 \nonumber\\
    \le & \sum_i \left(L_{F_i} + \beta_{F_i} + \max_{\lambda} \left( - \diff{F_i(\lambda)}{\lambda}\right)  \right)^2  \left|\lambda_i^{(1)} -\lambda_i^{(2)} \right|^2\nonumber\\ 
     & + \sum_j \left(L_{G_j} + \beta_{G_j} + \max_{\mu} \diff{G_j(\mu)}{\mu} \right)^2 
    \left|\mu_j^{(1)} - \mu_j^{(2)}\right|^2\nonumber\\
    \le & \beta_h^2 \left\|
            \begin{bmatrix}
            \boldsymbol{\lambda}^{(1)} - \boldsymbol{\lambda}^{(2)}\\
            \boldsymbol{\mu}^{(1)} - \boldsymbol{\mu}^{(2)}
            \end{bmatrix}
            \right\|_2^2,
\end{align*}
where 
\[
\beta_h^2\coloneqq \max \left\{\max_i \left(L_{F_i} + \beta_{F_i} + \max_{\lambda} \left( - \diff{F_i(\lambda)}{\lambda}\right)  \right)^2  ,\max_j \left(L_{G_j} + \beta_{G_j} + \max_{\mu} \diff{G_j(\mu)}{\mu} \right)^2 \right \}.
\]
Therefore, we have
\[
\left\| \nabla h (\boldsymbol{\lambda}^{(1)}, \boldsymbol{\mu}^{(1)}) - \nabla h (\boldsymbol{\lambda}^{(2)}, \boldsymbol{\mu}^{(2)}) \right\|_2 
\le \beta_h 
\left\|
\begin{bmatrix}
\boldsymbol{\lambda}^{(1)} - \boldsymbol{\lambda}^{(2)}\\
\boldsymbol{\mu}^{(1)} - \boldsymbol{\mu}^{(2)}
\end{bmatrix}
\right\|_2.
\]
Lemma~\ref{lemma:h-properties} is proved.

\clearpage
\newpage

\section{Proof of Lemma~\ref{lemma:avg-rate-balanced}}
\label{app:proof-lemma-avg-rate-balanced}

Let $L(t)\coloneqq \sum_i (-\kappa_{\mathrm{c},i}) Q_{\mathrm{c},i}(t) + \sum_j (-\kappa_{\mathrm{s},j}) Q_{\mathrm{s},j}(t)$. Consider the drift $\expt [L(t+1) - L(t)]$ as follows:
\begin{align}
    & \expt [L(t+1) - L(t)] \nonumber\\
    = & \expt \left[ \sum_i (- \kappa_{\mathrm{c},i}) \left( Q_{\mathrm{c},i}(t+1) - Q_{\mathrm{c},i}(t) \right) \right] \label{equ:drift-customer-side}\\
    & + \expt \left[ \sum_j ( - \kappa_{\mathrm{s},j}) \left( Q_{\mathrm{s},j}(t+1) - Q_{\mathrm{s},j}(t) \right) \right]. \label{equ:drift-server-side}
\end{align}
We first bound the customer-side drift term \eqref{equ:drift-customer-side}.
Note that the dynamics of the queue length for type $i$ customers are given by
\begin{align}\label{equ:dynamics}
    Q_{\mathrm{c}, i}(t+1) = Q_{\mathrm{c}, i}(t) + A_{\mathrm{c}, i}(t) - \sum_{j: (i,j)\in {\cal E}} X_{i,j}(t).
\end{align}
Hence, we have
\begin{align}\label{equ:drift-customer-side-1}
    \eqref{equ:drift-customer-side} = \sum_i \kappa_{\mathrm{c},i} \left( 
    \sum_j \expt [X_{i,j}(t)] -  \expt [ A_{\mathrm{c},i}(t) ] \right),
\end{align}
where
\begin{align}\label{equ:drift-customer-side-2}
    \expt [ A_{\mathrm{c},i}(t) ] 
    = & \expt \biggl[\mathbb{1}\{ 0 < Q_{\mathrm{c},i}(t) < q^{\mathrm{th}}\} A_{\mathrm{c},i}(t)\biggr]
    + \expt \biggl[\mathbb{1}\{  Q_{\mathrm{c},i}(t) = 0\} A_{\mathrm{c},i}(t)\biggr]\nonumber\\
    = & \expt\biggl[
    \mathbb{1}\{ 0 <  Q_{\mathrm{c},i}(t) < q^{\mathrm{th}}\} \expt \bigl[A_{\mathrm{c},i}(t) | \lambda_{\alpha,i}(t), Q_{\mathrm{c},i}(t)  \bigr]
    \biggr]\nonumber\\
    & + \expt\biggl[ \mathbb{1}\{  Q_{\mathrm{c},i}(t)  = 0\} 
    \expt \big[ A_{\mathrm{c},i}(t) | \lambda'_i(t), Q_{\mathrm{c},i}(t)  \bigr]
    \biggr]\nonumber\\
    = & \expt\biggl[
    \mathbb{1}\{ 0 <  Q_{\mathrm{c},i}(t) < q^{\mathrm{th}}\} 
    \lambda_{\alpha,i}(t)
    + \mathbb{1}\{  Q_{\mathrm{c},i}(t)  = 0\} 
    \lambda'_i(t)
    \biggr]
\end{align}
where the second equality is by the law of iterated expectation and the last equality holds since the arrival $A_{\mathrm{c},i}(t)$ is independent of the queue length $\boldsymbol{Q}_{\mathrm{c}}(t)$ given the reduced arrival rate $\lambda_{\alpha,i}(t)$ and $A_{\mathrm{c},i}(t)$ is also independent of the queue length $\boldsymbol{Q}_{\mathrm{c}}(t)$ given the regular arrival rate $\lambda'_i(t)$. Similarly, for the server-side drift term \eqref{equ:drift-server-side}, we have
\begin{align}\label{equ:drift-server-side-1}
    \eqref{equ:drift-server-side} = \sum_j \kappa_{\mathrm{s},j} \left( 
    \sum_i \expt [X_{i,j}(t)] -  \expt [ A_{\mathrm{s},j}(t) ] \right),
\end{align}
where
\begin{align}\label{equ:drift-server-side-2}
    \expt [ A_{\mathrm{s},j}(t) ] 
    = \expt\biggl[
    \mathbb{1}\{ 0 < Q_{\mathrm{s},j}(t) < q^{\mathrm{th}}\} 
    \mu_{\alpha,j}(t)
    + \mathbb{1}\{  Q_{\mathrm{s},j}(t) = 0\} 
    \mu'_j(t)
    \biggr].
\end{align}
Combine \eqref{equ:drift-customer-side}-\eqref{equ:drift-server-side-2}, we have
\begin{align}\label{equ:avg-rate-drift-2}
    \expt [L(t+1) - L(t)]
    = & \sum_{(i,j)\in{\cal E}} 
    \left( \kappa_{\mathrm{c},i} + \kappa_{\mathrm{s},j}\right)
    \expt [X_{i,j}(t)]\nonumber\\
    &-\sum_i \kappa_{\mathrm{c},i}  
     \expt\biggl[
    \mathbb{1}\{ 0 < Q_{\mathrm{c},i}(t) < q^{\mathrm{th}}\} 
    \lambda_{\alpha,i}(t)
    + \mathbb{1}\{  Q_{\mathrm{c},i}(t) = 0\} 
    \lambda'_i(t)
    \biggr]\nonumber\\
    & - \sum_j \kappa_{\mathrm{s},j} 
    \expt\biggl[
    \mathbb{1}\{  0 < Q_{\mathrm{s},j}(t) < q^{\mathrm{th}}\} 
    \mu_{\alpha,j}(t)
    + \mathbb{1}\{  Q_{\mathrm{s},j}(t) = 0\} 
    \mu'_j(t)
    \biggr]
\end{align}
Recall from \eqref{equ:taylor-final}:
\begin{align*}
    |\max\{\lambda'_i(t) - \tilde{\alpha}_{\mathrm{c},i}(t), 0\} 
    - \lambda_{\alpha,i}(t)| 
    \le &  \Theta(\alpha^2) \mbox{ for all } i\nonumber\\
    |\max\{\mu'_j(t) - \tilde{\alpha}_{\mathrm{s},j}(t), 0\}
    - \mu_{\alpha, j}(t)| \le &  \Theta(\alpha^2) \mbox{ for all } j.
\end{align*}
Hence, \eqref{equ:avg-rate-drift-2} can be upper bounded by
\begin{align*}
    & \expt [L(t+1) - L(t)]\nonumber\\
    \le & \sum_{(i,j)\in{\cal E}} 
    \left( \kappa_{\mathrm{c},i} + \kappa_{\mathrm{s},j}\right)
    \expt [X_{i,j}(t)]
    + \Theta(\alpha^2)\nonumber\\
    & - \sum_i \kappa_{\mathrm{c},i}  
     \expt\biggl[
    \mathbb{1}\{  0 < Q_{\mathrm{c},i}(t) < q^{\mathrm{th}}\} 
    \max\{\lambda'_i(t) - \tilde{\alpha}_{\mathrm{c},i}(t), 0\}
    + \mathbb{1}\{  Q_{\mathrm{c},i}(t) = 0\} 
    \lambda'_i(t)
    \biggr]\nonumber\\
    & - \sum_j \kappa_{\mathrm{s},j} 
    \expt\biggl[
    \mathbb{1}\{  0 < Q_{\mathrm{s},j}(t) < q^{\mathrm{th}}\} 
    \max\{\mu'_j(t) - \tilde{\alpha}_{\mathrm{s},j}(t), 0\}
    + \mathbb{1}\{  Q_{\mathrm{s},j}(t) = 0\} 
    \mu'_j(t)
    \biggr].
\end{align*}
Adding the indicator of the ``good'' event ${\cal C}$ and using Lemma~\ref{lemma:concentration}, we obtain
\begin{align*}
    & \expt [L(t+1) - L(t)]\nonumber\\
    \le  & \sum_{(i,j)\in{\cal E}} 
    \left( \kappa_{\mathrm{c},i} + \kappa_{\mathrm{s},j}\right)
    \expt [\mathbb{1}_{\cal C} X_{i,j}(t)]
    + \Theta(\alpha^2)
    + \Theta\left(T\epsilon^{\frac{\beta}{2}+1} + \epsilon^{\frac{\beta}{2}-1} \log (1/\epsilon) \right)
    \nonumber\\
    & - \sum_i \kappa_{\mathrm{c},i}  
     \expt\biggl[\mathbb{1}_{\cal C}
    \mathbb{1}\{  0 < Q_{\mathrm{c},i}(t) < q^{\mathrm{th}}\} 
    \max\{\lambda'_i(t) - \tilde{\alpha}_{\mathrm{c},i}(t), 0\}
    + \mathbb{1}_{\cal C} \mathbb{1}\{  Q_{\mathrm{c},i}(t) = 0\} 
    \lambda'_i(t)
    \biggr]\nonumber\\
    & - \sum_j \kappa_{\mathrm{s},j} 
    \expt\biggl[\mathbb{1}_{\cal C}
    \mathbb{1}\{  0 < Q_{\mathrm{s},j}(t) < q^{\mathrm{th}}\} 
    \max\{\mu'_j(t) - \tilde{\alpha}_{\mathrm{s},j}(t), 0\}
    + \mathbb{1}_{\cal C} \mathbb{1}\{  Q_{\mathrm{s},j}(t) = 0\} 
    \mu'_j(t)
    \biggr].
\end{align*}
By Lemma~\ref{lemma:improved-bisection-error}, we can relate $\lambda'_i(t)$ to $\lambda^{\mathrm{tg}}_i(t)=\sum_{j\in {\cal E}_{\mathrm{c},i}} x_{i,j}(k(t))$, and relate $\mu'_j(t)$ to $\mu^{\mathrm{tg}}_j(t)=\sum_{i\in {\cal E}_{\mathrm{s},j}} x_{i,j}(k(t))$. Then we have
\begin{align*}
    & \expt [L(t+1) - L(t)]\nonumber\\
    \le  & \sum_{(i,j)\in{\cal E}} 
    \left( \kappa_{\mathrm{c},i} + \kappa_{\mathrm{s},j}\right)
    \expt [\mathbb{1}_{\cal C} X_{i,j}(t)]
    + \Theta(\alpha^2)\nonumber\\
    & + \Theta\left(T\epsilon^{\frac{\beta}{2}+1} + \epsilon^{\frac{\beta}{2}-1} \log (1/\epsilon) \right)
    + \Theta\left( \frac{\eta\epsilon}{\delta} + \eta + \delta + \epsilon \right)
    \nonumber\\
    & - \sum_i \kappa_{\mathrm{c},i}  
     \expt\biggl[\mathbb{1}_{\cal C}
    \mathbb{1}\{  0 < Q_{\mathrm{c},i}(t) < q^{\mathrm{th}}\} 
    \max\{\lambda^{\mathrm{tg}}_i(t) - \tilde{\alpha}_{\mathrm{c},i}(t), 0\}
    + \mathbb{1}_{\cal C} \mathbb{1}\{  Q_{\mathrm{c},i}(t) = 0\} 
    \lambda^{\mathrm{tg}}_i(t)
    \biggr]\nonumber\\
    & - \sum_j \kappa_{\mathrm{s},j} 
    \expt\biggl[\mathbb{1}_{\cal C}
    \mathbb{1}\{  0 < Q_{\mathrm{s},j}(t) < q^{\mathrm{th}}\} 
    \max\{\mu^{\mathrm{tg}}_j(t) - \tilde{\alpha}_{\mathrm{s},j}(t), 0\}
    + \mathbb{1}_{\cal C} \mathbb{1}\{  Q_{\mathrm{s},j}(t) = 0\} 
    \mu^{\mathrm{tg}}_j(t)
    \biggr].
\end{align*}
Recall that $\lambda^{\mathrm{tg}}_i(t) - \tilde{\alpha}_{\mathrm{c},i}(t) \ge 0$ and $\mu^{\mathrm{tg}}_j(t) - \tilde{\alpha}_{\mathrm{s},j}(t) \ge 0$ for sufficiently large $T$. Then we have
\begin{align*}
    & \expt [L(t+1) - L(t)]\nonumber\\
    \le  & \sum_{(i,j)\in{\cal E}} 
    \left( \kappa_{\mathrm{c},i} + \kappa_{\mathrm{s},j}\right)
    \expt [\mathbb{1}_{\cal C} X_{i,j}(t)]
    + \Theta(\alpha^2)\nonumber\\
    & + \Theta\left(T\epsilon^{\frac{\beta}{2}+1} + \epsilon^{\frac{\beta}{2}-1} \log (1/\epsilon) \right)
    + \Theta\left( \frac{\eta\epsilon}{\delta} + \eta + \delta + \epsilon \right)
    \nonumber\\
    & - \sum_i \kappa_{\mathrm{c},i}  
     \expt\biggl[\mathbb{1}_{\cal C}
    \mathbb{1}\{  0 < Q_{\mathrm{c},i}(t) < q^{\mathrm{th}}\} 
    \left(\lambda^{\mathrm{tg}}_i(t) - \tilde{\alpha}_{\mathrm{c},i}(t)\right)
    + \mathbb{1}_{\cal C} \mathbb{1}\{  Q_{\mathrm{c},i}(t) = 0\} 
    \lambda^{\mathrm{tg}}_i(t)
    \biggr]\nonumber\\
    & - \sum_j \kappa_{\mathrm{s},j} 
    \expt\biggl[\mathbb{1}_{\cal C}
    \mathbb{1}\{  0 < Q_{\mathrm{s},j}(t) < q^{\mathrm{th}}\} 
    \left(\mu^{\mathrm{tg}}_j(t) - \tilde{\alpha}_{\mathrm{s},j}(t)\right)
    + \mathbb{1}_{\cal C} \mathbb{1}\{  Q_{\mathrm{s},j}(t) = 0\} 
    \mu^{\mathrm{tg}}_j(t)
    \biggr]\nonumber\\
    = & \sum_{(i,j)\in{\cal E}} 
    \left( \kappa_{\mathrm{c},i} + \kappa_{\mathrm{s},j}\right)
    \expt [\mathbb{1}_{\cal C} X_{i,j}(t)]
    + \Theta(\alpha^2)\nonumber\\
    & + \Theta\left(T\epsilon^{\frac{\beta}{2}+1} + \epsilon^{\frac{\beta}{2}-1} \log (1/\epsilon) \right)
    + \Theta\left( \frac{\eta\epsilon}{\delta} + \eta + \delta + \epsilon \right)\nonumber\\
    & + \sum_i \kappa_{\mathrm{c},i}  
     \expt\biggl[ \mathbb{1}_{\cal C}
    \mathbb{1}\{  0 < Q_{\mathrm{c},i}(t) < q^{\mathrm{th}}\} 
     \tilde{\alpha}_{\mathrm{c},i}(t)
    \biggr]
    + \sum_j \kappa_{\mathrm{s},j} 
    \expt\biggl[ \mathbb{1}_{\cal C}
    \mathbb{1}\{  0 < Q_{\mathrm{s},j}(t) < q^{\mathrm{th}}\} 
    \tilde{\alpha}_{\mathrm{s},j}(t)
    \biggr]\nonumber\\
    & + \expt\biggl[ \mathbb{1}_{\cal C}
    \sum_{(i,j)\in{\cal E}} x_{i,j}(k(t)) 
    \left(
    -\kappa_{\mathrm{c},i}
    \mathbb{1}\{  Q_{\mathrm{c},i}(t) < q^{\mathrm{th}}\} -\kappa_{\mathrm{s},j}
    \mathbb{1}\{  Q_{\mathrm{s},j}(t) < q^{\mathrm{th}}\}
    \right)
    \biggr].
\end{align*}
Adding the indicator of the event ${\cal H}_t$, we obtain
\begin{align*}
    & \expt [L(t+1) - L(t)]\nonumber\\
    \le  & \sum_{(i,j)\in{\cal E}} 
    \left( \kappa_{\mathrm{c},i} + \kappa_{\mathrm{s},j}\right)
    \expt [\mathbb{1}_{\cal C} \mathbb{1}_{{\cal H}_t} X_{i,j}(t)]
    + \Theta(\alpha^2)\nonumber\\
    & + \Theta\left(T\epsilon^{\frac{\beta}{2}+1} + \epsilon^{\frac{\beta}{2}-1} \log (1/\epsilon) \right)
    + \Theta\left( \frac{\eta\epsilon}{\delta} + \eta + \delta + \epsilon \right)
    \nonumber\\
    & +  
    \expt\Biggl[ \mathbb{1}_{\cal C}
    \mathbb{1}_{{\cal H}_t}
    \left(
    \sum_i \kappa_{\mathrm{c},i} 
    \mathbb{1}\{  0 < Q_{\mathrm{c},i}(t) < q^{\mathrm{th}}\} 
     \tilde{\alpha}_{\mathrm{c},i}(t)
    + 
    \sum_j \kappa_{\mathrm{s},j}
    \mathbb{1}\{  0 < Q_{\mathrm{s},j}(t) < q^{\mathrm{th}}\} 
    \tilde{\alpha}_{\mathrm{s},j}(t)
    \right)
    \Biggr]\nonumber\\
    & + \expt\biggl[ \mathbb{1}_{\cal C}
    \mathbb{1}_{{\cal H}_t}
    \sum_{(i,j)\in{\cal E}} x_{i,j}(k(t)) 
    \left(
    -\kappa_{\mathrm{c},i}
    -\kappa_{\mathrm{s},j}
    \right)
    \biggr]
    + \Theta \left( \expt [\mathbb{1}_{\cal C}
    \mathbb{1}_{{\cal H}^{\mathrm{c}}_t}]\right)\nonumber\\
    = &
    \expt\Biggl[ \mathbb{1}_{\cal C}
    \mathbb{1}_{{\cal H}_t}
    \left(
    \sum_i \kappa_{\mathrm{c},i} 
    \mathbb{1}\{  0 < Q_{\mathrm{c},i}(t) < q^{\mathrm{th}}\} 
     \tilde{\alpha}_{\mathrm{c},i}(t)
    + 
    \sum_j \kappa_{\mathrm{s},j}
    \mathbb{1}\{  0 < Q_{\mathrm{s},j}(t) < q^{\mathrm{th}}\} 
    \tilde{\alpha}_{\mathrm{s},j}(t)
    \right)
    \Biggr]\nonumber\\
    & + \expt\biggl[ \mathbb{1}_{\cal C}
    \mathbb{1}_{{\cal H}_t}
    \sum_{(i,j)\in{\cal E}}
    \left(
    \kappa_{\mathrm{c},i}
    +\kappa_{\mathrm{s},j}
    \right)\left(
    X_{i,j}(t) - x_{i,j}(k(t)) 
    \right)
    \biggr]
    + \Theta \left( \expt [\mathbb{1}_{\cal C}
    \mathbb{1}_{{\cal H}^{\mathrm{c}}_t}]\right)\nonumber\\
    & + \Theta(\alpha^2)
    + \Theta\left(T\epsilon^{\frac{\beta}{2}+1} + \epsilon^{\frac{\beta}{2}-1} \log (1/\epsilon) \right)
    + \Theta\left( \frac{\eta\epsilon}{\delta} + \eta + \delta + \epsilon \right).
\end{align*}
Taking summation over $t$ from $t=1$ to $t=T$, we obtain
\begin{align}\label{equ:avg-rate-drift-3}
    & \expt \left [\sum_i (-\kappa_{\mathrm{c},i}) Q_{\mathrm{c},i}(T+1) + \sum_j (-\kappa_{\mathrm{s},j}) Q_{\mathrm{s},j}(T+1) \right]\nonumber\\
    \le &  
    \expt\Biggl[
    \sum_{t=1}^{T}
    \mathbb{1}_{\cal C}
    \mathbb{1}_{{\cal H}_t}
    \left(
    \sum_i \kappa_{\mathrm{c},i} 
    \mathbb{1}\{  0 < Q_{\mathrm{c},i}(t) < q^{\mathrm{th}}\} 
     \tilde{\alpha}_{\mathrm{c},i}(t)
    + 
    \sum_j \kappa_{\mathrm{s},j}
    \mathbb{1}\{  0 < Q_{\mathrm{s},j}(t) < q^{\mathrm{th}}\} 
    \tilde{\alpha}_{\mathrm{s},j}(t)
    \right)
    \Biggr]\nonumber\\
    & + \expt\biggl[ 
    \sum_{t=1}^{T}
    \mathbb{1}_{\cal C}
    \mathbb{1}_{{\cal H}_t}
    \sum_{(i,j)\in{\cal E}}
    \left(
    \kappa_{\mathrm{c},i}
    +\kappa_{\mathrm{s},j}
    \right)\left(
    X_{i,j}(t) - x_{i,j}(k(t)) 
    \right)
    \biggr]
    + \Theta \left( \sum_{t=1}^{T} \expt [\mathbb{1}_{\cal C}
    \mathbb{1}_{{\cal H}^{\mathrm{c}}_t}]\right)\nonumber\\
    & + \Theta(T\alpha^2)
    + \Theta\left(T^2\epsilon^{\frac{\beta}{2}+1} + T\epsilon^{\frac{\beta}{2}-1} \log (1/\epsilon) \right)
    + \Theta\left( T \left(\frac{\eta\epsilon}{\delta} + \eta + \delta + \epsilon \right)\right)
\end{align}
By Lemma~\ref{lemma:theo-3-bound-time-large-queue} and \eqref{equ:theo-3-regret-term-3}, we have 
\begin{align}\label{equ:bound-H-t-c}
    \sum_{t=1}^{T} \expt [\mathbb{1}_{\cal C}
    \mathbb{1}_{{\cal H}^{\mathrm{c}}_t}] \le \Theta \left(
    \frac{T}{q^{\mathrm{th}}}
    + T^2\epsilon^{\frac{\beta}{2}+1} + T\epsilon^{\frac{\beta}{2}-1} \log (1/\epsilon) 
    + T \left(\frac{\eta \epsilon}{\delta} + \eta + \delta + \epsilon\right)  \right)
\end{align}
From \eqref{equ:avg-rate-drift-3} and \eqref{equ:bound-H-t-c}, we have
\begin{align*}
    & \expt \left [\sum_i (-\kappa_{\mathrm{c},i}) Q_{\mathrm{c},i}(T+1) + \sum_j (-\kappa_{\mathrm{s},j}) Q_{\mathrm{s},j}(T+1) \right]\nonumber\\
    \le &  
    \expt\Biggl[
    \sum_{t=1}^{T}
    \mathbb{1}_{\cal C}
    \mathbb{1}_{{\cal H}_t}
    \left(
    \sum_i \kappa_{\mathrm{c},i} 
    \mathbb{1}\{ 0 < Q_{\mathrm{c},i}(t) < q^{\mathrm{th}}\} 
     \tilde{\alpha}_{\mathrm{c},i}(t)
    + 
    \sum_j \kappa_{\mathrm{s},j}
    \mathbb{1}\{ 0 < Q_{\mathrm{s},j}(t) < q^{\mathrm{th}}\} 
    \tilde{\alpha}_{\mathrm{s},j}(t)
    \right)
    \Biggr]\nonumber\\
    & + \expt\biggl[ 
    \sum_{t=1}^{T}
    \mathbb{1}_{\cal C}
    \mathbb{1}_{{\cal H}_t}
    \sum_{(i,j)\in{\cal E}}
    \left(
    \kappa_{\mathrm{c},i}
    +\kappa_{\mathrm{s},j}
    \right)\left(
    X_{i,j}(t) - x_{i,j}(k(t)) 
    \right)
    \biggr]
    \nonumber\\
    & + \Theta(T\alpha^2)
    + \Theta\left(T^2\epsilon^{\frac{\beta}{2}+1} + T\epsilon^{\frac{\beta}{2}-1} \log (1/\epsilon) \right)
    + \Theta\left( T \left(\frac{\eta\epsilon}{\delta} + \eta + \delta + \epsilon \right)\right) 
    + \Theta \left(
    \frac{T}{q^{\mathrm{th}}} \right).
\end{align*}
Hence, we have
\begin{align}
    & \expt\Biggl[
    \sum_{t=1}^{T}
    \mathbb{1}_{\cal C}
    \mathbb{1}_{{\cal H}_t}
    \biggl(
    \sum_i (- \kappa_{\mathrm{c},i} )
    \mathbb{1}\{ 0 < Q_{\mathrm{c},i}(t) < q^{\mathrm{th}}\} 
     \tilde{\alpha}_{\mathrm{c},i}(t)\nonumber\\
    & \qquad \qquad \quad + 
    \sum_j ( - \kappa_{\mathrm{s},j} )
    \mathbb{1}\{ 0 < Q_{\mathrm{s},j}(t) < q^{\mathrm{th}}\} 
    \tilde{\alpha}_{\mathrm{s},j}(t)
    \biggr)
    \Biggr]\label{equ:avg-rate-drift-4}\\
    & + \expt\biggl[ 
    \sum_{t=1}^{T}
    \mathbb{1}_{\cal C}
    \mathbb{1}_{{\cal H}_t}
    \sum_{(i,j)\in{\cal E}}
    \left(
    \kappa_{\mathrm{c},i}
    +\kappa_{\mathrm{s},j}
    \right)\left(
    x_{i,j}(k(t)) - X_{i,j}(t)
    \right)
    \biggr]\label{equ:avg-rate-drift-5}\\
    \le &  
    \expt \left [\sum_i \kappa_{\mathrm{c},i} Q_{\mathrm{c},i}(T+1) + \sum_j \kappa_{\mathrm{s},j} Q_{\mathrm{s},j}(T+1) \right]
    \nonumber\\
    & + \Theta(T\alpha^2)
    + \Theta\left(T^2\epsilon^{\frac{\beta}{2}+1} + T\epsilon^{\frac{\beta}{2}-1} \log (1/\epsilon) \right)
    + \Theta\left( T \left(\frac{\eta\epsilon}{\delta} + \eta + \delta + \epsilon \right)\right) 
    + \Theta \left(
    \frac{T}{q^{\mathrm{th}}} \right)\nonumber\\
    \le &
    \Theta\left(T\alpha^2 + T^2\epsilon^{\frac{\beta}{2}+1} + T\epsilon^{\frac{\beta}{2}-1} \log (1/\epsilon) 
    + T \left(\frac{\eta\epsilon}{\delta} + \eta + \delta + \epsilon \right)
    + \frac{T}{q^{\mathrm{th}}} 
    + q^{\mathrm{th}} \right),\label{equ:avg-rate-drift-final}
\end{align}
where the last inequality holds since the length of any queue is no greater than $q^{\mathrm{th}}$ for all time because we use queue-length control all the time.
Note that if $Q_{\mathrm{c},i}(t) = 0$, $\tilde{\alpha}_{\mathrm{c},i}(t)=0$ and if $Q_{\mathrm{s},j}(t) = 0$, $\tilde{\alpha}_{\mathrm{s},j}(t)=0$. Also note that ${\cal H}_t$ holds in \eqref{equ:avg-rate-drift-4}.
Hence, we can drop the indicators $\mathbb{1}\{ 0 < Q_{\mathrm{c},i}(t) < q^{\mathrm{th}}\}$ and $\mathbb{1}\{ 0 < Q_{\mathrm{s},j}(t) < q^{\mathrm{th}}\}$ in \eqref{equ:avg-rate-drift-4}.
Then, writing the result $\eqref{equ:avg-rate-drift-4} + \eqref{equ:avg-rate-drift-5} \le \eqref{equ:avg-rate-drift-final}$ in a vector form, we have
\begin{align*}
    & \sum_{t=1}^{T} \expt \Biggl[
    \mathbb{1}_{\cal C} \mathbb{1}_{{\cal H}_t}  
    \left[
    \tilde{\boldsymbol{\alpha}}_{\mathrm{c}}(t)^\top 
    (-\boldsymbol{\kappa}_{\mathrm{c}})
    + \tilde{\boldsymbol{\alpha}}_{\mathrm{s}}(t)^\top 
    (-\boldsymbol{\kappa}_{\mathrm{s}})
    + (\boldsymbol{x}(k(t)) - \boldsymbol{X}(t))^\top \boldsymbol{\kappa}
    \right]
    \Biggr] \nonumber\\
    \le &  \Theta\left(T\alpha^2 + T^2\epsilon^{\frac{\beta}{2}+1} + T\epsilon^{\frac{\beta}{2}-1} \log (1/\epsilon) 
    + T \left(\frac{\eta\epsilon}{\delta} + \eta + \delta + \epsilon \right)
    + \frac{T}{q^{\mathrm{th}}} 
    + q^{\mathrm{th}} \right).
\end{align*}
Lemma~\ref{lemma:avg-rate-balanced} is proved.

\end{document}